\documentclass[journal]{IEEEtran}
\usepackage{pgf,tikz}
\usepackage{tkz-euclide}
\usepackage{tkz-graph}
\usepackage{pgfplots}
\newlength\figureheight 
\newlength\figurewidth 

\usepackage{algpseudocode}
\usepackage{algorithm}
\usepackage{xcolor}
\usepackage{graphicx,overpic}
\usepackage{amsmath,amssymb}

\usepackage{float,lipsum}
\restylefloat{table}
\usepackage{mdframed}

\usepackage{multicol}

\usepackage{amsmath}

\usepackage{paracol}
\usepackage{xparse}
\usepackage{kantlipsum}

\NewDocumentCommand\myframedtext{ s O{.9\linewidth} m }{%
    \IfBooleanTF{#1}{\begin{figure*}[h!t]}{\begin{figure}[h!t]}%
      \centering%
      \fboxsep=2.5mm
      \fboxrule=2pt
      \fcolorbox{lightgray}{white}{\parbox{#2}{%
        #3%
      }}%
    \IfBooleanTF{#1}{\end{figure*}}{\end{figure}}}

\newcommand {\R} {{\mathbb{R}}}
\newcommand {\E} {{\mathbb{E}}}
\newcommand {\w}{{\gamma}}
\newcommand {\W}{{\Gamma}}

\newcommand{\editMB}[1]{{\color{black}#1}}
\newcommand{\editJB}[1]{{\color{black}#1}}
\newcommand{\editADS}[1]{{\color{black}#1}}

\begin{document}
\title{ Geometric deep learning: \\going beyond Euclidean data}

\author{Michael M. Bronstein, Joan Bruna, Yann LeCun, Arthur Szlam, Pierre Vandergheynst
%Michael~Shell,~\IEEEmembership{Member,~IEEE,}
%        John~Doe,~\IEEEmembership{Fellow,~OSA,}
%        and~Jane~Doe,~\IEEEmembership{Life~Fellow,~IEEE}% <-this % stops a space
\thanks{MB is with USI Lugano, Switzerland, Tel Aviv University, and Intel Perceptual Computing, Israel. JB is with Courant Institute, NYU and UC Berkeley, USA. YL with with Facebook AI Research and NYU, USA. AS is with Facebook AI Research, USA. PV is with EPFL, Switzerland. }
}


\markboth{IEEE Sig Proc Mag}%
{Bronstein \MakeLowercase{\textit{et al.}}: Geometric deep learning}



\maketitle

%\begin{abstract}
%The abstract goes here.
%\end{abstract}

%\begin{IEEEkeywords}
%IEEE, IEEEtran, journal, \LaTeX, paper, template.
%\end{IEEEkeywords}
%
%

% \fi
%\IEEEpeerreviewmaketitle

%\IEEEPARstart{I}{n} 

\editADS{Many scientific fields study data with an underlying structure that is a non-Euclidean space. Some}
\editMB{ examples include social networks in computational social sciences, sensor networks in communications, functional networks in brain imaging, regulatory networks in genetics, and meshed surfaces in computer graphics. % used to represent 3D objects. 
%
%the characteristics of users can be modeled as signals on the vertices of the social graph \cite{lazer2009life}. Sensor networks are graph models of distributed interconnected sensors, whose readings are modelled as time-dependent signals on the vertices. In genetics, gene expression data are modeled as signals defined on the regulatory network \cite{davidson2002genomic}. 
%%
%In neuroscience, graph models are used to represent anatomical and functional structures of the brain. 
%%Modeling data given as points in a high-dimensional Euclidean space using nearest neighbor graphs is an increasingly popular trend in data science, allowing practitioners access to the intrinsic structure of the data.
%%
%In computer graphics and vision, 3D objects are modeled as Riemannian manifolds (surfaces) endowed with properties such as color texture. Even more complex examples include networks of operators, e.g., functional correspondences \cite{ovsjanikov2012functional} or difference operators \cite{rustamov2013map} in a collection of 3D shapes, or orientations of overlapping cameras in multi-view vision (``structure from motion'') problems \cite{arie2012global}. 
%
%
In many applications, such geometric data are} \editADS{ large and complex (in the case of social networks, on the scale of billions), and  are natural targets for machine learning techniques.   
In particular, we would like to use deep neural networks, which have recently proven to be powerful tools for a broad range of problems from computer vision, natural language processing, and audio analysis.  However, these tools have been most successful on data with an underlying Euclidean or grid-like structure, and in cases where the invariances of these structures are built into networks used to model them.  

{\em Geometric deep learning} is an umbrella term for emerging techniques attempting to generalize (structured) deep neural models to non-Euclidean domains such as graphs and manifolds. 
The purpose of this paper is to overview different examples of geometric deep learning problems and present available solutions, key difficulties, applications, and future research directions in this nascent field.
}

\section{Introduction}
\label{sec:intro}

``Deep learning'' refers to learning complicated concepts by building them from simpler ones in a hierarchical or Òmulti-layerÓ manner. Artificial neural networks are popular realizations of such deep multi-layer hierarchies. % inspired by the signal processing in the human brain. 
In the past few years, the growing computational power of modern GPU-based computers and the availability of large training datasets
%, and efficient stochastic optimization methods 
have allowed successfully  training neural networks with many layers and degrees of freedom \cite{lecun2015deep}. \editMB{This has led to qualitative breakthroughs on a wide variety of tasks, from speech recognition \cite{mikolov2011strategies,hinton2012deep} and machine translation \cite{sutskever2014sequence} to image analysis and computer vision \cite{lecun2010convolutional,cirecsan2011committee,krizhevsky2012imagenet,farabet2013learning,taigman2014deepface,simonyan2014very,he2015deep} (the reader is referred to \cite{deng2014deep,Goodfellow-et-al-2016-Book} for many additional examples of successful applications of deep learning). Nowadays, deep learning has matured into a technology that is widely used in commercial applications, including Siri speech recognition in Apple iPhone, Google text translation, and Mobileye vision-based technology for autonomously driving cars.}
 
%Several novel fields of applications of deep learning, such as Deep Genomics \cite{alipanahi2015predicting}, have the promise to become groundbreaking.

\editMB{
One of the key reasons for the success of deep neural networks is their ability to leverage statistical properties of the data such as stationarity and compositionality through local statistics, which are present in natural images, video, and speech \cite{simoncelli2001natural, field1989statistics}. %, are one of the key reasons for the success of deep neural networks in these domains.
% which were initially developed to model aspects of the visual cortex \cite{hopfield1982neural,olshausen1997sparse}. 
These statistical properties have been related to physics %via the renormalization group 
\cite{mehta2014exact} and formalized in specific classes of convolutional neural networks (CNNs) \cite{mallat2012group,bruna2013invariant,tygert2016mathematical}.
%{\color{blue} JB: updated with references}.
%{\color{green} ADS: lots more references in this seciton, no? specifically towards  images: http://www.naturalimagestatistics.net/, http://redwood.psych.cornell.edu/papers/field1989.pdf, and the Olshausen and Field sparse coding paper.    For compositionality and pooling etc, the renormalization paper http://arxiv.org/abs/1410.3831 and some scattering papers, and maybe Mark's version too }
%For example, one can think of images as functions on the Euclidean space (plane), sampled on a grid. 
In image analysis applications, one can consider images as functions on the Euclidean space (plane), sampled on a grid. }
In this setting, stationarity is owed to shift-invariance, locality is due to the local connectivity, and compositionality stems from the multi-resolution structure of the grid. These properties are exploited by convolutional architectures \cite{lecun1989backpropagation}, which are built of alternating convolutional and downsampling (pooling) layers. The use of convolutions has a two-fold effect. First, it allows extracting local features that are shared across the image domain and greatly reduces the number of parameters in the network with respect to generic deep architectures (and thus also the risk of overfitting), without sacrificing the expressive capacity of the network. Second, %as we will show in the following, 
the convolutional architecture itself imposes some priors about the data, which appear very suitable especially for natural images \cite{goodfellow2013maxout,bruna2013invariant,mallat2012group,tygert2016mathematical}. 

\editMB{
%Dealing with signals such as speech, images, or video on 1D-, 2D-, and 3D Euclidean domains, respectively, has been the main focus of research in deep learning for the past decades. 
%{\color{blue}
%While deep learning models have been particularly successful when dealing with signals such as speech, images, or video, in which there is an underlying Euclidean structure, recently there has been a growing interest in trying to apply learning on non-Euclidean geometric data.   For example in computer graphics and vision \cite{masci2015geodesic,boscaini2015learning,boscaini2016learning}, natural language processing \cite{bengio2003neural, mikolov2013efficient}, social network analysis \cite{perozzi2014deepwalk,tang2015line,cao2015grarep}, and biology \cite{alipanahi2015predicting,duvenaud2015convolutional}. 
While deep learning models have been particularly successful when dealing with signals such as speech, images, or video, in which there is an underlying Euclidean structure, recently there has been a growing interest in trying to apply learning on non-Euclidean geometric data. 
%, for example, in computer graphics and vision \cite{masci2015geodesic,boscaini2015learning,boscaini2016learning}, natural language processing \cite{bengio2003neural}, and biology \cite{alipanahi2015predicting,duvenaud2015convolutional}. 
%%{\color{green} ADS: many of these are not deep learning at all. word2vec and its variants are explicitly, expressly, purposefully *not* deep; that was Tomas Mikolov's point. we really should cut those references here... }  
%
Such kinds of data arise in numerous applications. For instance, in social networks, the characteristics of users can be modeled as signals on the vertices of the social graph \cite{lazer2009life}. Sensor networks are graph models of distributed interconnected sensors, whose readings are modelled as time-dependent signals on the vertices. In genetics, gene expression data are modeled as signals defined on the regulatory network \cite{davidson2002genomic}. 
In neuroscience, graph models are used to represent anatomical and functional structures of the brain. 
%Modeling data given as points in a high-dimensional Euclidean space using nearest neighbor graphs is an increasingly popular trend in data science, allowing practitioners access to the intrinsic structure of the data.
%
In computer graphics and vision, 3D objects are modeled as Riemannian manifolds (surfaces) endowed with properties such as color texture. 
}

%Even more complex examples include networks of operators, e.g., functional correspondences \cite{ovsjanikov2012functional} or difference operators \cite{rustamov2013map} in a collection of 3D shapes, or orientations of overlapping cameras in multi-view vision (``structure from motion'') problems \cite{arie2012global}. 

The non-Euclidean nature of such data implies that there are no such familiar properties as global parameterization, common system of coordinates, vector space structure, or shift-invariance. Consequently, basic operations like convolution that are taken for granted in the Euclidean case are even not well defined on non-Euclidean domains. %This major obstacle has hindered the application of successful deep learning methods such as convolutional networks on data with non-Euclidean underlying structure.  
\editADS{The purpose of our paper is to show different methods of translating the key ingredients of successful deep learning methods such as convolutional neural networks to non-Euclidean data.}

%As a result, the quantitative and qualitative breakthrough that deep learning methods have brought into speech recognition, natural language processing, and computer vision has not yet come to fields dealing with functions defined on more general geometric data. 

\pagebreak

\section{Geometric learning problems}
\label{sec:problems}
\editADS{
%\paragraph*{\bf Landscape} 
Broadly speaking, we can distinguish between two classes
of geometric learning problems. In the first class of problems,
the goal is to {\it characterize the structure} of the data.
The second class of problems deals with {\it analyzing functions}
defined on a given non-Euclidean domain.
These two classes are related, since understanding the properties of functions defined on a domain conveys certain information about the domain, and vice-versa, the structure of the domain imposes certain properties on the functions on it. 

%dealing with the structure of the .  
%%\paragraph*{\bf Building a domain} 
%

}

%%\editMB{
%%%\paragraph*{\bf Landscape} 
%%Broadly speaking, geometric learning problems can be divided into two classes. In the first class of problems, the data are represented or approximated as a non-Euclidean domain (e.g. graph or manifold) and the goal is to study the {\em structure} of that domain. In the second class of problems, the {\em data are given on a non-Euclidean domain} (e.g. in the form of a function defined on the graph or manifold), and the goal is to study these data accounting correctly for the structure of the domain. 
%%%
%%These two classes are related, since understanding the properties of functions defined on a domain conveys certain information about the domain, and vice-versa, the structure of the domain imposes certain properties on the functions on it. 
%%
%%%dealing with the structure of the .  
%%%%\paragraph*{\bf Building a domain} 
%%%
%%
%%}

{\bf{\em Structure of the domain:} }
\editMB{
As an example of the first class of problems, %the goal is to characterize the {\em structure} of the data.  
assume to be given a set of data points with some underlying lower dimensional structure embedded into a high-dimensional Euclidean space. Recovering that lower dimensional structure is often  referred to as {\em manifold learning}}\footnote{Note that the notion of ``manifold'' in this setting can be  considerably more general than a classical smooth manifold; see e.g. \cite{non-differentiable_image,Verma:adaptive}}  or {\em non-linear dimensionality reduction}, and is an instance of unsupervised learning.  
Many methods for non-linear dimensionality reduction consist of two steps: first, they start with constructing a representation of local affinity of the data points (typically, a sparsely connected graph). Second, the data points are embedded into a low-dimensional space trying to preserve some criterion of the original affinity. For example, spectral embeddings tend to map points with many connections between them to nearby locations, and MDS-type methods try to preserve global information such as graph geodesic distances.   Examples of manifold learning include different flavors of multidimensional scaling (MDS) \cite{tenenbaum2000global}, locally linear embedding (LLE) \cite{roweis2000nonlinear}, stochastic neighbor embedding (t-SNE) \cite{maaten2008visualizing}, spectral embeddings such as Laplacian eigenmaps \cite{belkin2003laplacian} and diffusion maps \cite{coifman2006diffusion}, and deep models \cite{hadsell2006dimensionality}. 
Most recent approaches \cite{perozzi2014deepwalk,tang2015line,cao2015grarep} tried to apply the successful word embedding model  \cite{mikolov2013efficient} to graphs.  Instead of embedding the vertices, the graph structure can be processed by decomposing it into small sub-graphs called {\em motifs} \cite{milo2002network} or  {\em graphlets} \cite{prvzulj2007biological}.

In some cases, the data are presented as a manifold or graph at the outset, and the first step of constructing the affinity structure described above is unnecessary. 
% (at the cost of having to use the graph structure given, instead of the practitioner getting to choose it). 
For instance, in computer graphics and vision applications, one can analyze 3D shapes represented as meshes by constructing local geometric descriptors capturing e.g. curvature-like properties \cite{sun2009concise,litman2014learning}. 
In network analysis applications such as computational sociology, the topological structure of the social graph  representing the social relations between people carries important insights allowing, for example, to classify the vertices and detect communities \cite{fortunato2010community}. 
In natural language processing, words in a corpus can be represented by the co-occurrence graph, where two words are connected  if they often appear near each other \cite{mikolov2013distributed}.

{\bf{\em  Data on a domain:} }
Our second class of problems deals with analyzing {\em functions} defined on a given non-Euclidean domain.   
We can further break down such problems into two subclasses: problems where the domain is {\em fixed} and those where {\em multiple domains} are given. 
For example, assume that we are given the geographic coordinates of the users of a social network, represented as a time-dependent signal on the vertices of the social graph. An important application in location-based social networks is to predict the position of the user given his or her past behavior, as well as that of his or her friends \cite{cho2011friendship}. 
In this problem, the domain (social graph) is assumed to be fixed; methods of {\em signal processing on graphs}, which have previously been reviewed in this Magazine \cite{shuman2013emerging}, can be applied to this setting, in particular, in order to define an operation similar to convolution in the spectral domain. This, in turn, allows generalizing  CNN models to graphs \cite{henaff2015deep,defferrard2016convolutional}. 

In computer graphics and vision applications, finding similarity and correspondence between shapes are examples of the second sub-class of problems: each shape is modeled as a manifold, and one has to work with multiple such domains. 
In this setting, a generalization of convolution in the spatial domain using local charting  \cite{atwood2016search,masci2015geodesic,boscaini2016learning} appears to be more appropriate.

%In computer graphics applications, a generalization of CNNs working on local intrinsic local patches was applied on 3D deformable shapes modeled as manifolds to address problems such as correspondence and similarity \cite{masci2015geodesic,boscaini2016learning}.

\editMB{
\paragraph*{\bf Brief history}
The main focus of this review is on this second class of problems, namely learning functions on non-Euclidean structured domains, and in particular, attempts to generalize the popular CNNs to such settings. 
First attempts to generalize neural networks to graphs we are aware of are due to Scarselli et al. \cite{gori2005new}, who proposed a scheme combining recurrent neural networks and random walk models. 
This approach went almost unnoticed, re-emerging in a modern form in \cite{GGSNN,comnets} due to the renewed recent interest in deep learning. 
The first formulation of CNNs on graphs is due to Bruna et al. \cite{bruna2013spectral}, who used the definition of convolutions in the spectral domain. Their paper, while being of conceptual importance, came with significant computational drawbacks that fell short of a truly useful method. These drawbacks were subsequently addressed in the followup works of Henaff et al. \cite{henaff2015deep} and Defferrard et al. \cite{defferrard2016convolutional}. In the latter paper, graph CNNs allowed achieving some state-of-the-art results.

In a parallel effort in the computer vision and graphics community, Masci et al. \cite{masci2015geodesic} showed the first CNN model on meshed surfaces, resorting to a spatial definition of the convolution operation based on local intrinsic patches. Among other applications, such models were shown to achieve state-of-the-art performance in finding correspondence between deformable 3D shapes. Followup works proposed  different construction of intrinsic patches on point clouds \cite{boscaini2016anisotropic,boscaini2016learning} and general graphs \cite{monti2016geometric}.

The interest in deep learning on graphs or manifolds has exploded in the past year, resulting in numerous attempts to apply these methods in a broad spectrum of problems ranging from biochemistry \cite{duvenaud2015convolutional} to recommender systems \cite{monti2017mc}. 
Since such applications originate in different fields that usually do not cross-fertilize, publications in this domain tend to use different terminology and notation, making it difficult for a newcomer to grasp the foundations and current state-of-the-art methods. We believe that our paper comes at the right time attempting to systemize and bring some order into the field. 
%However, coming from different fields, there is 

%We will consider both single- and multiple-domain cases. 
%

\paragraph*{\bf Structure of the paper}
We start with an overview of traditional Euclidean deep learning in Section III, summarizing the important assumptions about the data, and how they are realized in convolutional network architectures.\footnote{For a more in-depth review of CNNs and their applications, we refer the reader to \cite{deng2014deep,lecun2015deep,Goodfellow-et-al-2016-Book} and references therein.}

Going to the non-Euclidean world in Section IV, we then define basic notions in differential geometry and graph theory. These topics are insufficiently known in the signal processing community, and to our knowledge, there is no introductory-level reference treating these so different structures in a common way. One of our goals is to provide an accessible overview of these models resorting as much as possible to the intuition of traditional signal processing. 

In Sections V--VIII, we overview the main geometric deep learning paradigms, emphasizing the similarities and the differences between Euclidean and non-Euclidean learning methods. 
The key difference between these approaches is in the way a convolution-like operation is formulated on graphs and manifolds. One way is to resort to the analogy of the Convolution Theorem, defining the convolution in the spectral domain. An alternative is to think of the convolution as a template matching in the spatial domain. 
Such a distinction is, however, far from being a clear-cut: as we will see, some approaches though draw their formulation from the spectral domain, essentially boil down to applying filters in the spatial domain. 
It is also possible to combine these two approaches resorting to spatio-frequency analysis techniques, such as wavelets or the windowed Fourier transform. 
%and distinguish between spatial- and spectral domain learning methods. 
%We will further distinguish between methods that assume the domain to be fixed and those that can adapt to different domains. 

In Section IX, we show examples of selected problems from the fields of network analysis, particle physics, recommender systems, computer vision, and graphics. In Section X, we draw conclusions and outline current main challenges and potential future research directions in geometric deep learning.   
To make the paper more readable, we use inserts to illustrate important concepts. 
Finally, the readers are invited to visit a dedicated website {\tt geometricdeeplearning.com} for additional materials, data, and examples of code. 

}

%In the next sections, we will start with reviewing the basic ideas of classical deep learning on Euclidean domains (images) and then see the differences and challenges introduced by the need to deal with a non-Euclidean domains. 
% 
%
%
\myframedtext[0.95\linewidth]{
%
%\paragraph*{\bf Notation}\\
\begin{tabular}{ l l }
{\bf Notation} &\\
  $\mathbb{R}^m$ & $m$-dimensional Euclidean space\\
  $a, \mathbf{a}, \mathbf{A}$ & Scalar, vector, matrix\\
   $\bar{a}$ & Complex conjugate of $a$\\ 
  $\Omega, x$ & Arbitrary domain, coordinate on it \\
  $f\in L^2(\Omega)$ & Square-integrable function on $\Omega$ \\
    $\delta_{x'}(x), \delta_{ij}$ & Delta function at $x'$, Kronecker delta\\
   $\{ f_i, y_i\}_{i\in \mathcal{I}}$ & Training set\\ 
  $\mathcal{T}_v$ & Translation operator\\
  $\tau, \mathcal{L}_\tau$ & Deformation field, operator\\
  $\hat{f}$ & Fourier transform of $f$ \\
  $f \star g$ & Convolution of $f$ and $g$ \\
  $\mathcal{X}, T\mathcal{X}, T_x \mathcal{X}$ & Manifold, its tangent bundle, tangent \\ & space at $x$\\  
  $\langle \cdot, \cdot, \rangle_{T\mathcal{X}}$ & Riemannian metric \\  
  $f\in L^2(\mathcal{X})$ & Scalar field on manifold $\mathcal{X}$ \\  
  $F\in L^2(T\mathcal{X})$ & Tangent vector field on manifold $\mathcal{X}$ \\  
  $A^*$ & Adjoint of operator $A$ \\  
  $\nabla, \mathrm{div}, \Delta$ & Gradient, divergence, Laplace operators \\  
  $\mathcal{V}, \mathcal{E}, \mathcal{F}$ & Vertices and edges of a graph, \\  
  & faces of a mesh \\  
  $w_{ij}, \mathbf{W}$ & Weight matrix of a graph, \\
  $f\in L^2(\mathcal{V})$ & Functions on vertices of a graph\\  
  $F\in L^2(\mathcal{E})$ & Functions on edges of a graph \\  
  $\phi_i, \lambda_i$ & Laplacian eigenfunctions, eigenvalues \\  
  $h_t(\cdot, \cdot)$ & Heat kernel \\  
  $\boldsymbol{\Phi}_k$ & Matrix of first $k$ Laplacian eigenvectors \\  
  $\boldsymbol{\Lambda}_k$ & Diagonal matrix of first $k$ Laplacian  \\  
  & eigenvalues\\
  $\xi$ & Point-wise nonlinearity (e.g. ReLU) \\
  $\w_{l,l'}(x), \mathbf{\W}_{l,l'}$ & Convolutional filter in spatial\\& and spectral domain\\
\end{tabular}
%
%We denote by $\Omega$ an arbitrary domain, by $x \in \Omega$ a coordinate in this domain, 
%and by $f(x)$ a function or image defined on that domain. 
%$f \in L^2(\Omega, \R^m)$ if $f(x) \in \R^m$ and
%it is square integrable: $\|f \|_2^2= \int \|f(x)\|^2 dx < \infty$, 
%and $f \in L^1(\Omega,\R^m)$ if it is absolutely integrable: $\|f \|_1 = \int |f(x)| dx < \infty$.
%For simplicity, we write $L^p(\Omega) = L^p(\Omega, \R)$.
%The Fourier transform of $f \in L^2(\Omega, \R)$ is denoted as 
%$\hat{f}(\omega) = \int f(x) e^{-i x.\omega } dx$.
%Given $f, g \in L^1(\Omega)$, the convolution is 
%$f \star g(x) = \int f(x') g(x-x') dx'$.
%Matrices are denoted in bold-face $\mathbf{X}$.
%[Introduce here the notation concerning the Graph Laplacian].
%
}

%At this point, we should introduce a more precise terminology, which we will define in the next sections. 
%By non-Euclidean domains, we refer to two prototypical structures: graphs and manifolds. The latter, being a continuous structure, can be discretized in many ways, also as a graph. The main characteristic of a manifold when compared to 

\section{Deep learning on Euclidean domains}
\label{sec:deepl}

%In this Section we review traditional Convolutional Neural Network architectures, 
%highlighting the role of specific mathematical properties that shall be generalized later on on non-Euclidean domains. 
%For a more in-depth review of CNNs and applications we refer the reader to \cite{lecun2015deep} and references therein.
%
%
%[PROTOTYPICAL APPLICATIONS: image classification employing pooling, and pixel-wise classification]

\paragraph*{\bf Geometric priors}
%define the problems at hand. 
Consider a compact $d$-dimensional Euclidean domain $\Omega = [0,1]^d \subset \mathbb{R}^d$ on which  
square-integrable functions $f\in L^2(\Omega)$ are defined (for example, in image analysis applications, images can be thought of as functions on the unit square $\Omega = [0,1]^2$). 
We consider a generic supervised learning setting, in which an unknown function 
$y : ~ L^2(\Omega) \to \mathcal{Y}$ is observed on a training set 
\begin{equation}
\{ f_i \in L^2(\Omega), y_i = y(f_i) \}_{i \in \mathcal{I}}. 
\end{equation} 
In a supervised {\em classification} setting, the target space $\mathcal{Y}$ can be thought discrete with $| \mathcal{Y} |$ being 
the number of classes. 
In a {\em multiple object recognition} setting, we can replace $\mathcal{Y}$ by the $K$-dimensional simplex, 
which represents the posterior class probabilities $p( y | x)$. 
In {\em regression} tasks, we may consider $\mathcal{Y}=\mathbb{R}^m$. 
%
%$y$ is often referred to as {\em label}. 

In the vast majority of computer vision and speech analysis tasks, there are several  crucial 
%prior information 
prior assumptions on the unknown function $y$. 
As we will see in the following, these assumptions are effectively exploited by convolutional neural network architectures. 
% that is effectively exploited by Convolutional Neural Network architectures. 
%We will now describe these priors mathematically with the objective of extending them to non-Euclidean 
%domains later on. 

{\bf{\em Stationarity:}} 
Let 
\begin{equation}{\cal T}_v f(x) = f(x - v), \hspace{3mm} x, v \in \Omega, 
\end{equation}
%$T_v$ for some $v \in \mathbb{R}^d$ 
be a {\em translation operator}\footnote{
We assume periodic boundary conditions to 
ensure that the operation is well-defined over $L^2(\Omega)$.
} acting on functions $f \in L^2(\Omega)$. %as 
%$T_v f(x) = f(x - v)$ for 
%
Our first assumption is that the function $y$ is either {\em invariant} or {\em equivariant} with 
respect to translations, depending on the task. 
In the former case, we have $y({\cal T}_v f) = y(  f) $ for any $f \in L^2(\Omega)$ and $v\in \Omega$. This
is typically the case in object classification tasks. In the latter, we have $y({\cal T}_v f) = {\cal T}_v y(f)$, which 
is well-defined when the output of the model is a space in which translations can act upon (for example, in problems of object localization, 
semantic segmentation, or motion estimation). \editJB{Our definition of invariance should
not be confused with the traditional notion of \emph{translation invariant systems} in signal processing, 
which corresponds to translation equivariance in our language (since the output translates whenever the input translates).}

% Equivalently, we can describe this property in terms of the statistics of natural images. 
% If one considers that natural images are drawn from an underlying probability distribution, 
% the invariance/equivariance property implies that this distribution describes a stationary source \cite{simoncelli2001natural} in the 

{\bf{\em Local deformations and scale separation}:} 
Similarly, a deformation ${\cal L}_\tau$, where $\tau: \Omega \to \Omega$ is a smooth vector field,  acts on $L^2(\Omega)$ as 
${\cal L}_\tau f(x) = f(x -\tau(x))$. Deformations can model local translations, changes in point of view, rotations and
frequency transpositions \cite{bruna2013invariant}. 

Most tasks studied in computer vision are not only translation invariant/equivariant, but also stable with respect to local deformations \cite{mallat2016understanding,bruna2013invariant}. In tasks that are translation invariant we have
\begin{equation}
\label{deformstab}
| y( {\cal L}_\tau f) - y( f) | \approx \| \nabla \tau \|,
\end{equation}
for all $f, \tau$. 
 Here, $\| \nabla \tau \|$ measures the smoothness of a given deformation field. In other words, 
 the quantity to be predicted does not change much if the input image is slightly deformed. 
 In tasks that are translation equivariant, we have 
 \begin{equation}
 | y( {\cal L}_\tau f) - {\cal L}_\tau y( f) | \approx \| \nabla \tau \|.
 \end{equation}
% instead.
This property is much stronger than the previous one, since the space of local deformations has a high dimensionality, 
as opposed to the $d$-dimensional translation group.

It follows from (\ref{deformstab}) that we can extract sufficient statistics at a lower spatial resolution by 
downsampling demodulated localized filter responses without losing approximation power. 
An important consequence of this is that long-range dependencies can be broken into multi-scale 
local interaction terms, leading to hierarchical models in which spatial resolution is progressively reduced.
 To illustrate this principle,
denote by 
 \begin{equation}
Y(x_1, x_2;v) = \mathrm{Prob}( f(u) = x_1 \,\,\, \mathrm{and} \,\,\, f(u+v)=x_2)
 \end{equation}
 the joint distribution of two image pixels at an offset $v$ from each other. 
 In the presence of long-range dependencies, this joint distribution will not be separable for any $v$. However, the deformation stability prior states that $Y(x_1, x_2; v) \approx Y(x_1, x_2; v(1+\epsilon))$ for small $\epsilon$. In other words, whereas long-range dependencies indeed exist in natural images and are critical to object recognition, they can be captured and down-sampled at different scales.
This principle of stability to local deformations has been exploited in the computer vision community 
in models other than CNNs, for instance, deformable parts models \cite{felzenszwalb2010object}. 
% are graphical models that incorporate 
%a similar prior in their model specification. 

In practice, the Euclidean domain $\Omega$ is discretized using a regular grid with $n$ points; the translation and deformation operators are still well-defined so the above properties hold in the discrete setting.

\myframedtext*[\linewidth]{
\vspace{-5mm}
\begin{multicols}{2}
\paragraph*{\bf [IN1] Convolutional neural networks}
CNNs are currently among the most successful deep learning architectures in a variety of tasks, in particular, in computer vision. A typical CNN used in computer vision applications (see FIGS1) consists of multiple convolutional layers~(\ref{eq:convlayer}), passing the input image through a set of filters $\W$ followed by point-wise non-linearity $\xi$ (typically, half-rectifiers $\xi(z) = \max(0, z)$ are used, although practitioners 
have experimented with a diverse range of choices \cite{Goodfellow-et-al-2016-Book}). 
The model can also include a bias 
term, which is equivalent to adding a constant coordinate to the input. \\% at each layer.
A network composed of $K$ convolutional layers put together $U(f) = (C_{\W^{(K)}} \hdots  \circ C_{\W^{(2)}} \circ  C_{\W^{(1)}})(f)$ produces pixel-wise features that are covariant w.r.t. translation and approximately covariant to local deformations. 
Typical computer vision applications requiring covariance are semantic image segmentation \cite{farabet2013learning} or motion estimation \cite{dosovitskiy2015flownet}. 
\\
In applications requiring invariance, such as image classification \cite{krizhevsky2012imagenet}, the convolutional layers are typically interleaved with pooling layers~(\ref{eq:poolinglayer}) progressively reducing the resolution of the image passing through the network. 
Alternatively, one can integrate the convolution and downsampling in a single linear operator (convolution with stride). 
Recently, some authors have also experimented with convolutional layers which 
increase the spatial resolution using interpolation kernels \cite{springenberg2014striving}. These kernels 
can be learnt efficiently by mimicking the so-called \emph{algorithme \`{a} trous} \cite{mallat1999wavelet}, also referred to as {\em dilated convolution}.
\end{multicols}
\begin{center}
\begin{overpic}
[width=1\linewidth]{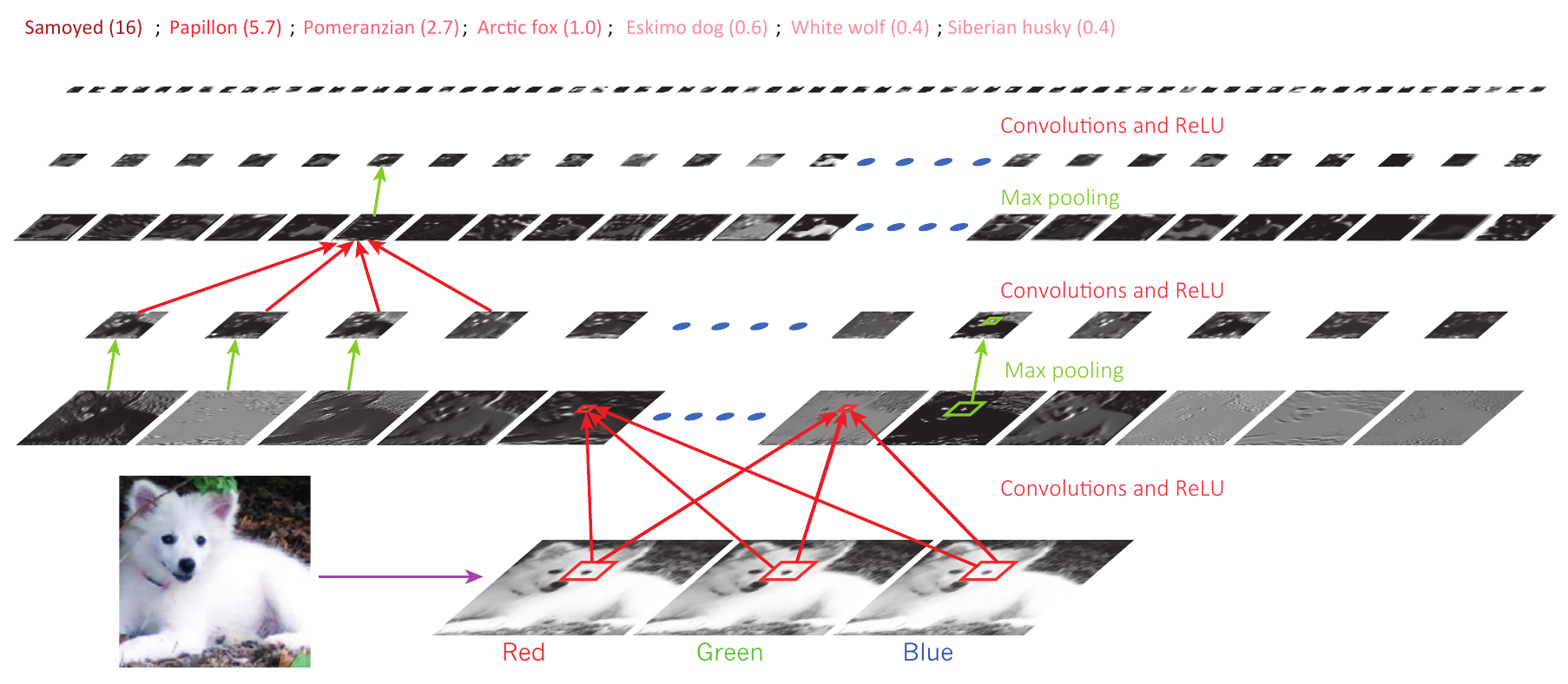} %cnn_typical}
\end{overpic}
\end{center}
{\small {\bf [FIGS1]} Typical convolutional neural network architecture used in computer vision applications  % [fig. credit J. Tompson]. 
(figure reproduced from \cite{lecun2015deep}).
}
%
%Third and fourth rows: examples of correspondence between different shapes affected by extreme transformations (missing parts) learned by intrinsic CNN . Corresponding points are encoded in similar color. Reference shape is shown on the left. 
%\caption{
%}
%\end{figure*}
}

\paragraph*{\bf Convolutional neural networks}

Stationarity and stability to local translations are both leveraged in 
convolutional neural networks (see insert IN1). % \cite{lecun1989backpropagation}. 
A CNN consists of several {\em convolutional layers} of the form $\mathbf{g} = C_\W(\mathbf{f})$, acting on a $p$-dimensional input $\mathbf{f}(x) = (f_1(x), \hdots, f_p(x))$ by applying a bank of filters $\W = (\w_{l,l'})$, $l = 1,\hdots, q, l' = 1,\hdots, p$ and point-wise non-linearity $\xi$, 
\begin{eqnarray}
\label{eq:convlayer}
 g_l(x) &=& \xi\left( \sum_{l'=1}^{p} (f_{l'} \star \w_{l,l'}) (x) \right), % \\ 
% &=&\xi\left(  \sum_{l'=1}^{m}\int  f_{l'}( x-x') w_{l,l'} (x')  dx' \right), \nonumber
%& =&  \sum_{l'=1}^{N_k} \sum_{\substack{ |v_x|\leq S_{k,x},\\ |v_y| \leq S_{k,y}} } x_k( u_x- v_x, u_y - v_y, l') W^{(k)}_{l,l'}(v_x, v_y) \nonumber ~,\\
%g_l(x) &=& \xi( \tilde{f}_l(x) )~,~(l \leq n)~.
\end{eqnarray}
producing a $q$-dimensional output $\mathbf{g}(x) = (g_1(x), \hdots, g_q(x))$ often referred to as the {\em feature maps}.  
%
%Let $f \in L^1(\Omega, \R^m)$ 
%and $W = ( W_{l,l'} \in L^1(\Omega) )_{l \leq m, l' \leq n}$.
%A convolutional layer is defined iteratively
%as a transformation of the form $g = C_W( f)$, with
%\begin{eqnarray*}
% \tilde{f}_l(x) &=& \sum_{l'=1}^{m} f_{l'} \star W_{l,l'} (x) \\ 
% &=&\sum_{l'=1}^{m}\int  f_{l'}( x-x') W_{l,l'} (x')  dx' \\
%%& =&  \sum_{l'=1}^{N_k} \sum_{\substack{ |v_x|\leq S_{k,x},\\ |v_y| \leq S_{k,y}} } x_k( u_x- v_x, u_y - v_y, l') W^{(k)}_{l,l'}(v_x, v_y) \nonumber ~,\\
%g_l(x) &=& \xi( \tilde{f}_l(x) )~,~(l \leq n)~.
%\end{eqnarray*}
%Here, $n$ is the number of filters or \emph{feature maps} created by the 
%convolutional tensor $W$,
%and $\xi(\cdot)$ is a point-wise nonlinearity. 
%
%A convolutional network is constructed by composing several 
%convolutional layers using $W^{(1)}, \hdots W^{(K)}$. 
%
Here, 
\begin{equation}
\label{eq:convdef}
(f \star \w)(x) = \int_\Omega  f( x-x') \w(x')  dx'
\end{equation}
denotes the standard convolution. 
According to the local deformation prior,
the filters $\W$ %at each layer $k$ 
have compact spatial support. 
% $(2 S_{k,x} +1 ) \times (2 S_{k,y}+1)$, and 
%the nonlinearities are typically 
%half-rectifiers $\xi(z) = \max(0, z)$, although practitioners 
%have experimented with a diverse range of choices \cite{Goodfellow-et-al-2016-Book}.
%The model can also include a bias 
%term, which is equivalent to adding a constant coordinate to the input at each layer.

Additionally, a downsampling or \emph{pooling} layer $\mathbf{g} = P(\mathbf{f})$ may be used, defined as 
\begin{equation}
\label{eq:poolinglayer}
g_l(x) = P( \{ f_l( x') \, : \, x' \in \mathcal{N}(x) \} ),  \,\,\, l = 1,\hdots, q, 
\end{equation}
where $\mathcal{N}(x)\subset\Omega$ is a neighborhood around $x$ 
and $P$ is a permutation-invariant function such as a $L_p$-norm  % $\mathcal{G}(y) = \|y \|_p$.
(in the latter case, the choice of $p=1,2$ or $\infty$ results in {\em average-}, {\em energy-}, or {\em max-pooling}). 
%Typical choices for $\mathcal{G}$ are $p=1$ (average-pooling), $p=2$ (energy-pooling) or $p=\infty$ (max-pooling).
%Alternatively, one can integrate the convolution and downsampling in a single linear operator (convolution with stride). 
%Recently, some authors have also experimented with convolutional layers which 
%increase the spatial resolution using interpolation kernels \cite{springenberg2014striving}. These kernels 
%can be learnt efficiently by mimicking the so-called \emph{algorithme \`{a} trous} \cite{mallat1999wavelet}, also referred as {\em dilated convolution}.

A convolutional network is constructed by composing several 
convolutional and optionally pooling layers, %layers using $W^{(1)}, \hdots W^{(K)}$. 
%By cascading convolutional and pooling layers, one obtains 
obtaining a generic hierarchical representation 
\begin{equation}
U_{\boldsymbol{\Theta}}(f) =  (C_{\W^{(K)}} \cdots P \cdots \circ C_{\W^{(2)}} \circ C_{\W^{(1)}} ) (f)
\end{equation}
where $\boldsymbol{\Theta} = \{\W^{(1)}, \hdots, \W^{(K)}\}$ is the hyper-vector of the network parameters (all the filter coefficients). 
The model is said to be {\em deep} if it comprises multiple layers, though this notion is rather vague and one can find examples of CNNs with as few as a couple and as many as hundreds of layers \cite{he2015deep}. 
The output features enjoy translation invariance/covariance depending on 
whether spatial resolution is progressively lost by means of pooling or kept fixed. 
 Moreover, 
if one specifies the convolutional tensors to be complex wavelet decomposition operators 
and uses complex modulus as point-wise nonlinearities, 
one can provably obtain stability to local deformations \cite{mallat2012group}. Although this 
stability is not rigorously proved for generic compactly supported convolutional tensors, 
it underpins the empirical success of CNN architectures 
across a variety of computer vision applications \cite{lecun2015deep}.

In supervised learning tasks, one can obtain the CNN parameters by 
minimizing a task-specific cost $L$ on the training set $\{ f_i, y_i\}_{i \in \mathcal{I}}$, 
%solving an 
%end-to-end optimization task
\begin{equation}
\min_{\boldsymbol{\Theta} } \,\,\, \sum_{i \in \mathcal{I}} L( U_{\boldsymbol{\Theta}}(f_i), y_i), 
\label{eq:cnnmin}
\end{equation}
%
%where $\boldsymbol{\Theta}$ is the hyper-vector of the network parameters (all the filter coefficients). 
%This is a process referred to as {\em learning}. 
%
for instance, $L(x,y) = \|x-y\|$. 
If the model is sufficiently complex and the training set is sufficiently representative, when applying the learned model to previously unseen data, one expects $U(f) \approx y(f)$.  
Although~(\ref{eq:cnnmin}) is a non-convex optimization problem, stochastic optimization 
methods offer excellent empirical performance. 
Understanding the structure of the optimization problems~(\ref{eq:cnnmin}) and finding efficient strategies for its solution is an active area of research in deep learning 
 \cite{choromanska2015loss, safran2015quality, kawaguchi2016deep, chen2015net2net, freeman2016topology}.

A key advantage of CNNs explaining their success in numerous tasks is that the geometric priors on which CNNs are based result in a learning complexity 
that avoids the curse of dimensionality. Thanks to the stationarity and local 
deformation priors, the 
linear operators at each layer have a constant number of parameters, independent 
of the input size $n$ (number of pixels in an image). 
\editJB{Moreover, thanks to the multiscale hierarchical property, 
the number of layers grows at a rate $\mathcal{O}(\log n)$, resulting in a total learning 
complexity of $\mathcal{O}(\log n)$ parameters. 
}

%\paragraph*{\bf Application examples of CNNs}

\section{The geometry of manifolds and graphs}
\label{sec:noneucl}

Our main goal is to generalize CNN-type constructions to non-Euclidean domains. 
In this paper, by non-Euclidean domains, we refer to two prototypical structures: {\em manifolds} and {\em graphs}. 
While arising in very different fields of mathematics (differential geometry and graph theory, respectively), in our context, these structures share several common characteristics that we will try to emphasize throughout our review. 
%
%Being aware that these subjects are typically not studied in standard signal processing curricular, we nevertheless believe that .

%\paragraph*{\bf Graphs}

%The former, being a continuous structure, can be discretized in many ways, also as a graph. The main characteristic of a manifold is that it can be locally associated with a low-dimensional vector space. 

\paragraph*{\bf Manifolds}

Roughly, a manifold is a space that is locally Euclidean. One of the simplest examples is a spherical surface modeling our planet: around a point, it seems to be planar, which has led generations of people to believe in the flatness of the Earth. 
Formally speaking, a (differentiable) $d$-{\em dimensional manifold} $\mathcal{X}$ is a topological space where each point $x$ has a neighborhood that is topologically equivalent (homeomorphic) to a $d$-dimensional Euclidean space, called the {\em tangent space} and denoted by $T_x\mathcal{X}$ (see Figure~1, top). %\ref{fig:isometries}, top)
%\footnote{We warn readers that the ``manifolds'' that occur in many data analysis applications can have less structure then a classical smooth manifold.   For example, a set of points that ``looks locally Euclidean'' in practice may have self intersections,  high curvature, different dimensions depending on the scale and location at which you look,  extreme variations in density,  and ``noise'' with confounding structure.  Nevertheless, as we discuss below, many of the tools for analyzing classical manifolds extend to these cases.}. 
%
The collection of tangent spaces at all points (more formally, their disjoint union) is referred to as the {\em tangent bundle} and denoted by $T\mathcal{X}$. 
On each tangent space, we define an inner product $\langle \cdot, \cdot \rangle_{T_x\mathcal{X}} : T_x\mathcal{X} \times T_x\mathcal{X} \rightarrow \mathbb{R}$, which is additionally assumed to depend smoothly on the position $x$. This inner product is called a {\em Riemannian metric} in differential geometry and allows performing local measurements of angles, distances, and volumes. A manifold equipped with a metric is called a {\em Riemannian manifold}. %, after Bernhard Riemann who first formalized these notions.

It is important to note that the definition of a Riemannian manifold is completely abstract and does not require a geometric realization in any space. However, a Riemannian manifold can be realized as a subset of a Euclidean space (in which case it is said to be {\em embedded} in that space) by using the structure of the Euclidean space to induce a Riemannian metric. The celebrated {\em Nash Embedding Theorem} guarantees that any sufficiently smooth Riemannian manifold can be realized in a Euclidean space of sufficiently high dimension \cite{nash1956imbedding}. 
An embedding is not necessarily unique; two different realizations of a Riemannian metric are called {\em isometries}. 

Two-dimensional manifolds (surfaces) embedded into $\mathbb{R}^3$ are used in computer graphics and vision to describe boundary surfaces of 3D objects, colloquially referred to as `3D shapes'. This term is somewhat misleading since `3D' here refers to the dimensionality of the embedding space rather than that of the manifold.  
Thinking of such a shape as made of infinitely thin material, inelastic deformations that do not stretch or tear it are isometric. Isometries do not affect the metric structure of the manifold and consequently, preserve any quantities that can be expressed in terms of the Riemannian metric (called {\em intrinsic}). Conversely, properties related to the specific realization of the manifold in the Euclidean space are called {\em extrinsic}.

As an intuitive illustration of this difference, imagine an insect that lives on a two-dimensional surface (Figure~1, bottom). %\ref{fig:isometries}, bottom). The surface can be placed in the Euclidean space in any way, but as long as it is transformed isometrically, the insect would not notice any difference. The insect in fact does not even know of the existence of the embedding space, as its only world is 2D. This is an intrinsic viewpoint. 
A human observer, on the other hand, sees a surface in 3D space - this is an extrinsic point of view.

%intrinsic vs extrinsic

\begin{figure}
\begin{overpic}
[width=1\linewidth]{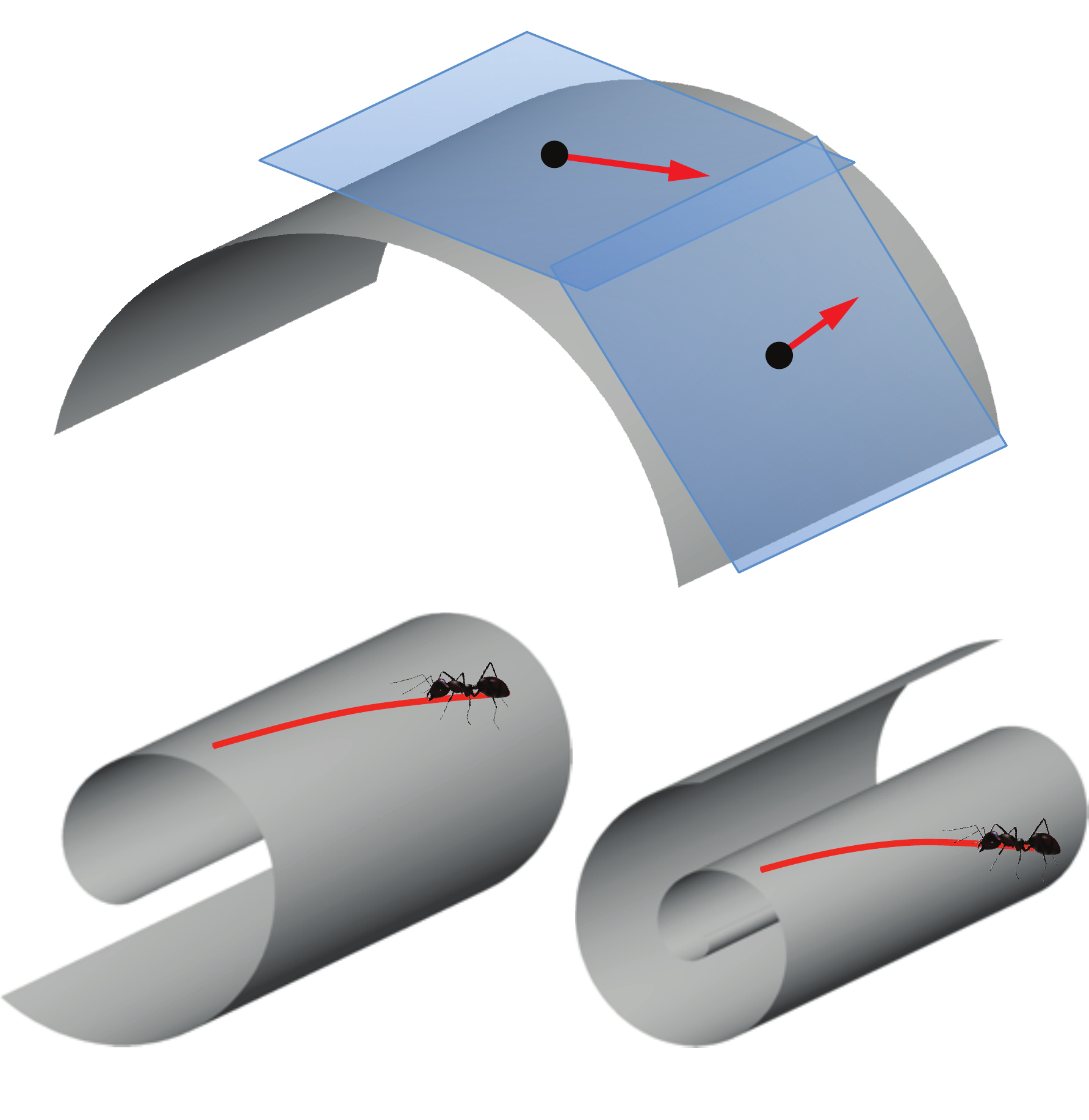}
\put(28.5,84.5){\small $T_x \mathcal{X}$}
\put(48.35,82.5){\small $x$}
\put(56.5,87.25){\small $\color{red}{F(x)}$}
\put(65,53){\small $T_{x'} \mathcal{X}$}
\put(68.35,64){\small $x'$}
\put(70,75){\small $\color{red}{F(x')}$}
\end{overpic}\vspace{-7mm}
\label{fig:isometries}
\caption{Top: tangent space and tangent vectors on a two-dimensional manifold (surface). 
Bottom: Examples of isometric deformations. }
\end{figure}

\myframedtext*[\linewidth]{
\vspace{-5mm}
\begin{multicols}{2}
\vspace{5mm}
\begin{center}
  \begin{overpic}
[width=0.47\columnwidth]{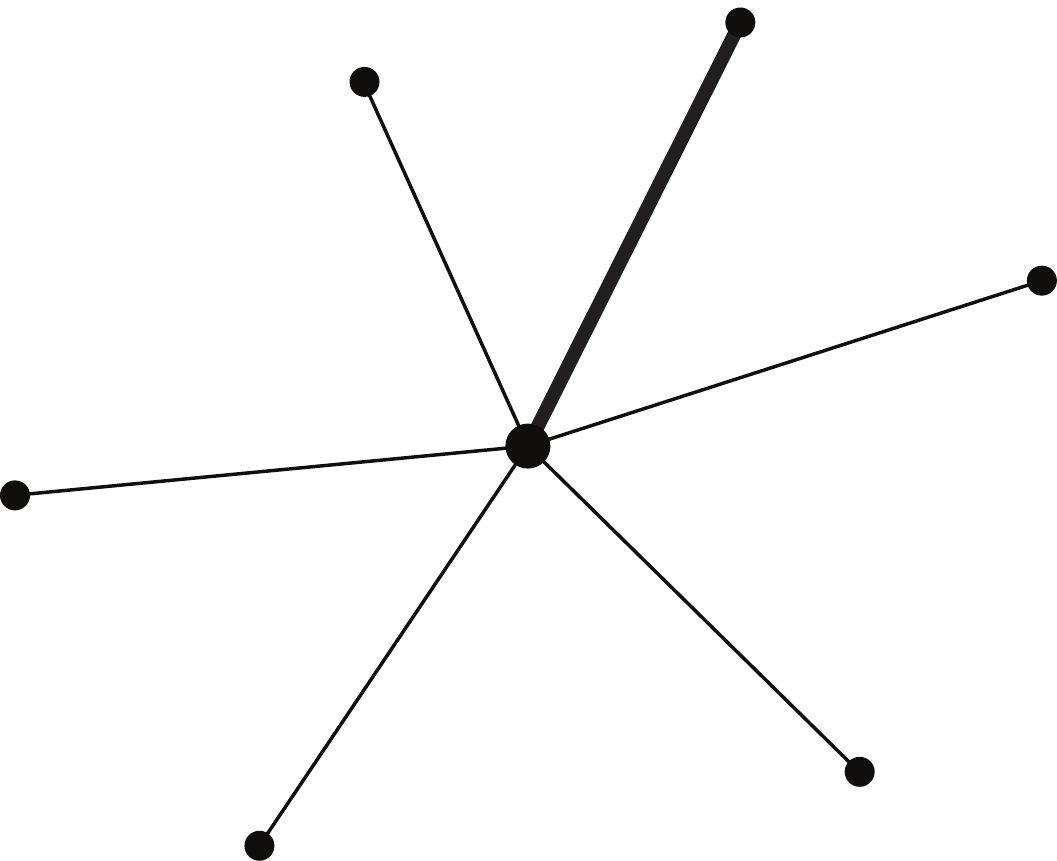}
	\put(69,83.5){\small $j$} 		
	\put(48,32){\small $i$} 		
	\put(64,60){\small $w_{ij}$} 	
	\put(31,-12){\small Undirected graph} 	
	\end{overpic}
	\hspace{1mm}
\begin{overpic}
[width=0.47\columnwidth]{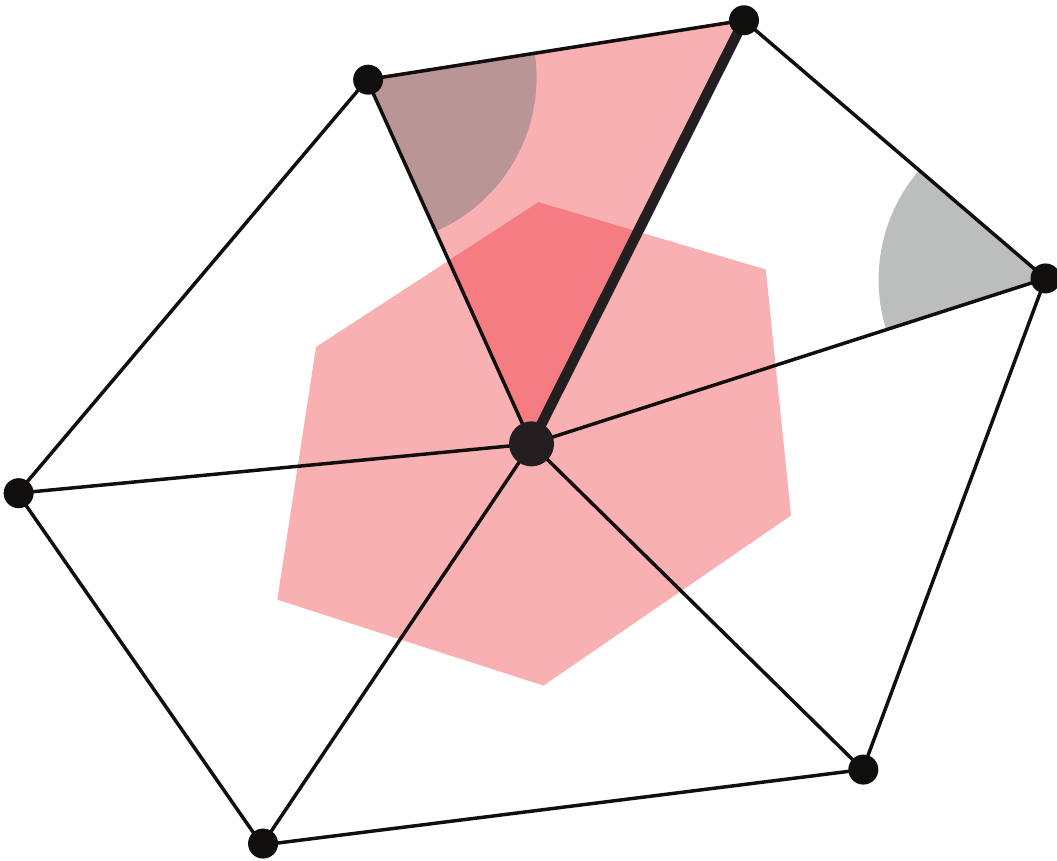}
	\put(69,83.5){\small $j$} 		
	\put(48,32){\small $i$} 		
	\put(28,72.5){\small$k$} 	
	\put(101,53){\small$h$} 	
	\put(38,69){\small$\alpha_{ij}$} 	
	\put(85,56){\small$\beta_{ij}$} 	
	\put(35.5,44.5){\color{red}\small$a_i$} 	
	\put(50,67){\color{red}\small$a_{ijk}$} 	
	\put(65,60){\small$\ell_{ij}$} 	
	\put(31,-12){\small Triangular mesh} 			
	\end{overpic}
	\vspace{2mm}
\end{center}
{\small {\bf [FIGS2]} Two commonly used discretizations of a two-dimensional manifold: a graph and a triangular mesh.}
\vspace{3mm}
\paragraph*{\bf [IN2] Laplacian on discrete manifolds}
%The Laplacian of a graph or a manifold discretized as a mesh with $n$ points can be written in the form 
%\begin{eqnarray}
%(\Delta f)_{i} &=& \frac{1}{a_i} \sum_{(i,j) \in E} w_{ij}(f_i - f_j)
%\label{eq:dlaplacian}
%\end{eqnarray}
%%
%where $E$ denotes the edge set of a graph or a mesh, and $w_{ij}$ and $a_i$ are the edge- and vertex-weights, respectively. 
%%
%Denoting by $\mathbf{W}$ the $n\times n$ matrix of edge weights and $\mathbf{A} = \mathrm{diag}(a_1, \hdots, a_n)$, the Laplacian application to a function $\mathbf{f} = (f_1, \hdots, f_n)^\top$ can be written in matrix-vector form as 
%%
%\begin{eqnarray}
%\boldsymbol{\Delta}\mathbf{f} &=& \mathbf{A}^{-1}\mathbf{W} \mathbf{f}.
%\end{eqnarray}
%%
%%
%Different discretizations of the Laplacian amount to difference choices of the edge and vertex weights in~(\ref{eq:dlaplacian}). 
%For undirected weighted graphs, it is typical to use the given edge weights $w_{ij}$ and set $a_i = \sum_{ij: (i,j) \in E} w_{ij}$. \\
%%
%%
In computer graphics and vision applications, two-dimensional manifolds are commonly used to model 3D shapes. %boundary surfaces of 3D objects. 
%, e.g. acquired with range sensors.\footnote{The reader might find it a bit confusing to refer to 3D shapes as 2D-manifolds. In fact, these are 2D-manifolds embedded in 3D Euclidean space. }
%
There are several common ways of discretizing such manifolds. First, the manifold is assumed to be sampled at $n$ points. Their embedding coordinates $\mathbf{x}_1, \hdots, \mathbf{x}_n$ are referred to as {\em point cloud}. Second, a graph is constructed upon these points, acting as its vertices. The edges of the graph represent the local connectivity of the manifold, telling whether two points belong to a neighborhood or not, e.g. with Gaussian edge weights 
\begin{eqnarray}
w_{ij} =  e^{-\| \mathbf{x}_i - \mathbf{x_j}\|^2 /2\sigma^2}.
\label{eq:graph_lap}
\end{eqnarray}
This simplest discretization, however, does not capture correctly the geometry of the underlying continuous manifold (for example, the graph Laplacian would typically not converge to the continuous Laplacian operator of the manifold with the increase of the sampling density \cite{wardetzky2007discrete}). 
A geometrically consistent discretization is possible with an additional structure of {\em faces} $\mathcal{F} \in \mathcal{V}\times \mathcal{V}\times \mathcal{V}$, where $(i,j,k) \in \mathcal{F}$ implies $(i,j), (i,k), (k,j) \in \mathcal{E}$. The collection of faces represents the underlying continuous manifold as a polyhedral surface consisting of small triangles glued together. The triplet $(\mathcal{V}, \mathcal{E}, \mathcal{F})$ is referred to as {\em triangular mesh}. 
To be a correct discretization of a manifold (a {\em manifold mesh}), every edge must be shared by exactly two triangular faces; if the manifold has a boundary, any boundary edge must belong to exactly one triangle. \\
On a triangular mesh, the simplest discretization of the Riemannian metric is given by assigning each edge a length $\ell_{ij} > 0$, which must additionally satisfy the triangle inequality in every triangular face. The mesh Laplacian is given by formula~(\ref{eq:dlaplacian}) with  
\begin{eqnarray}
\label{eq:wcotan}
w_{ij} &=&\dfrac{-\ell_{ij}^2+\ell_{jk}^2+\ell_{ik}^2}{8 a_{ijk}} + \dfrac{-\ell_{ij}^2+\ell_{jh}^2+\ell_{ih}^2}{8 a_{ijh}}; \\
\label{eq:acotan}
a_i &=& \tfrac{1}{3} \sum_{jk : (i,j,k) \in \mathcal{F}} a_{ijk},
\end{eqnarray}
where $a_{ijk} = \sqrt{s_{ijk}(s_{ijk}-\ell_{ij})(s_{ijk}-\ell_{jk})(s_{ijk}-\ell_{ik})}$ is the area of triangle $ijk$ given by the Heron formula, and $s_{ijk} = \tfrac{1}{2}(\ell_{ij} + \ell_{jk} + \ell_{ki})$ is the semi-perimeter of triangle $ijk$. 
The vertex weight $a_i$ is interpreted as the local area element (shown in red in FIGS2). Note that the weights (\ref{eq:wcotan}-\ref{eq:acotan}) are expressed solely in terms of the discrete metric $\ell$ and are thus intrinsic. When the mesh is infinitely refined under some technical conditions, such a construction can be shown to converge to the continuous Laplacian of the underlying manifold \cite{wardetzky2008convergence}. \\
An embedding of the mesh (amounting to specifying the vertex coordinates $\mathbf{x}_1, \hdots, \mathbf{x}_n$) induces a discrete metric $\ell_{ij} = \| \mathbf{x}_i - \mathbf{x}_j \|_2$, whereby~(\ref{eq:wcotan}) become the {\em cotangent weights} 
\begin{eqnarray}
w_{ij} =  \tfrac{1}{2}\left( \cot \alpha_{ij} + \cot \beta_{ij}\right)
\label{eq:cotan}
\end{eqnarray}
ubiquitously used in computer graphics \cite{pinkall1993computing}.
%
%\vspace{7mm}
%
\end{multicols}
}

\paragraph*{\bf Calculus on manifolds}
%We are interested in studying functions on manifolds. 
Our next step is to consider functions defined on manifolds. We are particularly interested in two types of functions: 
A {\em scalar field} is a smooth real function $f: \mathcal{X} \rightarrow \mathbb{R}$ on the manifold. 
A {\em tangent vector field} $F : \mathcal{X} \rightarrow T\mathcal{X}$ is a mapping attaching a tangent vector $F(x) \in T_x \mathcal{X}$ to each point $x$.  %(or, in differential geometric parlance, a 1-form). 
As we will see in the following, tangent vector fields are used to formalize the notion of infinitesimal displacements on the manifold. 
%
%The term `smooth' here implies that the field does not change abruptly when moving from point $x$ to a nearby point $\xi$. 
%
We define the Hilbert spaces of scalar and vector fields on manifolds, denoted by $L^2(\mathcal{X})$ and $L^2(T\mathcal{X})$, respectively, with the following inner products: 
\begin{eqnarray}
\label{eq:inner1}
\langle f, g \rangle_{L^2(\mathcal{X})} &=& \int_{\mathcal{X}} f(x)g(x) dx ; \\
\langle F, G \rangle_{L^2(T\mathcal{X})} &=& \int_{\mathcal{X}} \langle F(x), G(x) \rangle_{T_x \mathcal{X}} dx;   
\label{eq:inner2}
\end{eqnarray}
$dx$ denotes here a $d$-dimensional volume element induced by the Riemannian metric. 

In calculus, the notion of derivative describes how the value of a function changes with an infinitesimal change of its argument. 
%\begin{eqnarray}
%df(x) &=& f(x+dx) - f(x) = f'(x) dx
%\end{eqnarray}
%
One of the big differences distinguishing classical calculus from differential geometry is a lack of vector space structure on the manifold, prohibiting us from na{\"i}vely using expressions like $f(x+dx)$. The conceptual leap that is required to generalize such notions to manifolds is the need to work locally in the tangent space. 

To this end, we define the {\em differential} of $f$ as an operator $df : T\mathcal{X} \rightarrow \mathbb{R}$ acting on tangent vector fields. 
%
%} $\mathbf{V} : \mathcal{X} \rightarrow T\mathcal{X}$. 
%
At each point $x$, the differential can be defined as a linear functional ($1$-form) $df(x) = \langle \nabla f(x), \, \cdot \, \rangle_{T_x \mathcal{X}}$ acting on tangent vectors $F(x) \in T_x \mathcal{X}$, which model a small displacement around $x$. %\footnote{\editMB{Considering $df(x)$ as a dual vector allows to think of $\nabla f(x)$ as a .  }}
% is modeled as a tangent vector $F(x) \in T_x \mathcal{X}$; 
The change of the function value as the result of this displacement is given by applying the form to the tangent vector, $df(x) F(x) = \langle \nabla f(x), F(x) \rangle_{T_x \mathcal{X}}$, and can be thought of as an extension of the notion of the classical directional derivative.

The operator $ \nabla f : L^2(\mathcal{X}) \rightarrow L^2(T\mathcal{X})$ in the definition above is called the {\em intrinsic gradient}, and is similar to the classical notion of the gradient defining the direction of the steepest change of the function at a point, with the only difference that the direction is now a tangent vector. 
Similarly, the {\em intrinsic divergence} is an operator $ \mathrm{div} : L^2(T\mathcal{X}) \rightarrow L^2(\mathcal{X})$ acting on tangent vector fields and (formal) adjoint to the gradient operator \cite{rosenberg1997laplacian}, 
\begin{eqnarray}
\langle F, \nabla f \rangle_{L^2(T\mathcal{X})} =  \langle \nabla^* F, f \rangle_{L^2(\mathcal{X})} = \langle -\mathrm{div} F, f \rangle_{L^2(\mathcal{X})}.
\label{eq:adjoint}
\end{eqnarray} 
Physically, a tangent vector field can be thought of as a flow of material on a manifold. The divergence measures the net flow of a field at a point, allowing to distinguish between field `sources' and `sinks'. 
Finally, the {\em Laplacian} (or {\em Laplace-Beltrami operator} in differential geometric jargon) $\Delta : L^2(\mathcal{X}) \rightarrow L^2(\mathcal{X})$ is an operator  
\begin{eqnarray}
\Delta f = -\mathrm{div} (\nabla f) 
\label{eq:lbo}
\end{eqnarray} 
acting on scalar fields. Employing relation~(\ref{eq:adjoint}), it is easy to see that the Laplacian is self-adjoint (symmetric), 
\begin{eqnarray}
\langle \nabla f, \nabla f \rangle_{L^2(T\mathcal{X})} =  \langle \Delta f, f \rangle_{L^2(\mathcal{X})} = \langle f, \Delta f \rangle_{L^2(\mathcal{X})}.
\label{eq:adjoint1}
\end{eqnarray} 
The lhs in equation~(\ref{eq:adjoint1}) is known as the {\em Dirichlet energy} in physics and measures the smoothness of a scalar field on the manifold (see insert IN3). 
The Laplacian can be interpreted as the difference between the average of a function on an infinitesimal sphere around a point and the value of the function at the point itself. It is one of the most important operators in mathematical physics, used to describe phenomena as diverse as heat diffusion (see insert IN4), quantum mechanics, and wave propagation. 
As we will see in the following, the Laplacian plays a center role in signal processing and learning on non-Euclidean domains, as its eigenfunctions generalize the classical Fourier bases, allowing to perform spectral analysis on manifolds and graphs.

It is important to note that all the above definitions are {\em coordinate free}. By defining a basis in the tangent space, it is possible to express tangent vectors as $d$-dimensional vectors and the Riemannian metric as a $d\times d$ symmetric positive-definite matrix. 

\paragraph*{\bf Graphs and discrete differential operators}
Another type of constructions we are interested in are graphs, which are popular models of networks, interactions, and similarities between different objects. For simplicity, we will consider {\em weighted undirected graphs}, formally defined as a pair $(\mathcal{V},\mathcal{E})$, where $\mathcal{V} = \{1, \hdots, n\}$ is the  set of $n$ vertices, 
 and $\mathcal{E} \subseteq \mathcal{V} \times \mathcal{V}$ is the set of edges, where the graph being undirected implies that $(i,j) \in \mathcal{E}$ iff $(j,i) \in \mathcal{E}$. 
 Furthermore, we associate a weight $a_i > 0$ with each vertex $i \in \mathcal{V}$, and a weight $w_{ij} \geq 0$ with each edge $(i,j) \in \mathcal{E}$.

Real functions $f : \mathcal{V} \rightarrow \mathbb{R}$ and $F : \mathcal{E} \rightarrow \mathbb{R}$ on the vertices and edges of the graph, respectively, are roughly the discrete analogy of continuous scalar and tangent vector fields %(or 0- and 1-forms) 
in differential geometry.\footnote{It is tacitly assumed here that $F$ is {\em alternating}, i.e., $F_{ij} = - F_{ji}$.} 
We can define Hilbert spaces $L^2(\mathcal{V})$ and $L^2(\mathcal{E})$ of such functions by specifying the respective inner products, 
\begin{eqnarray}
\label{eq:dinner1}
\langle f, g \rangle_{L^2(\mathcal{V})} &=& \sum_{i\in \mathcal{V}} a_i f_i g_i ; \\
\langle F, G \rangle_{L^2(\mathcal{E})} &=& \sum_{i\in \mathcal{E}} w_{ij} F_{ij} G_{ij}.   
\label{eq:dinner2}
\end{eqnarray}
%
%for all $a,b \in $. 

Let $f \in L^2(\mathcal{V})$ and $F \in L^2(\mathcal{E})$ be functions on the vertices and edges of the graphs, respectively. 
We can define differential operators acting on such functions analogously to differential operators on manifolds \cite{lim2015hodge}. 
The {\em graph gradient} is an operator $\nabla: L^2(\mathcal{V}) \rightarrow L^2(\mathcal{E})$ mapping functions defined on vertices to functions defined on edges, 
\begin{eqnarray}
(\nabla f)_{ij} &=& f_i - f_j, 
\label{eq:dgradient}
\end{eqnarray}
automatically satisfying $(\nabla f)_{ij} = - (\nabla f)_{ji}$. 
The {\em graph divergence} is an operator $\mathrm{div}: L^2(\mathcal{E}) \rightarrow L^2(\mathcal{V})$ doing the converse, 
\begin{eqnarray}
(\mathrm{div} F)_{i} &=& \frac{1}{a_i} \sum_{j: (i,j) \in \mathcal{E}} w_{ij} F_{ij}. 
\label{eq:ddiv}
\end{eqnarray}
It is easy to verify that the two operators are adjoint w.r.t. the inner products (\ref{eq:dinner1})--(\ref{eq:dinner2}), 
\begin{eqnarray}
\langle F, \nabla f \rangle_{L^2(\mathcal{E})} = \langle \nabla^* F, f \rangle_{L^2(\mathcal{V})} = \langle -\mathrm{div} F, f \rangle_{L^2(\mathcal{V})}.
\label{eq:dadjoint}
\end{eqnarray}

The {\em graph Laplacian} is an operator $\Delta: L^2(\mathcal{V}) \rightarrow L^2(\mathcal{V})$ 
%of a graph or a manifold discretized as a mesh with $n$ points can be written in the form 
defined as $\Delta = - \mathrm{div} \, \nabla$. Combining definitions~(\ref{eq:dgradient})--(\ref{eq:ddiv}), it can be expressed in the familiar form   
\begin{eqnarray}
(\Delta f)_{i} &=& \frac{1}{a_i} \sum_{(i,j) \in \mathcal{E}} w_{ij}(f_i - f_j). 
\label{eq:dlaplacian}
\end{eqnarray}
Note that formula~(\ref{eq:dlaplacian}) captures the intuitive geometric interpretation of the Laplacian as the difference between the local average of a function around a point and the value of the function at the point itself.  
%where $E$ denotes the edge set of a graph or a mesh, and $w_{ij}$ and $a_i$ are the edge- and vertex-weights, respectively. 
%

Denoting by $\mathbf{W} = (w_{ij})$ the $n\times n$ matrix of edge weights (it is assumed that $w_{ij} = 0$ if $(i,j) \notin \mathcal{E}$), by $\mathbf{A} = \mathrm{diag}(a_1, \hdots, a_n)$ the diagonal matrix of vertex weights, 
and by $\mathbf{D} = \mathrm{diag}(\sum_{j: j \neq i} w_{ij})$ the {\em degree matrix}, 
the graph Laplacian application to a function $f\in L^2(\mathcal{V})$ represented as a column vector $\mathbf{f} = (f_1, \hdots, f_n)^\top$ can be written in matrix-vector form as 
\begin{eqnarray}
\label{eq:grlaplacian}
\boldsymbol{\Delta}\mathbf{f} &=& \mathbf{A}^{-1}(\mathbf{D} - \mathbf{W}) \mathbf{f}.
\end{eqnarray}
The choice of $\mathbf{A} = \mathbf{I}$ in~(\ref{eq:grlaplacian}) is referred to as the {\em unnormalized graph Laplacian}; another popular choice is $\mathbf{A} = \mathbf{D}$ producing the {\em random walk Laplacian} \cite{von2007tutorial}.

\paragraph*{\bf Discrete manifolds} 
As we mentioned, there are many practical situations in which one is given a sampling of points arising from a manifold but not the manifold itself. In computer graphics applications, reconstructing a correct discretization of a manifold from a point cloud is a difficult problem of its own, referred to a {\em meshing} (see insert IN2). 
In manifold learning problems, the manifold is typically approximated as a graph capturing the local affinity structure. 
We warn the reader that the term ``manifold'' as used in the context of generic data science is not geometrically rigorous, and can have less structure than a classical smooth manifold we have defined beforehand.   For example, a set of points that ``looks locally Euclidean'' in practice may have self intersections, infinite curvature, different dimensions depending on the scale and location at which one looks,  extreme variations in density,  and ``noise'' with confounding structure.  
%Nevertheless, as we discuss below, many of the tools for analyzing classical manifolds extend to these cases.

\myframedtext*[\linewidth]{
\vspace{-5mm}
\begin{multicols}{2}
\paragraph*{\bf [IN3] Physical interpretation of Laplacian eigenfunctions}
Given a function $f$ on the domain $\mathcal{X}$, the {\em Dirichlet energy}
\begin{eqnarray}
\label{eq:dirichlet}
\mathcal{E}_\mathrm{Dir}(f) &=& \int_{\mathcal{X}} \| \nabla f(x) \|^2_{T_x \mathcal{X}} dx =  \int_{\mathcal{X}} f(x) \Delta f(x) dx,
%\mathcal{E}_\mathrm{dir}(f) &=& \int_{-\pi}^{+\pi} \| \nabla f(x) \| dx = \int_{-\pi}^{+\pi} \langle \nabla f(x) , \nabla f(x) \rangle dx  \nonumber \\
%&=& \int_{-\pi}^{+\pi} f(x) \Delta f(x) dx,
\end{eqnarray}
measures how smooth it is (the last identity in~(\ref{eq:dirichlet}) stems from~(\ref{eq:adjoint1})). We are looking for an orthonormal basis on $\mathcal{X}$, containing $k$ smoothest possible functions (FIGS3), by solving the optimization problem 
\begin{eqnarray}
\label{eq:courant}
\min_{\phi_0}  \,\, \mathcal{E}_\mathrm{Dir}(\phi_0)  & \mathrm{s.t.} & \| \phi_0 \| = 1 \\
\min_{\phi_i} \,\, \mathcal{E}_\mathrm{Dir}(\phi_i)  & \mathrm{s.t.} & \| \phi_i \| = 1,  \hspace{3mm} i = 1,2,\hdots k-1\nonumber\\
&& \phi_i \perp  \mathrm{span}\{\phi_0, \hdots, \phi_{i-1}\}. \nonumber 
%\mathcal{E}_\mathrm{dir}(f) &=& \int_{-\pi}^{+\pi} \| \nabla f(x) \| dx = \int_{-\pi}^{+\pi} \langle \nabla f(x) , \nabla f(x) \rangle dx  \nonumber \\
%&=& \int_{-\pi}^{+\pi} f(x) \Delta f(x) dx,
\end{eqnarray}
In the discrete setting, when the domain is sampled at $n$ points, problem~(\ref{eq:courant}) can be rewritten as 
\begin{eqnarray}
\label{eq:eigen}
\min_{\boldsymbol{\Phi}_k\in \mathbb{R}^{n\times k}}  \,\,\mathrm{trace}(\boldsymbol{\Phi}_k^\top \boldsymbol{\Delta} \boldsymbol{\Phi}_k)  & \mathrm{s.t.} & \boldsymbol{\Phi}_k^\top \boldsymbol{\Phi}_k = \mathbf{I},
\end{eqnarray}
where $\boldsymbol{\Phi}_k = (\boldsymbol{\phi}_0, \hdots \boldsymbol{\phi}_{k-1})$. 
The solution of~(\ref{eq:eigen}) is given by the first $k$ eigenvectors of $\boldsymbol{\Delta}$ satisfying 
\begin{eqnarray}
\boldsymbol{\Delta}\boldsymbol{\Phi}_k = \boldsymbol{\Phi}_k\boldsymbol{\Lambda}_k,
\label{eq:lapeigs}
\end{eqnarray}
where $\boldsymbol{\Lambda}_k = \mathrm{diag}(\lambda_0, \hdots, \lambda_{k-1})$ is the diagonal matrix of corresponding eigenvalues.  
The eigenvalues $0=\lambda_0 \leq \lambda_1 \leq \hdots \lambda_{k-1}$ are non-negative due to the positive-semidefiniteness of the Laplacian and can be interpreted as `frequencies', where $\boldsymbol{\phi}_0 = \mathrm{const}$ with the corresponding eigenvalue $\lambda_0 = 0$  play the role of the DC. \\
The Laplacian eigendecomposition can be carried out in two ways. First, equation~(\ref{eq:lapeigs}) can be rewritten as a generalized eigenproblem $(\mathbf{D} - \mathbf{W})\boldsymbol{\Phi}_k = \mathbf{A}\boldsymbol{\Phi}_k\boldsymbol{\Lambda}_k$, resulting in $\mathbf{A}$-orthogonal eigenvectors, $\boldsymbol{\Phi}_k^\top \mathbf{A} \boldsymbol{\Phi}_k = \mathbf{I}$. 
Alternatively, introducing a change of variables $\boldsymbol{\Psi}_k = \mathbf{A}^{1/2}\boldsymbol{\Phi}_k$, we can obtain a standard eigendecomposition problem $\mathbf{A}^{-1/2}(\mathbf{D} - \mathbf{W}) \mathbf{A}^{-1/2} \boldsymbol{\Psi}_k = \boldsymbol{\Psi}_k\boldsymbol{\Lambda}_k$ with orthogonal eigenvectors $\boldsymbol{\Psi}_k^\top \boldsymbol{\Psi}_k = \mathbf{I}$. When  $\mathbf{A} = \mathbf{D}$ is used, the matrix ${\boldsymbol{\Delta}} = \mathbf{A}^{-1/2}(\mathbf{D} - \mathbf{W}) \mathbf{A}^{-1/2}$ is referred to as the {\em normalized symmetric Laplacian}. 
\end{multicols}
\vspace{8mm}
\begin{center}
\begin{overpic}
[width=1\linewidth]{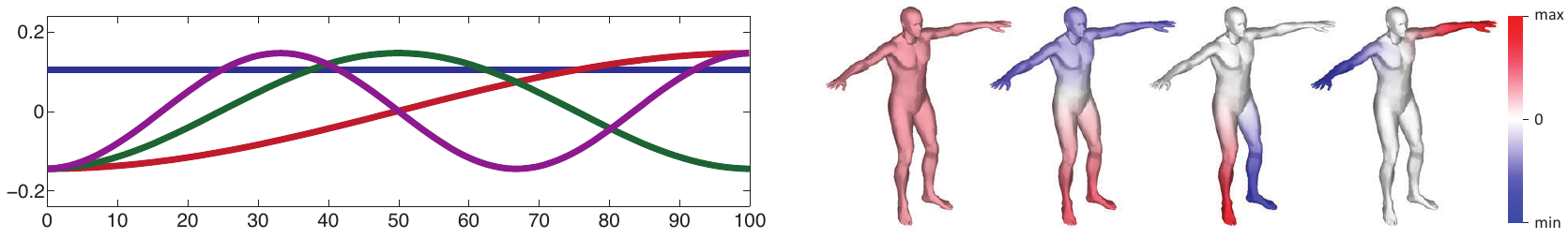}
\put(20,-2){\small Euclidean}
\put(72,-2){\small Manifold}
\put(4.5,12){\small $\phi_0$}
\put(20,5.2){\small $\phi_1$}
\put(12.75,6.35){\small $\phi_2$}
\put(7.5,8){\small $\phi_3$}
\put(61.5,1.5){\small $\phi_0$}
\put(71.5,1.5){\small $\phi_1$}
\put(82.5,1.5){\small $\phi_2$}
\put(92,1.5){\small $\phi_3$}
\end{overpic}\vspace{9mm}
\begin{overpic}
[width=1\linewidth]{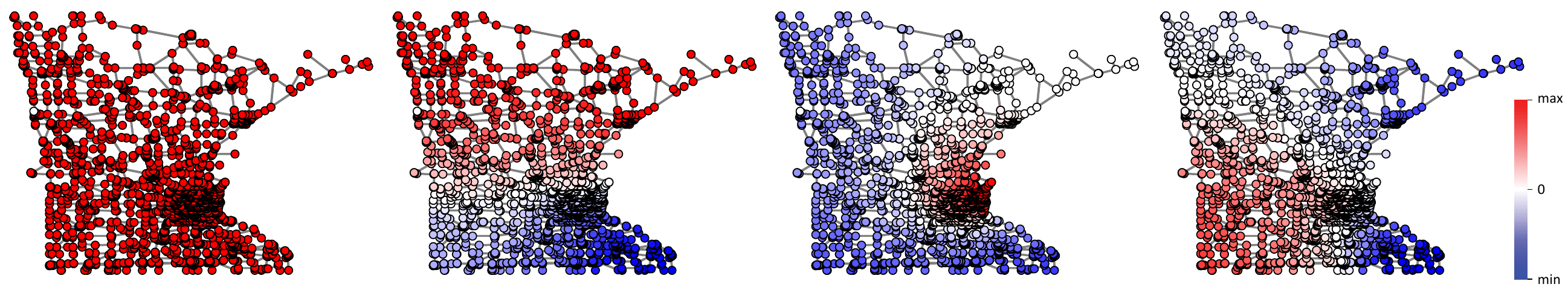}
\put(45,-2){\small Graph}
\put(19.8,1.5){\small $\phi_0$}
\put(44,1.5){\small $\phi_1$}
\put(68.8,1.5){\small $\phi_2$}
\put(93,1.5){\small $\phi_3$}
\end{overpic}
\end{center}\vspace{2mm}
{\small {\bf [FIGS3]} Example of the first four Laplacian eigenfunctions $\phi_0,\hdots, \phi_3$ on a Euclidean domain (1D line, top left) and non-Euclidean domains (human shape modeled as a 2D manifold, top right; and Minnesota road graph, bottom). In the Euclidean case, the result is the standard Fourier basis comprising sinusoids of increasing frequency. In all cases, the eigenfunction $\phi_0$ corresponding to zero eigenvalue is constant (`DC').}
%
%Third and fourth rows: examples of correspondence between different shapes affected by extreme transformations (missing parts) learned by intrinsic CNN . Corresponding points are encoded in similar color. Reference shape is shown on the left. 
%\caption{
%}
%\end{figure*}
}

\paragraph*{\bf Fourier analysis on non-Euclidean domains}
The Laplacian operator %, whether on a graph, manifold, or a discrete manifold, 
is a self-adjoint positive-semidefinite operator, admitting on a compact domain\footnote{In the Euclidean case, the Fourier transform of a function defined on a finite interval (which is a compact set) or its periodic extension is discrete. 
In practical settings, all domains we are dealing with are compact. 
}  an eigendecomposition with a discrete set of orthonormal eigenfunctions $\phi_0, \phi_1, \hdots $ (satisfying $\langle \phi_i, \phi_j \rangle_{L^2(\mathcal{X})} = \delta_{ij}$) and non-negative real eigenvalues $0 = \lambda_0 \leq \lambda_ 1 \leq \hdots$ (referred to as the {\em spectrum} of the Laplacian), 
\begin{eqnarray}
\Delta \phi_i = \lambda_i \phi_i, \hspace{3mm} i = 0, 1, \hdots
\end{eqnarray}
The eigenfunctions are the smoothest functions in the sense of the Dirichlet energy (see insert IN3) and can be interpreted as a generalization of the standard Fourier basis (given, in fact, by the eigenfunctions of the 1D Euclidean Laplacian, $-\tfrac{d^2}{x^2} e^{i\omega x} = \omega^2 e^{i\omega x}$) to a non-Euclidean domain. 
It is important to emphasize that the Laplacian eigenbasis is intrinsic due to the intrinsic construction of the Laplacian itself. 

A square-integrable function $f$ on $\mathcal{X}$ can be decomposed into {\em Fourier series} as 
\begin{eqnarray}
\label{eq:fourier}
f(x) &=& \sum_{i\geq 0} \underbrace{\langle f, \phi_i \rangle_{L^2(\mathcal{X})} }_{\hat{f}_i} \phi_i(x),
%\int_\mathcal{X} f(x) \phi_i(x) 
\end{eqnarray}
where the projection on the basis functions producing a discrete set of Fourier coefficients $(\hat{f}_i)$ generalizes the {\em analysis} (forward transform) stage in  classical  signal processing, and summing up the basis functions with these coefficients is the {\em synthesis} (inverse transform) stage.

A centerpiece of classical Euclidean signal processing is the property of the Fourier transform diagonalizing the convolution operator, colloquially referred to as the {\em Convolution Theorem}. This property allows to express the convolution $f \star g$ of two functions 
%%
%\begin{eqnarray}
%(f \star g)(x) &=& \int_{-\infty}^\infty f(x') g(x - x') dx'
%%\int_\mathcal{X} f(x) \phi_i(x) 
%\end{eqnarray}
%
in the spectral domain as the element-wise product of their Fourier transforms,
\begin{eqnarray}
\hspace{-2mm}(\widehat{f \star g})(\omega) &=& \int_{-\infty}^\infty f(x) e^{-i\omega x} dx \int_{-\infty}^\infty g(x) e^{-i\omega x} dx.  
%\int_\mathcal{X} f(x) \phi_i(x) 
\end{eqnarray}
Unfortunately, in the non-Euclidean case we cannot even define the operation $x-x'$ on the manifold or graph, so the notion of convolution~(\ref{eq:convdef}) does not directly extend to this case. %, let alone the convolution. 
One possibility to generalize convolution to non-Euclidean domains is by using the Convolution Theorem as a {\em definition},
\begin{eqnarray}
\label{eq:gconv}
(f \star g)(x) &=& \sum_{i\geq 0} \langle f, \phi_i \rangle_{L^2(\mathcal{X})} \langle g, \phi_i \rangle_{L^2(\mathcal{X})} \phi_i(x).  
%\int_\mathcal{X} f(x) \phi_i(x) 
\end{eqnarray}
One of the key differences of such a construction from the classical convolution is the lack of shift-invariance. 
In terms of signal processing, it can be interpreted as a position-dependent filter. While parametrized by a fixed number of coefficients in the frequency domain, the spatial representation of the filter can vary dramatically at different points (see FIGS4). %across the domain. 

%{\color{green} ADS: more on this.  Emphasize that the filter as it applies to one location may not even be well defined on another; and more generally neighborhoods are not comparable.  Also the consequences in terms of parameters.  Or maybe forward pointer to the discussion in section V?  but then that one could be expanded; and I think it is better here with a backward pointer in section V.}

The discussion above also applies to graphs instead of manifolds, where one only has to replace the inner product in equations (\ref{eq:fourier}) and (\ref{eq:gconv}) with the discrete one (\ref{eq:dinner1}). All the sums over $i$ would become finite, as the graph Laplacian $\boldsymbol{\Delta}$ has $n$ eigenvectors. 
% rather than infinitely many continuous eigenfunctions of the continuous Laplacian. 
%
In matrix-vector notation, the generalized convolution $f\star g$ can be expressed as $\mathbf{G}\mathbf{f} = \boldsymbol{\Phi} \, \mathrm{diag}(\hat{\mathbf{g}}) \boldsymbol{\Phi}^\top \mathbf{f}$, where $\hat{\mathbf{g}} = (\hat{g}_1, \hdots, \hat{g}_n)$ is the spectral representation of the filter and $\boldsymbol{\Phi} = (\boldsymbol{\phi}_1,\hdots, \boldsymbol{\phi}_n)$ denotes the Laplacian eigenvectors~(\ref{eq:lapeigs}). 
The lack of shift invariance results in the absence of circulant (Toeplitz) structure in the matrix $\mathbf{G}$, which characterizes the Euclidean setting. 
Furthermore, it is easy to see that the convolution operation commutes with the Laplacian, $\mathbf{G}\boldsymbol{\Delta}\mathbf{f} = \boldsymbol{\Delta}\mathbf{G}\mathbf{f}$.
%\begin{eqnarray}
%\label{eq:lapcommute}
%\mathbf{G}\boldsymbol{\Delta}\mathbf{f} &=& \boldsymbol{\Phi} \, \mathrm{diag}(\hat{\mathbf{g}}) \boldsymbol{\Phi}^\top \boldsymbol{\Delta}\mathbf{f} = \boldsymbol{\Delta} \boldsymbol{\Phi} \, \mathrm{diag}(\hat{\mathbf{g}}) \boldsymbol{\Phi}^\top \mathbf{f}.
%\end{eqnarray}
%which follows from the symmetry of all the involved matrices. 

%resulting which is not shift-invariant anymore. 
%As we will see next, such a construction allows generalizing convolutional neural networks to graphs and manifolds. 
%We will see next how to use it to generalize convolutional neural networks to graphs. %and manifolds

%\begin{figure*}[t!]
%\begin{mdframed}[linecolor=red,middlelinewidth=2]

\myframedtext*[\linewidth]{
\vspace{-5mm}
\begin{multicols}{2}
\paragraph*{\bf [IN4] Heat diffusion on non-Euclidean domains}
An important application of spectral analysis, and historically, the main motivation for its development by Joseph Fourier, is the solution of partial differential equations (PDEs). 
In particular, we are interested in heat propagation on non-Euclidean domains. This process is governed by the {\em heat diffusion equation}, which in the simplest setting of homogeneous and isotropic diffusion has the form
\begin{eqnarray}
\label{eq:diffusion}
\left\{
\begin{array}{l}
\displaystyle 
f_t(x,t) = - c \Delta f(x,t)\vspace{2mm}\\
f(x,0) = f_0(x) \,\,\,\, (\text{Initial condition})
\end{array}
\right.
\end{eqnarray}
with additional boundary conditions if the domain has a boundary. $f(x,t)$ represents the temperature at point $x$ at time $t$. 
Equation~(\ref{eq:diffusion}) encodes the {\em Newton's law of cooling}, according to which the rate of temperature change of a body (lhs) is proportional to the difference between its own temperature and that of the surrounding (rhs). The proportion coefficient $c$ is referred to as the {\em thermal diffusivity constant}; here, we assume it to be equal to one for the sake of simplicity. 
The solution of~(\ref{eq:diffusion}) is given by applying the {\em heat operator} $H^t = e^{-t \Delta}$ to the initial condition and can be expressed in the spectral domain as 
\begin{eqnarray}
\label{eq:heat1}
\hspace{-3mm}
f(x,t) &=& e^{-t\Delta} f_0(x) = \sum_{i\geq 0} \langle f_0, \phi_i\rangle_{L^2(\mathcal{X})} e^{- t \lambda_i} \phi_i(x) \\
&=& \int_{\mathcal{X}} f_0(x') \underbrace{\sum_{i \geq 0} e^{-t\lambda_i} \phi_i(x) \phi_i(x') }_{h_t(x,x')}dx'. \nonumber
\end{eqnarray}
$h_t(x,x')$ is known as the {\em heat kernel} and represents the solution of the heat equation with an initial condition $f_0(x) = \delta_{x'}(x)$, or, in signal processing terms, an `impulse response'. 
In physical terms, $h_t(x,x')$  describes how much heat flows from a point $x$ to point $x'$ in time $t$. 
In the Euclidean case, the heat kernel is {\em shift-invariant}, $h_t(x,x') = h_t(x -x')$, allowing to interpret the integral in~(\ref{eq:heat1}) as a convolution $f(x,t) = (f_0 \star h_t)(x)$. 
In the spectral domain, convolution with the heat kernel amounts to low-pass filtering with frequency response $e^{-t\lambda}$. Larger values of diffusion time $t$ result in lower effective cutoff frequency and thus smoother solutions in space (corresponding to the intuition that longer diffusion smoothes more the initial heat distribution).  \\
The `cross-talk' between two heat kernels positioned at points $x$ and $x'$ allows to measure an intrinsic distance %between these points, 
\begin{eqnarray}
\label{eq:norm1}
d_t^2(x,x') &=& \int_\mathcal{X} (h_t(x,y) - h_t(x',y))^2 dy \\
&=& \sum_{i\geq 0} e^{- 2 t \lambda_i}(\phi_i(x) - \phi_i(x'))^2
\label{eq:norm2}
\end{eqnarray}
referred to as the {\em diffusion distance} \cite{coifman2006diffusion}. 
Note that interpreting (\ref{eq:norm1}) and (\ref{eq:norm2}) as spatial- and frequency-domain norms $\| \, \cdot \, \|_{L^2(\mathcal{X})}$ and $\| \, \cdot \, \|_{\ell^2}$, respectively, their equivalence is the consequence of the {\em Parseval identity}. 
Unlike {\em geodesic distance} that measures the length of the shortest path on the manifold or graph, the diffusion distance has an effect of averaging over different paths. It is thus more robust to perturbations of the domain, for example, introduction or removal of edges in a graph, or `cuts' on a manifold. 
%The lack of shift invariance in the non-Euclidean case is one of the major differences from the Euclidean counterpart.  
%
\vspace{2mm}
\begin{center}
\begin{overpic}
[width=1\linewidth]{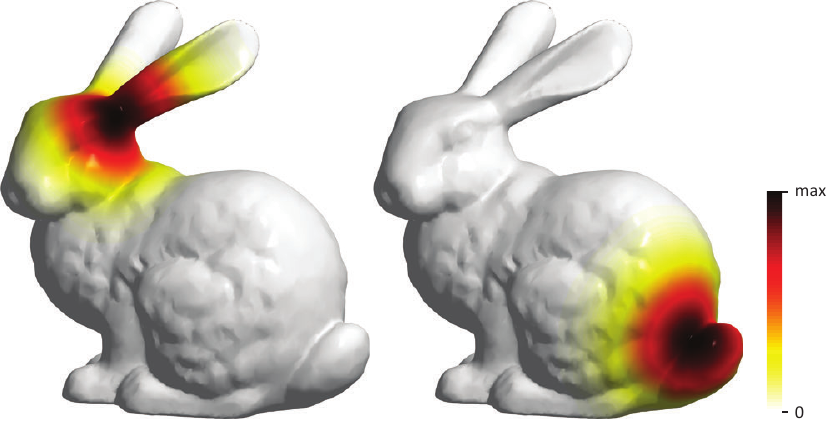}
%\put(20,-2){\small Euclidean}
\end{overpic}\vspace{2mm}
\begin{overpic}
[width=1\linewidth]{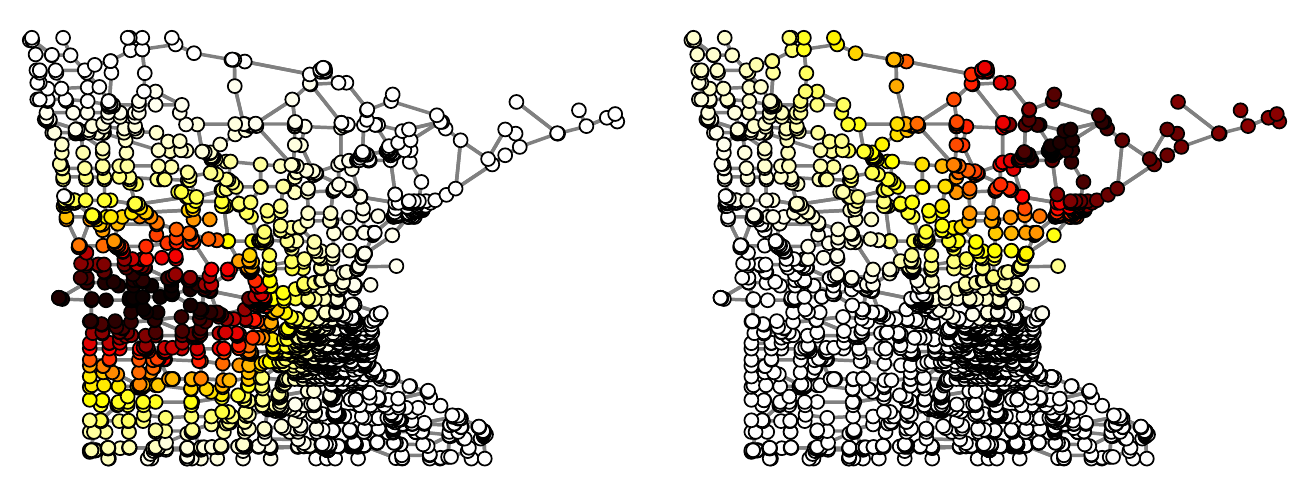}
%\put(20,-2){\small Euclidean}
\end{overpic}
\end{center}\vspace{0mm}
{\small {\bf [FIGS4]} Examples of heat kernels on non-Euclidean domains (manifold, top; and graph, bottom). Observe how moving the heat kernel to a different location changes its shape, which is an indication of the lack of shift-invariance. }
\end{multicols}
%
%Third and fourth rows: examples of correspondence between different shapes affected by extreme transformations (missing parts) learned by intrinsic CNN . Corresponding points are encoded in similar color. Reference shape is shown on the left. 
%\caption{
%}
%\end{figure*}
}

\paragraph*{\bf Uniqueness and stability}

Finally, it is important to note that the Laplacian eigenfunctions are not uniquely defined. 
To start with, they are defined up to sign, i.e., $\Delta (\pm\phi) = \lambda (\pm \phi)$.  
Thus, even isometric domains might have different Laplacian eigenfunctions. 
Furthermore, if a Laplacian eigenvalue has multiplicity, then the associated eigenfunctions can be defined as orthonormal basis spanning the corresponding eigen-subspace (or said differently, the eigenfunctions are defined up to an orthogonal transformation in the eigen-subspace). 
A small perturbation of the domain can lead to very large changes in the Laplacian eigenvectors, especially those associated with high frequencies.  
At the same time, the definition of heat kernels~(\ref{eq:heat1}) and diffusion distances~(\ref{eq:norm2}) does not suffer from these ambiguities -- for example, the sign ambiguity disappears as the eigenfunctions are squared. Heat kernels also appear to be robust to domain perturbations.

\section{Spectral methods}
\label{sec:freq}

We have now finally got to our main goal, namely, constructing a generalization of the CNN architecture on non-Euclidean domains. %to graphs. 
We will start with the assumption that the domain on which we are working is fixed, and for the rest of this section will use the problem of classification of a signal on a fixed graph as the prototypical application.  

%\paragraph*{\bf Spectral CNN}

%As in section \ref{sec:noneucl}, 
%let $W$ be a  weighted graph with index set denoted by $\Omega$, and let $\boldsymbol{\Phi}$
% be the unitary matrix containing the 
% eigenvectors of the graph Laplacian $\boldsymbol{\Delta}$, ordered by eigenvalue.
%In Section \ref{sec:noneucl}, 
%\paragraph*{\bf Spectral approaches}
We have seen that convolutions are linear operators that commute 
 with the Laplacian operator. Therefore, 
given a weighted graph, a first route to generalize a convolutional architecture is by first
restricting our interest to linear operators that commute with the graph Laplacian. % \cite{bruna2013spectral}.
This property, in turn, implies operating on the spectrum of the 
graph weights, given by the eigenvectors of the graph Laplacian.

% m->p (input), n->q

{\bf{\em Spectral CNN (SCNN)}} \cite{bruna2013spectral}: Similarly to the convolutional layer~(\ref{eq:convlayer}) of a classical Euclidean CNN, Bruna et al. \cite{bruna2013spectral} define a {\em spectral convolutional layer} as 
%
%For simplicity, let us first describe a construction where each layer $k=1\dots K$ 
%transforms an input vector $f_k$ of size $|\Omega| \times N_{k-1}$ into an output $f_{k+1}$ 
%of dimensions $|\Omega| \times N_{k}$, that is, without spatial subsampling:
\begin{equation} 
\label{spectral_construction_eq}
\mathbf{g}_l =   \xi \left(  \sum_{l'=1}^{q} \boldsymbol{\Phi}_k \mathbf{\W}_{l,l'} \boldsymbol{\Phi}_k^\top \mathbf{f}_{l'} \right),
%f_{k+1, l}  =   \rho \left( \boldsymbol{\Phi} \sum_{l'=1}^{N_{k-1}} W_{k,l,l'} \boldsymbol{\Phi}^T f_{k,l'} \right)~,~l=1\dots N_k~,
\end{equation}
where the $n\times p$ and $n\times q$ matrices $\mathbf{F} = (\mathbf{f}_1, \hdots, \mathbf{f}_p)$  and $\mathbf{G} = (\mathbf{g}_1, \hdots, \mathbf{g}_q)$ represent the $p$- and $q$-dimensional input and output signals on the vertices of the graph, respectively (we use $n = |\mathcal{V}|$ to denote the number of vertices in the graph), 
$\mathbf{\W}_{l,l'}$ is a $k\times k$ diagonal matrix of spectral multipliers representing a filter in the frequency domain, and $\xi$ is a nonlinearity applied on the vertex-wise function values.
Using only the first $k$ eigenvectors in~(\ref{spectral_construction_eq}) sets a cutoff frequency which depends on the intrinsic regularity of the graph and also the sample size. Typically,  $k \ll n$, since only the first Laplacian eigenvectors describing the smooth structure of the graph are useful in practice. 

%
%Often, only the first $d$ eigenvectors of the Laplacian are useful in practice, which  
%carry the smooth geometry of the graph.  $\boldsymbol{\Phi}$ by
%$\boldsymbol{\Phi}_d$, obtained by keeping the first $d$ columns of $\boldsymbol{\Phi}$.

If the graph has an underlying group invariance, such a construction can discover it. In particular, 
standard CNNs can be redefined from the spectral domain (see insert IN5). %; see \ref{sec:rediscover}. 
However, in many cases the graph does not 
have a group structure, or the group structure does not commute with the Laplacian, 
and so we cannot think of each filter as passing a template across $\mathcal{V}$ and recording the correlation 
of the template with that location. 

%\paragraph*{\bf Limitations of Spectral CNNs}

We should stress that a fundamental limitation of the spectral construction is its limitation to a single domain. 
The reason is that spectral filter coefficients~(\ref{spectral_construction_eq}) are {\em basis dependent}. It implies that if we learn a filter w.r.t. basis $\boldsymbol{\Phi}_k$ on one domain, and then try to apply it on another domain with another basis $\boldsymbol{\Psi}_k$, the result could be very different (see Figure~2 and insert IN6). 
%~\ref{fig:blobs} and insert IN6). 
%
%Consequently, spectral CNN can be considered as method for learning on a graph, rather than on multiple graphs. 
%
It is possible to construct compatible orthogonal bases across different domains resorting to a joint diagonalization procedure \cite{kovnatsky2013coupled,eynard2015multimodal}. However, such a construction requires the knowledge of some correspondence between the domains. In applications such as social network analysis, for example, where dealing with two time instances of a social graph in which new vertices and edges have been added, such a correspondence can be easily computed and is therefore a reasonable assumption. 
Conversely, in computer graphics applications, finding correspondence between shapes is in itself a very hard problem, so assuming known correspondence between the domains is a rather unreasonable assumption. 
%
%In the next section, we will see a different generalization of the convolution that allows applying CNN-type models across different domains. 

\begin{figure}[h!]
\centering
\begin{overpic}
[width=1\linewidth]{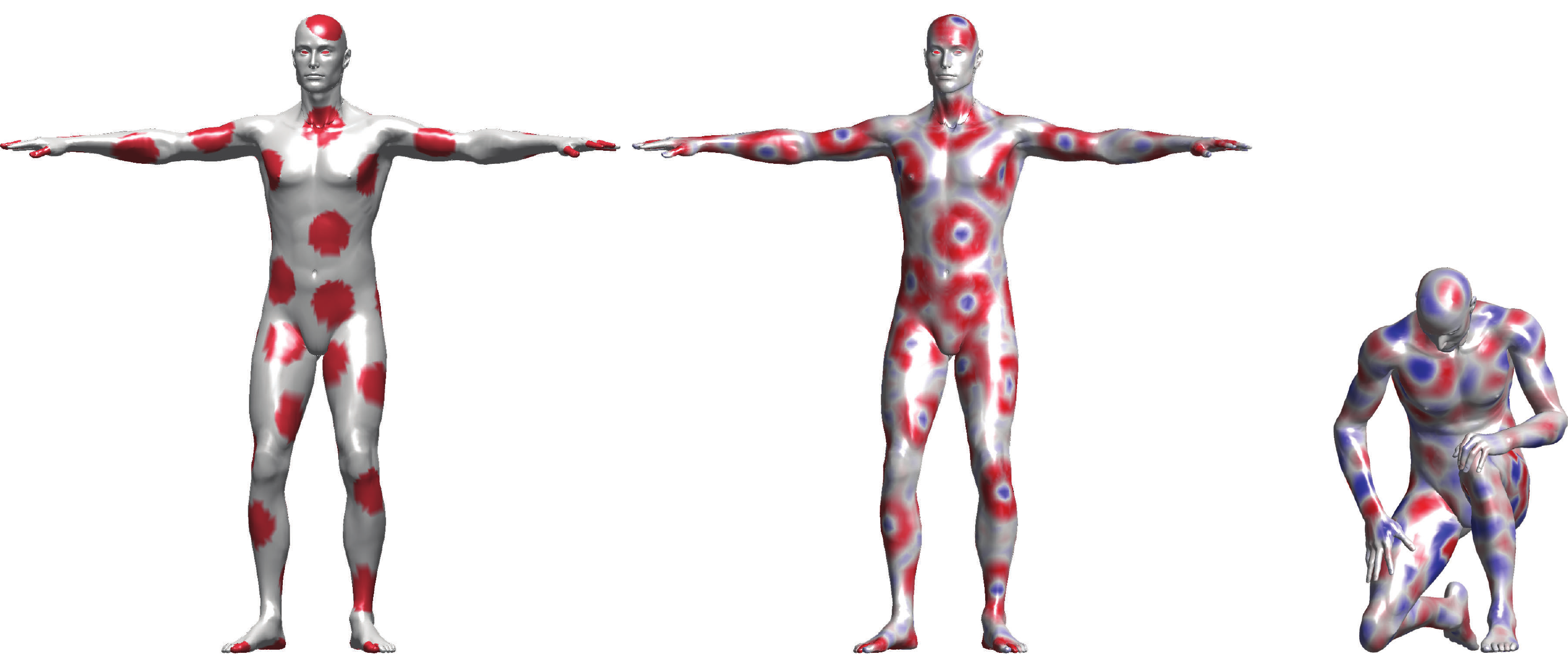}
\put(0,-4){\small Domain}
\put(0,-9){\small Basis}
\put(0,-14){\small Signal}
\put(18,-4){\small $\mathcal{X}$}
\put(18,-9){\small $\boldsymbol{\Phi}$ }%$\{ \phi \}_{i\geq 0}$}
\put(18,-14){\small $\mathbf{f}$}
\put(58.5,-4){\small $\mathcal{X}$}
\put(58.5,-9){\small $\boldsymbol{\Phi}$ }%$\{ \phi \}_{i\geq 0}$}
\put(54.5,-14){\small $\boldsymbol{\Phi} \mathbf{W} \boldsymbol{\Phi}^\top\mathbf{f}$}%$\sum_{i\geq 0} \hat{w}_i \langle f,\phi_i \rangle_{L^2(\mathcal{X})} \phi_i$}
\put(90,-4){\small $\mathcal{Y}$}
\put(90,-9){\small $\boldsymbol{\Psi}$} %$\{ \psi \}_{i\geq 0}$}
\put(85,-14){\small $\boldsymbol{\Psi} \mathbf{W} \boldsymbol{\Psi}^\top\mathbf{f}$} %$\{ \psi \}_{i\geq 0}$}
\end{overpic}
\vspace{10mm}
\label{fig:blobs}
\caption{
\small A toy example illustrating the difficulty of generalizing spectral filtering across non-Euclidean domains. 
Left: a function defined on a manifold (function values are represented by color); middle: result of the application of an edge-detection filter in the frequency domain; right: the same filter applied on the same function but on a different (nearly-isometric) domain produces a completely different result. The reason for this behavior is that the Fourier basis is domain-dependent, and the filter coefficients learnt on one domain cannot be applied to another one in a straightforward manner.
}
\end{figure}
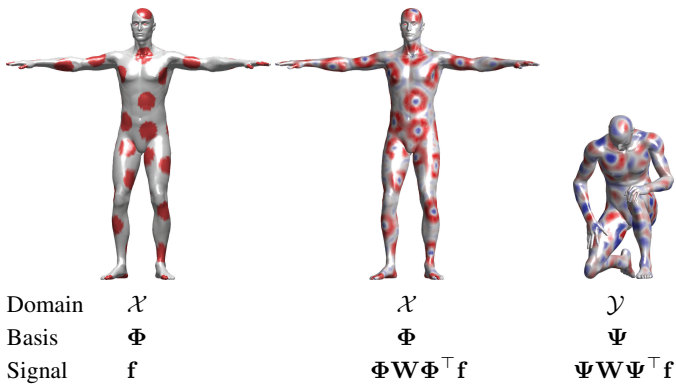

%Number of parameters
Assuming that $k = O(n)$ eigenvectors of the Laplacian are kept, 
a convolutional layer~(\ref{spectral_construction_eq}) requires 
$p q k = O(n)$ parameters to train. 
We will see next how the global and local 
regularity of the graph can be combined to produce layers with constant number of
parameters (i.e., such that the number of learnable parameters per layer does not depend upon 
the size of the input), which is the case in classical Euclidean CNNs. 

%Another important property that relates classical Euclidean CNN with the 
%spectral domain construction is the downsampling or pooling operation. 

The non-Euclidean analogy of pooling is {\em graph coarsening}, in which only a fraction $\alpha < 1$ of the graph vertices is retained. 
The eigenvectors of graph Laplacians at two different resolutions are related by the following
multigrid property: Let $\boldsymbol{\Phi}$, $\tilde{\boldsymbol{\Phi}}$ denote the $n \times n$ and $\alpha n \times \alpha n$ matrices of Laplacian eigenvectors of the original and the coarsened graph, respectively.  
%denote the eigenvectors of the original graph Laplacian and the coarse graph Laplacian 
%respectively (where we kept a fraction $\alpha$ of the nodes), then 
Then, 
\begin{equation}
\tilde{\boldsymbol{\Phi}} \approx \mathbf{P} \boldsymbol{\Phi} \left( 
\begin{array}{c} 
\mathbf{I}_{\alpha n} \\
\mathbf{0}
\end{array}
\right), 
\end{equation}
where $\mathbf{P}$ is a $\alpha n \times n$ binary matrix whose $i$th row encodes the position of the $i$th vertex of the coarse graph on the original graph. 
It follows that strided convolutions can be generalized using the spectral construction by keeping
only the low-frequency components of the spectrum.
This property also allows us to interpret (via interpolation) 
the local filters at deeper layers in the spatial construction to be low frequency.  
However, since in~(\ref{spectral_construction_eq}) the non-linearity is applied in the spatial domain, 
in practice one has to recompute the graph Laplacian eigenvectors at each resolution and apply them directly 
after each pooling step.

%The spectral construction can suffer from the fact that most graphs have meaningful eigenvectors 
%only for the very top of the spectrum. 
%Even when the individual high frequency eigenvectors are not 
%meaningful, a cohort of high frequency eigenvectors may contain meaningful information.  However 
%this construction may not be able to access this information because it is nearly diagonal at the highest frequencies.  
The spectral construction (\ref{spectral_construction_eq}) assigns a degree of freedom for each 
eigenvector of the graph Laplacian. In most graphs, individual high-frequency eigenvectors become highly unstable.  
%as one accesses smaller eigenvalues, which are associated with higher frequency information. 
However, 
similarly as the wavelet construction in Euclidean domains, by appropriately grouping high frequency 
eigenvectors in each octave one can recover meaningful and stable information. As we shall see 
next, this principle also entails better learning complexity. 

 %{\color{red}MB: This is not clear to me- Joan, can you explain what is meant here?}

% it is not obvious how to do either the forwardprop or the backprop efficiently while applying the nonlinearity 
%on the space side, as we have to make the expensive multiplications by $V$ and $V^T$; and it is not obvious 
%how to do standard nonlinearities on the spectral side. However, see \ref{sec:multigrid}.  
% 
%Why is the previous spectral construction not always good? 
%Stability with respect to noise / sampling. Only the regular part of the spectrum is in practice stable. 
%High frequencies are important for recognition and are not efficiently captured with 
%the Laplacian eigenmaps.
 
 \myframedtext*[\linewidth]{
\vspace{-5mm}
\begin{multicols}{2}
 \paragraph*{\bf [IN5] Rediscovering standard CNNs using correlation kernels}
 \label{sec:rediscover}
 In situations where the graph is constructed from the data, a straightforward choice of the edge weights~(\ref{eq:graph_lap}) of the graph 
is the covariance of the data.  
Let $\mathbf{F}$ denote the input data distribution and %, where each row represents a data point, and denote by 
\begin{equation}
\boldsymbol{\Sigma} = \E (\mathbf{F} - \E \mathbf{F}) (\mathbf{F} - \E \mathbf{F})^\top
\end{equation}
be the data covariance matrix. If each point has the same variance $\sigma_{ii} = \sigma^2$, 
%%
%Let $\mathbf{F}=(f_k)_k$ be the input data distribution, with $f_k \in \mathbb{R}^n$.
%If each coordinate $j=1\dots n$ has the same variance,  
%$$\sigma^2_j = \E\left( | f(j) -  \E(f(j))|^2\right)~,$$
then diagonal operators on the Laplacian simply scale the principal components of $\mathbf{F}$.  \\
In natural images, since their distribution is approximately stationary, 
%its covariance operator 
%$$\Sigma(j,j) = \E\left( (f(j) -  \E(f(j)))(f(j') -  \E(f(j')))\right) $$
%satisfies $\Sigma(j,j') = \Sigma(j-j')$, hence it is 
the covariance matrix has a circulant structure $\sigma_{ij} \approx \sigma_{i-j}$ and is thus 
diagonalized by the standard Discrete Cosine Transform (DCT) basis. 
It follows that the principal 
components of $\mathbf{F}$ roughly correspond to the DCT basis vectors %Discrete Cosine Transform basis, 
ordered by frequency. 
 Moreover, natural images exhibit a power spectrum $\E |\widehat{f}(\omega)|^2 \sim |\omega|^{-2} $, 
since nearby pixels are more correlated than far away pixels \cite{simoncelli2001natural}.
It results that principal components of the covariance are essentially ordered from
low to high frequencies, which is consistent with the standard group structure of the 
Fourier basis.
When applied to natural images represented as graphs with weights defined by the covariance, the spectral CNN construction   
recovers the standard CNN, 
 without any prior knowledge %. This property was also studied in 
 \cite{roux2008learning}. 
 Indeed, the linear operators $\boldsymbol{\Phi} \mathbf{\W}_{l,l'} \boldsymbol{\Phi}^\top$ in~(\ref{spectral_construction_eq}) 
 are by the previous argument diagonal in the Fourier basis, hence translation invariant, 
 hence classical convolutions. 
 Furthermore, Section \ref{sec:freq} explains how spatial 
 subsampling can also be obtained via dropping the last part of the spectrum of the Laplacian, 
 leading to pooling, and ultimately to standard CNNs.
\begin{center}\vspace{3mm}
\begin{overpic}
[width=0.45\linewidth]{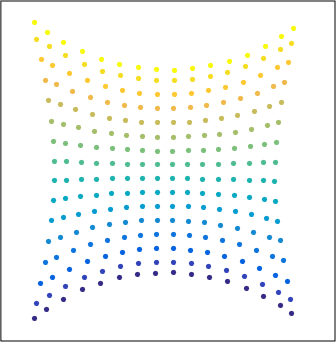}
\end{overpic}\hspace{5mm}
\begin{overpic}
[width=0.45\linewidth]{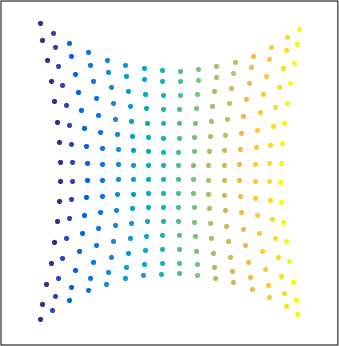}
\end{overpic}
\end{center}\vspace{0mm}
{\small {\bf [FIG5a]} Two-dimensional embedding of pixels in $16\times 16$ image patches using a Euclidean RBF kernel. 
The RBF kernel is constructed as in (\ref{eq:graph_lap}), by using the covariance 
$\sigma_{ij}$ as Euclidean distance between two features. The pixels are embedded in a 2D space using the first two eigenvectors of the resulting graph Laplacian. The colors in the left and right figure represent the horizontal and vertical coordinates of the pixels, respectively. 
The spatial arrangement of pixels is roughly recovered from correlation measurements.
}
  \end{multicols}
}

%\paragraph*{\bf 
{\bf{\em Spectral CNN with smooth spectral multipliers}} \cite{bruna2013spectral,henaff2015deep}:
%Learning with smooth spectral multipliers}
%\label{smoothspect}
%
In order to reduce the risk of overfitting, it is important to adapt the learning 
complexity to reduce the number of free parameters of the model. 
On Euclidean domains, this is achieved by learning convolutional kernels with small spatial support,  which enables the model to learn a number of parameters independent of the input size. 
In order to achieve a similar learning complexity in the spectral domain, it is thus necessary to restrict the class of spectral multipliers to those corresponding to localized filters. 

For that purpose, we have to express spatial localization of filters in the frequency domain. 
In the Euclidean case, smoothness in the frequency domain corresponds to  spatial decay, since
%$$\left| \frac{\partial^k \hat{f}(\omega)}{\partial \omega^k} \right| \leq C \int^{+\infty}_{-\infty} |x|^k |f(x)| dx.$$
\begin{equation}
\int_{-\infty}^{+\infty} |x|^{2k} |f(x)|^2 dx = \int_{-\infty}^{+\infty} \left| \frac{\partial^k \hat{f}(\omega)}{\partial \omega^k} \right|^2 d\omega,
\end{equation}
by virtue of the Parseval Identity.
%where $\hat{f}(\omega)$ is the Fourier transform of $f$.
%
% In the standard grid, we do not need a parameter for each 
% frequency because the filters 
% are compactly supported in space, but in (\ref{spectral_construction_eq}), 
% each filter requires one parameter for each eigenvector on which it acts. 
% Even if the filters were compactly supported in space in this construction, 
% we still would not get less than $O(N)$ parameters per filter because the spatial 
% response would be different at each location.  
 %
% One possibility for getting around this is to generalize the duality of the grid.  
% On the Euclidian grid, the decay of a function in the spatial domain is translated 
% into smoothness in the Fourier domain, and viceversa. 
%It results that a function $x$ which is spatially localized 
%  has a smooth frequency response $\hat{x} = V^T x$. 
%In that case, the eigenvectors of the Laplacian can be thought of as being arranged on a grid 
%isomorphic to the original spatial grid.
%
This suggests that, in order to learn a layer in which features will be not only 
shared across locations but also well localized in the original domain,
one can learn spectral multipliers which are smooth. 
Smoothness can be prescribed by learning only a subsampled set
of frequency multipliers and using an interpolation kernel 
to obtain the rest, such as cubic splines.

However, the notion of smoothness also requires some geometry in the spectral domain.  
In the Euclidean setting, such a geometry naturally arises from the notion of frequency; for example, in the plane, the similarity between two Fourier atoms $e^{i \boldsymbol{\omega}^\top \mathbf{x}}$ and $e^{i \boldsymbol{\omega'}^\top \mathbf{x}}$
 can be quantified by the distance $\| \boldsymbol{\omega} - \boldsymbol{\omega'}\|$, where $\mathbf{x}$ denotes the two-dimensional planar coordinates, and $\boldsymbol{\omega}$ is the two-dimensional frequency vector. 
On graphs, such a relation 
% but this cannot be directly generalized to other graphs. 
can be defined by means of a dual graph %$\widetilde{W}=(\tilde{w}_{i,j})$ 
with weights $\tilde{w}_{ij}$ encoding the similarity between two eigenvectors $\phi_i$ and $\phi_j$.

A particularly simple choice consists in choosing a 
one-dimensional arrangement, obtained by ordering the eigenvectors according
to their eigenvalues. \footnote{
In the mentioned 2D example, this would correspond to ordering the Fourier basis function according to the sum of the corresponding  frequencies $\omega_1 + \omega_2$. 
Although numerical results on simple low-dimensional graphs show that the 1D arrangement 
given by the spectrum of the Laplacian is efficient at creating spatially localized filters \cite{bruna2013spectral}, an open fundamental question is how to define a dual graph on the eigenvectors of the Laplacian in which smoothness (obtained by 
applying the diffusion operator) corresponds to localization in the original graph. 
}
In this setting, 
the spectral multipliers are parametrized as %diagonal of each filter $\mathbf{W}_{l,l'}$ (of size at most $|\Omega|)$ is parametrized by
\begin{equation}
\label{eq:bspline}
\mbox{diag}(\mathbf{\W}_{l,l'}) = \mathbf{B} \boldsymbol{\alpha}_{l,l'}, 
\end{equation}
where $\mathbf{B} = (b_{ij}) = (\beta_j(\lambda_i))$ is a $k \times q$ 
fixed interpolation kernel (e.g., $\beta_j(\lambda)$ can be cubic splines) and 
$\boldsymbol{\alpha}$ is a vector of $q$ interpolation coefficients. 
In order to obtain filters with constant spatial support (i.e., independent of the input size $n$), 
one should choose a sampling step $\gamma \sim n$ in the spectral domain, which 
results in a constant number $n \gamma^{-1} = \mathcal{O}(1)$ of coefficients $\boldsymbol{\alpha}_{l,l'}$ per filter.
\editJB{
Therefore, by combining spectral layers with graph coarsening, this model has  
$\mathcal{O}( \log n)$ total trainable parameters for inputs of size $n$, thus recovering the same 
learning complexity as CNNs on Euclidean grids. }

Even with such a parametrization of the filters, the spectral CNN~(\ref{spectral_construction_eq}) entails a high computational complexity of performing 
forward and backward passes, since they require an expensive step of matrix multiplication  by $\boldsymbol{\Phi}_k$ and $\boldsymbol{\Phi}_k^\top$. While on Euclidean domains such a multiplication can be efficiently carried in $\mathcal{O}(n\log n)$ operations using FFT-type algorithms, for general graphs such algorithms do not exist and the complexity is $\mathcal{O}(n^2)$.
We will see next %in Section \ref{sec:spat_freq} the following 
how to alleviate this cost by avoiding explicit computation of the Laplacian eigenvectors. 

%A possible algorithmic strategy is to consider an input distribution $X=(x_k)_k$ consisting on spatially localized 
%signals and to construct a dual graph $\widehat{W}$ by measuring the similarity of 
%in the spectral domain: $\widehat{X}= V^T X$. 
%The similarity could be measured for instance
%with $E( ( |\hat{x}| -E(|\hat{x})|))^T( |\hat{x}| -E(|\hat{x}|))$.

%\paragraph*{\bf Spectrum-free computation}

%Chebyshev polynomials \cite{defferrard2016convolutional}

\section{Spectrum-free methods}
\label{sec:freq}

\editMB{
A polynomial of the Laplacian acts as a polynomial on the eigenvalues.   
Thus, instead of explicitly operating in the frequency domain with spectral multipliers as in equation~\eqref{eq:bspline}, it is possible to represent the filters 
%Another convenient parametric way of representing the convolution filters is 
via a polynomial expansion:}
\begin{eqnarray}
\label{eq:poly_filter}
	g_{\boldsymbol{\alpha}}(\boldsymbol{\Delta}) &=& \boldsymbol{\Phi} g_{\boldsymbol{\alpha}}(\boldsymbol{\Lambda}) \boldsymbol{\Phi}^\top, %\\
%	%
%	g_{\boldsymbol{\alpha}}(\boldsymbol{\Lambda}) &=& 
%	\left(
%	\begin{array}{ccc} 
%	g_{\boldsymbol{\alpha}}(\lambda_1) &&\\
%	&\cdots&\\
%	&&	g_{\boldsymbol{\alpha}}(\lambda_{n})
%\end{array}
%	\right)
\end{eqnarray}
corresponding to %~\cite{defferrard2016convolutional}. 
\begin{equation} \label{eq:filt_poly}
	g_{\boldsymbol{\alpha}}(\lambda) = \sum_{j=0}^{r-1} \alpha_j \lambda^j.
\end{equation}
Here $\boldsymbol{\alpha}$ is the $r$-dimensional vector of polynomial coefficients, and $g_{\boldsymbol{\alpha}}(\boldsymbol{\Lambda}) = \mathrm{diag}( g_{\boldsymbol{\alpha}}(\lambda_1), \hdots, g_{\boldsymbol{\alpha}}(\lambda_n) )$,  
resulting in filter matrices $\mathbf{\W}_{l,l'} = g_{\boldsymbol{\alpha}_{l,l'}}(\boldsymbol{\Lambda})$ whose entries have an explicit form in terms of the eigenvalues.

%
%In this case, the filter matrix $W_{k,l,l'}$ is a diagonal matrix whose entries have an explicit form in terms of the eigenvalues. Ignoring the filter index for the sake of clarity, we write $W_{l,l} = g_\theta(\lambda_l)$, where~:
%\begin{equation} \label{eq:filt_poly}
%	g_\theta(\Lambda) = \sum_{j=0}^{J-1} \theta_j \Lambda^j,
%\end{equation}
%and the parameter $\theta \in \mathbb{R}^J$ is a vector of polynomial coefficients. 

An important property of this representation is that it automatically yields localized filters, for the following reason.  
Since the Laplacian is a local operator (working on 1-hop neighborhoods), the action of its $j$th power is constrained to $j$-hops. Since the filter is a linear combination of powers of the Laplacian, overall  %$g_{\boldsymbol{\alpha}}(\boldsymbol{\Delta})$ 
(\ref{eq:filt_poly}) behaves like a diffusion operator limited to $r$-hops around each vertex. 

%Indeed, applying the spectral theorem, we can write $V W V^T = g_\theta(L)$ and we observe that filtering is obtained by applying a polynomial of the Laplacian operator. Such filters behave like diffusion operators, whose action is constrained to $J$-hop neighbourhoods. 

{\bf{\em Graph CNN (GCNN) a.k.a. ChebNet}} \cite{defferrard2016convolutional}: Defferrard et al. used 
%$g_{\boldsymbol{\alpha}}(\lambda)$ as a polynomial that can be computed recursively.  
%%Besides their ease of interpretation, polynomial filters can also be applied very efficiently if one chooses
%$g_{\boldsymbol{\alpha}}(\lambda)$ as a polynomial that can be computed recursively. For instance, the 
Chebyshev polynomial %$T_j(\lambda)$ of order $j$ 
generated by the recurrence relation 
\begin{eqnarray}
\label{eq:cheby}
T_j(\lambda) &=& 2\lambda T_{j-1}(\lambda) - T_{j-2}(\lambda);\\ 
T_0(\lambda) &=&1; \nonumber \\
T_1(\lambda) &=& \nonumber \lambda. 
\end{eqnarray}
A filter can thus be parameterized uniquely via an
 expansion of order $r-1$ such that
\begin{eqnarray} \label{eq:filt_cheby}
	g_{\boldsymbol{\alpha}}(\tilde{\boldsymbol{\Delta}}) &=& \sum_{j=0}^{r-1} \alpha_j  \boldsymbol{\Phi} T_j(\tilde{\boldsymbol{\Lambda}})\boldsymbol{\Phi}^\top \\
	&=& \sum_{j=0}^{r-1} \alpha_j T_j(\tilde{\boldsymbol{\Delta}}),  \nonumber
\end{eqnarray}
%where the parameter $\theta \in \mathbb{R}^J$ is a vector of Chebyshev coefficients and
%$T_j(\tilde{\Lambda}) \in \mathbb{R}^{n \times n}$ is the Chebyshev polynomial of order
%$j$ evaluated at 
where $\tilde{\boldsymbol{\Delta}} = 2 \lambda_{n}^{-1}\boldsymbol{\Delta}  - \mathbf{I}$ and $\tilde{\boldsymbol{\Lambda}} = 2 \lambda_{n}^{-1} \boldsymbol{\Lambda}  - \mathbf{I}$ denotes a rescaling of the Laplacian mapping its eigenvalues from the interval $[0, \lambda_{n}]$ to $[-1,1]$ (necessary since the Chebyshev polynomials form an orthonormal basis in $[-1,1]$).

%Filtering a signal $\mathbf{f}$ can now be written as 
%\begin{equation}
%g_{\boldsymbol{\alpha}}(\boldsymbol{\Delta}) \mathbf{f}
%= \sum_{j=0}^{r-1} \alpha_j T_j(\tilde{\boldsymbol{\Delta}}) \mathbf{f}, 
%\end{equation}
%where $\tilde{\boldsymbol{\Delta}} = 2 \lambda_{n}^{-1}\boldsymbol{\Delta}  - \mathbf{I}$ is the rescaled Laplacian. 
%$T_j(\tilde{L}) \in \mathbb{R}^{n
%\times n}$ is the Chebyshev polynomial of order $j$ evaluated at the re-scaled
%Laplacian $\tilde{L} = 2 L / \lambda_{max} - I_n$. %Note that the spectrum of the
%normalized Laplacian is bounded by $2$ \cite{book:Chung97Spectral}, such that
%the scaling can simply be $\tilde{L} = L - I_n$. 
Denoting $\bar{\mathbf{f}}^{(j)} =
T_j(\tilde{\boldsymbol{\Delta}})\mathbf{f}$, we can use the recurrence relation~(\ref{eq:cheby}) to compute
$\bar{\mathbf{f}}^{(j)} = 2\tilde{\boldsymbol{\Delta}} \bar{\mathbf{f}}^{(j-1)} - \bar{\mathbf{f}}^{(j-2)}$ with $\bar{\mathbf{f}}^{(0)} = \mathbf{f}$ and
$\bar{\mathbf{f}}^{(1)} = \tilde{\boldsymbol{\Delta}}\mathbf{f}$. The computational complexity of this procedure is therefore $\mathcal{O}(rn)$ operations and does not require an explicit computation of the Laplacian eigenvectors.
%\paragraph*{\bf Spectrum-free computation}
%
%Chebyshev polynomials \cite{defferrard2016convolutional}

{\bf{\em Graph Convolutional Network (GCN)}} \cite{welling2016}: Kipf and Welling simplified this  construction by further assuming $r=2$ and $\lambda_n \approx 2$, resulting in filters of the form
\begin{eqnarray}
g_{\boldsymbol{\alpha}}( \mathbf{f} ) &=& \alpha_0 \mathbf{f} + \alpha_1 (\boldsymbol{\Delta} - \mathbf{I}) \mathbf{f}   \nonumber \\
&=& \alpha_0 \mathbf{f} - \alpha_1 \mathbf{D}^{-1/2} \mathbf{W} \mathbf{D}^{-1/2} \mathbf{f}. 
\end{eqnarray}
Further constraining $\alpha = \alpha_0 = -\alpha_1$, one obtains filters represented by a single parameter, 
\begin{eqnarray}
g_\alpha( \mathbf{f} ) &=& \alpha ( \mathbf{I} + \mathbf{D}^{-1/2} \mathbf{W} \mathbf{D}^{-1/2}) \mathbf{f}. 
\end{eqnarray}
Since the eigenvalues of $\mathbf{I} + \mathbf{D}^{-1/2} \mathbf{W} \mathbf{D}^{-1/2}$ are now in the range $[0,2]$, repeated application of such a filter can result in numerical instability. This can be remedied by a renormalization
\begin{eqnarray}
\label{eq:welling}
g_\alpha( \mathbf{f} ) &=& \alpha \tilde{\mathbf{D}}^{-1/2} \tilde{\mathbf{W}} \tilde{\mathbf{D}}^{-1/2} \mathbf{f}, 
\end{eqnarray}
where $\tilde{\mathbf{W}} = \mathbf{W} + \mathbf{I}$ and $\tilde{\mathbf{D}} = \mathrm{diag}(\sum_{j \neq i} \tilde{w}_{ij})$. 

\editADS{
Note that though we arrived at the constructions of ChebNet and GCN starting in the spectral domain, they boil down to applying simple filters acting on the $r$- or 1-hop neighborhood of the graph in the spatial domain.   We consider these constructions to be examples of the more general Graph Neural Network (GNN) framework:}
%We can thus consider these as spatial-domain methods, which are discussed in the following section. 
%Furthermore, GCN can be formulated as an instance of the Graph Neural Network (GNN) model described in the next section.  
%The more general ChebNet can also be considered a spatial-domain method with $r$-hop spatial filters. 
 
%Note that this also can be considered an example of the GNN construction discussed in Section \ref{sec:spat} and Equation \eqref{eq:gnn}

{\bf{\em Graph Neural Network (GNN)}} \cite{GNN}: 
%Many of the spatial-domain methods described above require notions of charts and local orientations as a ``filter'' needs to be matched against a neighborhood.  
%The spatial-domain methods above use some notion of local Euclidean structure to define filtering operations. 
Graph Neural Networks generalize the notion of applying the filtering operations directly on the graph via the graph weights.
%
%A more general spatial construction on graphs has been proposed \cite{gori2005new,GNN} and has been simplified in \cite{GGSNN,comnets}.  
\editJB{
Similarly as Euclidean CNNs learn generic filters as linear combinations of localized, oriented bandpass and lowpass filters, a Graph Neural Network learns
at each layer a generic linear combination of graph low-pass and high-pass operators. These are given respectively by $f \mapsto \mathbf{W}f$ and $f \mapsto \mathbf{\Delta} f$, 
and are thus generated by the degree matrix $\mathbf{D}$ and the diffusion matrix $\mathbf{W}$.  
Given a $p$-dimensional input signal on the vertices of the graph, represented by the $n\times p$ matrix $\mathbf{F}$, %, the application of the Laplacian is an intrinsic operation that can be broken down into $\mathbf{W}\mathbf{F}$ and $\mathbf{D}\mathbf{F}$. 
%
%we have so far 
%considered two fundamental intrinsic linear operators:
%$$(\mathbf{D}\mathbf{f})_i = \left( \sum_{j \neq i} w_{ij}\right) \mathbf{f}_i~,\text{and}~(\mathbf{W}\mathbf{f})_i = \sum_{j \neq i} w_{ij}  \mathbf{f}_j~.$$
the GNN considers a generic nonlinear function $\eta_{\boldsymbol{\theta}}: \R^p \times \R^p \to \R^q$, 
parametrized by trainable parameters $\boldsymbol{\theta}$ that is applied to all nodes of the graph, 
\begin{equation}
\label{eq:gnn}
\mathbf{g}_i = \eta_{\boldsymbol{\theta}} \left( (\mathbf{W}\mathbf{f})_i, (\mathbf{D}\mathbf{f})_i \right).
\end{equation}
In particular, choosing $\eta( \mathbf{a},\mathbf{b}) = \mathbf{b} - \mathbf{a}$ one recovers the Laplacian operator $\mathbf{\Delta}\mathbf{f}$, but more general, nonlinear choices for $\eta$ yield trainable, task-specific diffusion operators.}  
Similarly as with a CNN architecture, one can stack the resulting GNN layers $\mathbf{g} = C_{\boldsymbol{\theta}}(\mathbf{f})$ and interleave them with graph pooling operators.
Chebyshev polynomials  $T_r(\mathbf{\Delta})$  can be obtained with $r$ layers of (\ref{eq:gnn}), making it possible, in principle, to consider ChebNet and GCN as particular instances of the GNN framework. 
%Some of the previous constructions can be cast as special cases of~(\ref{eq:gnn}). 

\editADS{
Historically, a version of GNN was
the first formulation of deep learning on graphs, proposed
in \cite{gori2005new,GNN}.  These works optimized over the parameterized steady state of some diffusion
process (or random walk) on the graph.  This can be interpreted as in equation \eqref{eq:gnn}, but using a large number of layers where each $C_{\boldsymbol{\theta}}$ is identical, as the forwards through the $C_{\boldsymbol{\theta}}$ approximate the steady state.   Recent works \cite{duvenaud2015convolutional,GGSNN,comnets,physical_dynamics,interaction_networks} relax the requirements of approaching the steady state or using repeated applications of the same $C_{\boldsymbol{\theta}}$.

%The construction uses parameterized functions $f^k$ corresponding to the $k$th layer of the network; these input a vector $h^{k-1}_i$ associated to vertex $i$ in the graph, and a vector $c^{k-1}_i$ associated to vertex $i$ representing the communication of the neighbors of $i$ to $i$.   The $h$ and $c$ are updated via   
%\begin{eqnarray}
%h^{k+1}_i &=& f^k(h^{k}_j, c^{k}_j)  \\ 
%c^{k+1}_i &=& \frac{1}{s_i}\sum_{j}  w_{ij} h^{k+1}_{j}.
%\label{eq:gnn}
%\end{eqnarray}
%Note that the computation does not need the graph to defined until the network is to be evaluated.   Thus the learnable functions $f^k$ are decoupled from the graph structure, and are not fixed to use one specific graph.  This is the crux of the models flexibility: the model eats a graph, and outputs vectors on the nodes.   

Because the communication at each layer is local to a vertex neighborhood, one may worry that it would take many layers to get information from one part of the graph to another, requiring multiple hops (indeed, this was one of the reasons for the use of the steady state in \cite{GNN}). 
 However, for many applications, it is not necessary for information to completely traverse the graph.   Furthermore, note that the graphs at each layer of the network need not be the same.  Thus we can replace the original neighborhood structure with one's favorite multi-scale coarsening of the input graph, and operate on that to obtain the same flow of information as with the convolutional nets above (or rather more like a ``locally connected network'' \cite{coates2011selecting}).  This also allows producing a single output for the whole graph (for ``translation-invariant" tasks), rather than a per-vertex output, by connecting each to a special output node.  Alternatively, one can allow $\eta$ to use not only $\mathbf{W}\mathbf{f}$ and $\mathbf{\Delta}\mathbf{f}$ at each node, but also $\mathbf{W}^s \mathbf{f}$ for several diffusion scales $s>1$, (as in \cite{defferrard2016convolutional}), giving the GNN the ability to learn algorithms such as the power method, and more directly accessing spectral properties of the graph.
 
The GNN model can be further generalized to replicate other operators on graphs. For instance, 
the point-wise nonlinearity $\eta$ can depend on the vertex type, allowing extremely rich  architectures \cite{duvenaud2015convolutional,GGSNN,comnets,physical_dynamics,interaction_networks}. 
 
}

%\editMB{It thus allows formulating spectrum-free methods such as GCN as instances of GNN.}

\section{Charting-based methods}
\label{sec:spat}

We will now consider the second sub-class of non-Euclidean learning problems, where we are given multiple domains. A prototypical application the reader should have in mind throughout this section is the problem of finding correspondence between shapes, modeled as manifolds (see insert IN7). % in the field of computer graphics and vision. 
As we have seen, defining convolution in the frequency domain has an inherent drawback of inability to adapt the model across different domains. We will therefore need to resort to an alternative generalization of the convolution in the spatial domain that does not suffer from this drawback. % (see insert IN6). 

Furthermore, note that in the setting of multiple domains, there is no immediate way to define a meaningful spatial pooling operation, as the number of points on different domains can vary, and their order be arbitrary. 
It is however possible to pool point-wise features produced by a network by aggregating all the local information into a single vector. 
One possibility for such a pooling is computing the statistics of the point-wise features, e.g. the mean or covariance \cite{masci2015geodesic}. 
%%
%\begin{eqnarray}
%\label{eq:covpooling}
%\mathbf{F}^\mathrm{out} &=& \int_\mathcal{X} (\mathbf{f}^\mathrm{in}(x) - \boldsymbol{\mu}) (\mathbf{f}^\mathrm{in}(x) - \boldsymbol{\mu})^\top dx; \\
% \boldsymbol{\mu} &=& \int_\mathcal{X} \mathbf{f}^\mathrm{in}(x) dx. \nonumber
%\end{eqnarray}
%%
%(here $\mathbf{f}^\mathrm{in}(x)$ denotes a $q$-dimensional feature vector at point $x$, and $\mathbf{F}^\mathrm{out}$ is the $q\times q$ covariance matrix). 
Note that after such a pooling all the spatial information is lost.

%The key feature of CNNs is the convolutional layer, implementing the idea of ``weight sharing'', wherein a small set of templates (filters) is applied to different parts of the data.
%%
%%

%\paragraph*{\bf Charting-based methods}%Intrinsic patch operator}
%An alternative way is to generalize convolution in the spatial domain.  
%
%In image analysis applications, the input into the CNN is a function representing pixel values given on a Euclidean domain (plane); 
On a Euclidean domain, due to shift-invariance the convolution can be thought of as passing a template at each point of the domain and  recording the correlation of the template with the function at that point. Thinking of image filtering, this amounts to extracting a (typically square) patch of pixels, multiplying it element-wise with a template and summing up the results, then moving to the next position in a sliding window manner. Shift-invariance implies that the very operation of extracting the patch at each position is always the same. 

%{\color{green} ADS: should we talk about locally connected networks a bit?}

One of the major problems in applying the same paradigm to non-Euclidean domains is the lack of shift-invariance, implying that the `patch operator' extracting a local `patch' would be position-dependent. 
Furthermore, the typical lack of meaningful global parametrization for a graph or manifold forces to represent the patch in some local intrinsic system of coordinates. 
%While there is usually no meaningful global parametrization for a graph or manifold, it is possible to create a local system of coordinates $\mathbf{u}(x,\xi)$ mapping a point $\xi$ in the vicinity of $x$ to a low-dimensional Euclidean space. 
%
%
Such a mapping can be obtained by defining a set of weighting functions $v_1(x,\cdot), \hdots, v_J(x,\cdot)$ localized to positions near $x$ (see examples in Figure~3). %\ref{fig:patch1}). 
%For example, the weights could represent a discrete geodesic polar system of coordinates around $x$. 
Extracting a patch amounts to averaging the function $f$ at each point by these weights,  
\begin{eqnarray}
\label{eq:intpatch}
D_j(x) f &=& \int_{\mathcal{X}} f(x') v_j(x,x') dx', \hspace{3mm} j = 1,\hdots, J, 
\end{eqnarray}
%template now has to be location-dependent. 
%
providing for a spatial definition of an intrinsic equivalent of convolution %can then be defined as 
\begin{eqnarray}
\label{eq:intconv}
(f \star g)(x) &=& \sum_j  g_j D_j (x)f, 
\end{eqnarray}
where $g$ denotes the template coefficients applied on the patch extracted at each point. 
Overall, (\ref{eq:intpatch})--(\ref{eq:intconv}) act as a kind of non-linear filtering of $f$, and the patch operator $D$ is specified by defining the weighting functions $v_1, \hdots, v_J$. \editMB{Such filters are localized by construction, and the number of parameters is equal to the number of weighting functions $J = \mathcal{O}(1)$. 
Several frameworks for non-Euclidean CNNs essentially amount to different choice of these weights. The spectrum-free methods (ChebNet and GCN) described in the previous section can also be thought of in terms of local weighting functions, as it is easy to see the analogy between formulae~(\ref{eq:intconv}) and (\ref{eq:filt_cheby}).
}

%
%%\paragraph*{\bf 
%{\em Diffusion CNN:}  
%The simplest local charting on a non-Euclidean domain $\mathcal{X}$ is a one-dimensional coordinate measuring the intrinsic (e.g. geodesic or diffusion) distance $d(x,\cdot)$ \cite{atwood2016search}. The weighting functions in this case, for example chosen as Gaussians
%\begin{eqnarray}
%v_i(x,x') &=& e^{-(d(x,x') - \rho_i)^2/2\sigma^2}
%\end{eqnarray}
%%in this case 
%have the shape of rings of width $\sigma$ at distances $\rho_1, \hdots, \rho_J$ (Figure~\ref{fig:patch1}, left). 
%%, allowing to distinguish between neighboring vertices that are one-, two- or more hops away. 
%%
%
%

%\paragraph*{\bf 
{\bf{\em Geodesic CNN}} \cite{masci2015geodesic}: % on manifolds} 
Since manifolds naturally come with a low-dimensional tangent space associated with each point, it is natural to work in a local system of coordinates in the tangent space. 
In particular, on two-dimensional manifolds one can create a polar system of coordinates around $x$ where the radial coordinate is given by some intrinsic distance $\rho(x') = d(x,x')$, and the angular coordinate $\theta(x)$ is obtained by ray shooting from a point at equi-spaced angles. The weighting functions in this case can be obtained as a product of Gaussians 
\begin{eqnarray}
v_{ij}(x,x') &=& e^{-(\rho(x') - \rho_i)^2/2\sigma_\rho^2} \, \, e^{-(\theta(x') - \theta_j)^2/2\sigma_\theta^2},
\end{eqnarray}
where $i = 1,\hdots, J$ and $j = 1,\hdots, J'$ denote the indices of the radial and angular bins, respectively. 
The resulting $JJ'$ weights are bins of width $\sigma_\rho \times \sigma_\theta$ in the polar coordinates (Figure~3, right). 
%\ref{fig:patch1}, right). 
 %
%The diffusion CNN can be considered as a particular setting where only one angular bin and $\sigma_\theta = \infty $ are used. 

%\paragraph*{\bf 
{\bf{\em Anisotropic CNN}} \cite{boscaini2016learning}: % on manifolds}
We have already seen the non-Euclidean heat equation~(\ref{eq:diffusion}), whose heat kernel $h_t(x,\cdot)$ produces localized blob-like weights around the point $x$ (see~FIGS4). Varying the diffusion time $t$ controls the spread of the kernel. However, such kernels are {\em isotropic}, meaning that the heat flows equally fast in all the directions. 
A more general {\em anisotropic diffusion} equation on a manifold 
\begin{equation}
\label{eq:heat_eqa}
f_t(x,t) = -\text{div}(\mathbf{A}(x) \nabla f(x,t) ),
\end{equation}
%
%Here $\nabla$ and $\text{div}$ denote the intrinsic gradient and divergence operators, and $f(x,t)$ is the temperature at point $x$ at time $t$. 
involves the {\em thermal conductivity tensor} 
$\mathbf{A}(x)$ (in case of two-dimensional manifolds, a $2\times 2$ matrix applied to the intrinsic gradient in the tangent plane at each point), allowing modeling heat flow that is position- and direction-dependent \cite{andreux2014anisotropic}. % ({\em anisotropic}). 
A particular choice of the heat conductivity tensor proposed in \cite{boscaini2016anisotropic} is 
\begin{equation}
\label{eq:aniso_tensor}
\mathbf{A}_{\alpha \theta}(x) = \mathbf{R}_\theta(x)
\begin{pmatrix}
\alpha & \\
 & 1 
\end{pmatrix}
\mathbf{R}^\top_\theta(x),
\end{equation}
where the $2\times 2$ matrix $\mathbf{R}_\theta(x)$ performs rotation of $\theta$ w.r.t. to some reference (e.g. the maximum curvature) direction and  
%Such a thermal conductivity tensor drives the diffusion in some direction, where 
$\alpha > 0$ is a parameter controlling the degree of anisotropy ($\alpha = 1$ corresponds to the classical isotropic case). 
%
%\paragraph*{Anisotropic Laplacian.}
The heat kernel of such anisotropic diffusion equation is given by the spectral expansion 
\begin{equation}
\label{eq:ahks}
h_{\alpha\theta t}(x,x') = \sum_{i\geq 0} e^{-t \lambda_{\alpha\theta i}} \phi_{\alpha\theta i}(x) \phi_{\alpha\theta i}(x'),  
\end{equation}
where $\phi_{\alpha\theta 0}(x), \phi_{\alpha\theta 1}(x), \hdots$ are the eigenfunctions and 
$ \lambda_{\alpha\theta 0},  \lambda_{\alpha\theta 1}, \hdots$ the corresponding eigenvalues 
%$\{ \phi_{\alpha\theta i}, \lambda_{\alpha\theta i} \}_{i\geq 0}$ 
%are the eigenfunctions and eigenvalues 
of the {\em anisotropic Laplacian}
\begin{equation}
\Delta_{\alpha\theta}f(x) = -\text{div} (\mathbf{A}_{\alpha \theta}(x) \nabla f(x)). 
\end{equation}
The discretization of the anisotropic Laplacian is a modification of the cotangent formula~(\ref{eq:cotan}) on meshes or graph Laplacian~(\ref{eq:graph_lap}) on point clouds \cite{boscaini2016learning}.

The anisotropic heat kernels $h_{\alpha\theta t}(x,\cdot)$ look like elongated rotated blobs (see Figure~3, center), %\ref{fig:patch1}, center), 
where the parameters $\alpha, \theta$ and $t$ control the elongation, orientation, and scale, respectively. 
Using such kernels as weighting functions $v$ in the construction of the patch operator~(\ref{eq:intpatch}), it is possible to obtain a charting similar to the geodesic patches (roughly, $\theta$ plays the role of the angular coordinate and $t$ of the radial one).

%A limitation of the spatial generalization of CNNs based on patch operators is the assumption of some local low-dimensional structure in which a meaningful system of coordinates can be defined. While very natural on manifolds (where the tangent space is such a low-dimensional space), such a definition is significantly more challenging on graphs. 
%%
%In particular, defining anisotropic diffusion on general graphs seems to be an intriguing but hard problem.

\editMB{

{\bf{\em Mixture model network (MoNet)}} \cite{monti2016geometric}: Finally, as the most general construction of patches, 
Monti et al. \cite{monti2016geometric} proposed defining at each point a local system of $d$-dimensional pseudo-coordinates $\mathbf{u}(x,x')$ around $x$.  On these coordinates, a set of parametric kernels $v_1(\mathbf{u}), \hdots, v_J(\mathbf{u}))$ is applied, producing the weighting functions in~(\ref{eq:intpatch}). Rather than using fixed kernels as in the previous constructions, Monti et al. use Gaussian kernels 
$$
v_j(\mathbf{u}) = \exp \left( -\tfrac{1}{2}(\mathbf{u}-\boldsymbol{\mu}_j)^\top\boldsymbol{\Sigma}_j^{-1}(\mathbf{u}-\boldsymbol{\mu}_j) \right)
$$
whose parameters ($d\times d$ covariance matrices $\boldsymbol{\Sigma}_1, \hdots, \boldsymbol{\Sigma}_J$ and $d\times 1$ mean vectors $\boldsymbol{\mu}_1, \hdots, \boldsymbol{\mu}_J$) are learned.\footnote{This choice allow interpreting intrinsic convolution~(\ref{eq:intconv}) as a mixture of Gaussians, hence the name of the approach. }
Learning not only the filters but also the patch operators in~(\ref{eq:intconv}) affords additional degrees of freedom to the MoNet architecture, which makes it currently the state-of-the-art approach in several applications.  
It is also easy to see that this approach generalizes the previous models, and e.g. classical Euclidean CNNs as well as Geodesic- and Anisotropic CNNs can be obtained as particular instances thereof \cite{monti2016geometric}. 
MoNet can also be applied on general graphs using as the pseudo-coordinates $\mathbf{u}$ some local graph features such as vertex degree, geodesic distance, etc. 
}

\begin{figure}[h!]
\centering
\begin{overpic}
[width=1\linewidth]{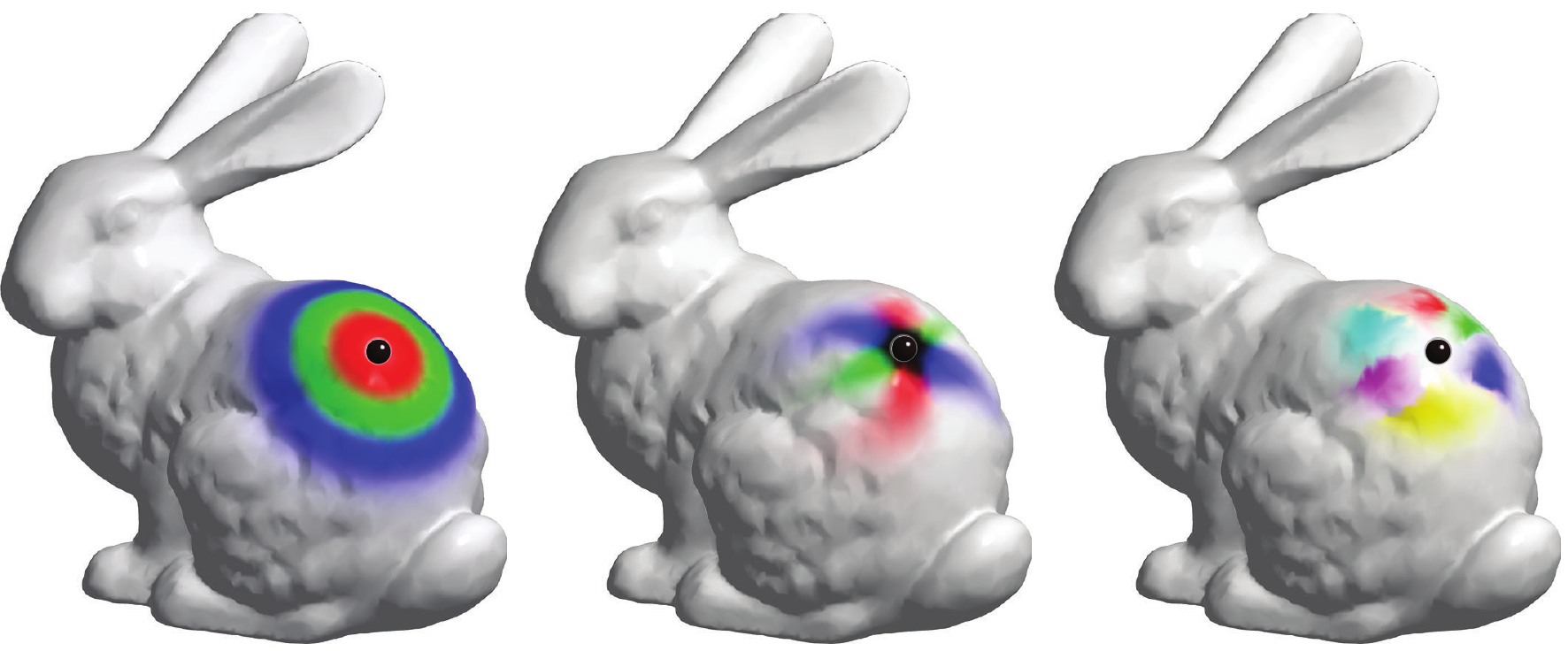}
\put(5,-4){\small Diffusion distance}
\put(45,-4){\small Anisotropic}
\put(45.25,-8){\small heat kernel}
\put(75,-4){\small Geodesic polar}
\put(78.5,-8){\small coordinates}
\end{overpic}\vspace{8mm}\\
\begin{minipage}{0.32\linewidth}
	\centering
	\setlength\figureheight{3cm} 
	\setlength\figurewidth{3cm} % \linewidth
	\input{polar_plot_concentric_kernels}%\\
%		\footnotesize GCNN
\end{minipage}
\begin{minipage}{0.32\linewidth}
	\centering
	\setlength\figureheight{3cm} 
	\setlength\figurewidth{3cm}
	\input{polar_plot_acnn_kernels}%\\
%		\footnotesize ACNN
\end{minipage}
\begin{minipage}{0.32\linewidth}
	\centering
	\setlength\figureheight{3cm} 
	\setlength\figurewidth{3cm}
	\input{polar_plot_gcnn_kernels}%\\
%		\footnotesize MoNet
\end{minipage}\\
\label{fig:patch1}
\caption{
\small Top: examples of intrinsic weighting functions used to construct a patch operator at the point marked in black (different colors represent different weighting functions). 
Diffusion distance (left) allows to map neighbor points according to their distance from the reference point, thus defining a one-dimensional system of local intrinsic coordinates. 
Anisotropic heat kernels (middle) of different scale and orientations and geodesic polar weights (right) are two-dimensional systems of coordinates. 
\editMB{Bottom: representation of the weighting functions in the local polar $(\rho,\theta)$ system of coordinates (red curves represent the 0.5 level set). }
%  
%Shown are (left to right): diffusion distance rings, anisotropic heat kernels, geodesic polar coordinates (radial and angular coordinates). 
}
\end{figure}

\section{Combined spatial/spectral methods}
\label{sec:spat_freq}

The third alternative for constructing convolution-like operations of non-Euclidean domains is jointly in spatial-frequency domain. 

\paragraph*{\bf Windowed Fourier transform}

One of the notable drawbacks of classical Fourier analysis is its lack of spatial localization. By virtue of the {\em Uncertainty Principle}, one of the fundamental properties of Fourier transforms, spatial localization comes at the expense of frequency localization, and vice versa.  
In classical signal processing, this problem is remedied by localizing frequency analysis in a window $g(x)$, leading to the definition of the {\em Windowed Fourier Transform} (WFT, also known as {\em short-time Fourier transform} or {\em spectrogram} in signal processing), 
\begin{eqnarray}
\label{eq:wft1}
(Sf)(x,\omega) &=& \int_{-\infty}^\infty f(x') \underbrace{g(x'-x) e^{-i\omega x'}}_{\overline{g_{x,\omega}}(x')} dx' \\
&=& \langle f, g_{x,\omega}\rangle_{L^2(\mathbb{R})}.
\label{eq:wft2} 
\end{eqnarray}
The WFT is a function of two variables: spatial location of the window $x$ and the modulation frequency $\omega$. The choice of the window function $g$ allows to control the tradeoff between spatial and frequency localization (wider windows result in better frequency resolution). 
Note that WFT can be interpreted as inner products~(\ref{eq:wft2})  of the function $f$ with translated and modulated windows $g_{x,\omega}$, referred to as the WFT {\em atoms}.

The generalization of such a construction to non-Euclidean domains requires the definition of translation and modulation operators \cite{shuman2016vertex}. While modulation simply amounts to multiplication by a Laplacian eigenfunction, translation is not well-defined due to the lack of shift-invariance. 
It is possible to resort again to the spectral definition of a convolution-like operation~(\ref{eq:gconv}), defining translation as  convolution with a delta-function, 
\begin{eqnarray}
(g \star \delta_{x'})(x) &=& \sum_{i\geq 0} \langle g, \phi_i \rangle_{L^2(\mathcal{X})} \langle \delta_{x'}, \phi_i \rangle_{L^2(\mathcal{X})} \phi_i(x) \nonumber \\
%& = & \sum_{i \geq 0}  \langle g, \phi_i \rangle_{L^2(\mathcal{X})}  \phi_i(\xi) \phi_i(x) 
&=& \sum_{i \geq 0}  \hat{g}_i  \phi_i(x') \phi_i(x). 
\end{eqnarray}
The translated and modulated atoms can be expressed as  
\begin{eqnarray}
g_{x', j}(x) & = & \phi_j(x') \sum_{i \geq 0}  \hat{g}_i \phi_i(x) \phi_i(x'), 
\end{eqnarray}
where the window is specified in the spectral domain by its Fourier coefficients $\hat{g}_i$; 
the WFT on non-Euclidean domains thus takes the form 
\begin{eqnarray}
\label{eq:gwft}
(Sf)(x',j) =  \langle f, g_{x',j}\rangle_{L^2(\mathcal{X})} = \sum_{i \geq 0}  \hat{g}_i \phi_i(x')  \langle f, \phi_i \phi_j\rangle_{L^2(\mathcal{X})}.  
\end{eqnarray}
Due to the intrinsic nature of all the quantities involved in its definition, the WFT is also intrinsic.

\paragraph*{\bf Wavelets}

Replacing the notion of frequency in time-frequency representations by that of scale leads to wavelet decompositions. Wavelets have 
been extensively studied in general graph domains \cite{coifman2006diffusionw}. Their objective is to define 
 stable linear decompositions with atoms well localized both in space and frequency that can efficiently approximate 
 signals with isolated singularities. Similarly to the Euclidean setting, 
 wavelet families can be constructed either from its spectral constraints or from its spatial constraints.

 The simplest of such families are Haar wavelets. Several bottom-up wavelet constructions on graphs were studied in \cite{szlam2005diffusion} and \cite{gavish2010multiscale}. In \cite{rustamov2013wavelets}, the authors developed an unsupervised method that learns wavelet decompositions on graphs by optimizing a sparse reconstruction objective.
In \cite{cheng2016deep}, ensembles of Haar wavelet decompositions were used to define deep wavelet scattering transforms on general domains, obtaining excellent numerical performance. Learning amounts to finding optimal pairings of nodes at each scale, which can be efficiently solved in polynomial time.

\editMB{
{\bf{\em Localized Spectral CNN (LSCNN)}} \cite{boscaini2015learning}: 
Boscaini et al. used the WFT as a way of constructing patch operators~(\ref{eq:intpatch}) on manifolds and point clouds and used in an intrinsic convolution-like construction~(\ref{eq:intconv}). 
The WFT allows expressing a function around a point in the spectral domain in the form $D_j(x) f = (Sf)(x,j)$. Applying learnable filters to such `patches' (which in this case can be interpreted as spectral multipliers), it is possible to extract meaningful features that also appear to generalize across different domains.  
An additional degree of freedom is the definition of the window, which can also be learned \cite{boscaini2015learning}.  
}

\myframedtext[0.95\linewidth]{
\editMB{
%\paragraph*{\bf Notation}\\
%\paragraph*{\bf Geometric deep learning methods} 
\begin{tabular}{ lcc }
\multicolumn{3}{c} {\bf Dichotomy of Geometric deep learning methods} \vspace{2mm}\\
{\bf Method} & {\bf Domain} & {\bf Data}  \\
  {\em Spectral CNN} \cite{bruna2013spectral} & spectral & graph\\
  {\em GCNN/ChebNet} \cite{defferrard2016convolutional} & spec. free & graph\\
  {\em GCN} \cite{welling2016} & spec. free & graph \\
    {\em GNN} \cite{GNN} & spec. free & graph \\
  %
%  {\em Diffusion CNN} \cite{atwood2016search} & spatial & graph\\ 
  %
  {\em Geodesic CNN} \cite{masci2015geodesic} & charting & mesh \\
  {\em Anisotropic CNN} \cite{boscaini2016learning}\hspace{-2.5mm} & charting & mesh/point cloud \\
  {\em MoNet} \cite{monti2016geometric} & charting &\hspace{-1mm} graph/mesh/point cloud\hspace{-1mm}\\
  {\em LSCNN} \cite{boscaini2015learning} & combined & mesh/point cloud \\
\end{tabular}
}
}

\section{Applications}
\label{sec:applications}

\editJB{
\paragraph*{\bf Network analysis}
One of the classical examples used in many works on network analysis are citation networks. Citation network is a graph where vertices represent papers and there is a directed edge $(i,j)$ if paper $i$ cites paper $j$. Typically, vertex-wise features representing the content of the paper (e.g. histogram of frequent terms in the paper) are available. 
A prototypical classification application is to attribute each paper to a field. 
Traditional approaches work vertex-wise, performing classification of each vertex's feature vector individually. More recently, it was shown that classification can be considerably improved using information from neighbor vertices, e.g. with a CNN on graphs \cite{defferrard2016convolutional,welling2016}. 
Insert IN6 shows an example of application of spectral and spatial graph CNN models on a citation network.

\myframedtext[0.90\linewidth]{
%\vspace{-5mm}
\paragraph*{\bf [IN6] Citation network analysis application}
The  CORA citation network \cite{sen2008collective} is a graph containing $2708$ vertices representing papers and $5429$ edges representing citations. Each paper is described by a $1433$-dimensional bag-of-words feature vector and 
belongs to seven classes. 
For simplicity, the network is treated as an undirected graph. 
Applying the spectral CNN with two spectral convolutional layers parametrized according to~(\ref{eq:welling}), the authors of \cite{welling2016} obtained classification accuracy of $81.6\%$ (compared to $75.7\%$ previous best result). \editMB{In \cite{monti2016geometric}, this result was slightly improved further, reaching $81.7\%$ accuracy with the use of MoNet architecture. }\vspace{3mm}\\
\begin{overpic}
[width=1\linewidth]{cora_monet.pdf}
%\put(19,-2){\small Spectral CNN (25\% error)}
%\put(70,-2){\small Diffusion CNN (19\% error)}
\end{overpic}\vspace{2mm}
{\small {\bf [FIGS6a]} \editMB{Classifying research papers in the CORA dataset with MoNet.} Shown is the citation graph, where each node is a paper, and an edge represents a citation. Vertex fill and outline colors represents the predicted and groundtruth labels, respectively; ideally, the two colors should coincide. (Figure reproduced from \cite{monti2016geometric}). 
%For visualization purposes, only test set vertices are visualized. \color{red}{[MB: this figure must be updated]} 
}
}

Another fundamental problem in network analysis is {\em ranking} and {\em community detection}. These can be estimated by solving an eigenvalue problem on an appropriately defined 
operator on the graph: for instance, the \emph{Fiedler vector} (the eigenvector associated with the smallest non-trivial eigenvalue of the Laplacian) carries information on the graph partition with minimal cut \cite{von2007tutorial}, and the popular PageRank algorithm approximates page ranks with the principal eigenvector of a modified Laplacian operator. In some contexts, one may want develop data-driven versions of such algorithms, that can adapt to model mismatch and perhaps provide a faster alternative to diagonalization methods. By unrolling power iterations, one obtains a Graph Neural Network architecture whose parameters can be learnt with backpropagation from labeled examples, similarly to the Learnt Sparse Coding paradigm \cite{gregor2010learning}. We are currently exploring this connection by constructing multiscale versions of graph neural networks. 

%
%\myframedtext*[\linewidth]{
%\vspace{-5mm}
%\begin{multicols}{2}
%\paragraph*{\bf [IN4] Citation network analysis example}
%...
%\end{multicols}
%\begin{overpic}
%[width=1\linewidth]{cora_dcnn.pdf}
%\put(19,-2){\small Spectral CNN (25\% error)}
%\put(70,-2){\small Diffusion CNN (19\% error)}
%\end{overpic}\vspace{6mm}
%{\small {\bf [FIGS4a]} Classifying research papers in the CoRA dataset. Shown is the citation graph, where each node is a paper, and an edge represents a citation. Vertex fill color represents the label predicted by the spectral (left) and diffusion CNN (right) models; vertex outline color represents the groundtruth label (ideally, the two colors should coincide). For visualization purposes, only test set vertices are visualized.}
%}
%

}

\begin{figure}[th!]
\centering
\scalebox{1}{		
\begin{overpic}
	[trim=0cm 0cm 0cm 0cm,clip,width=1.1\linewidth]{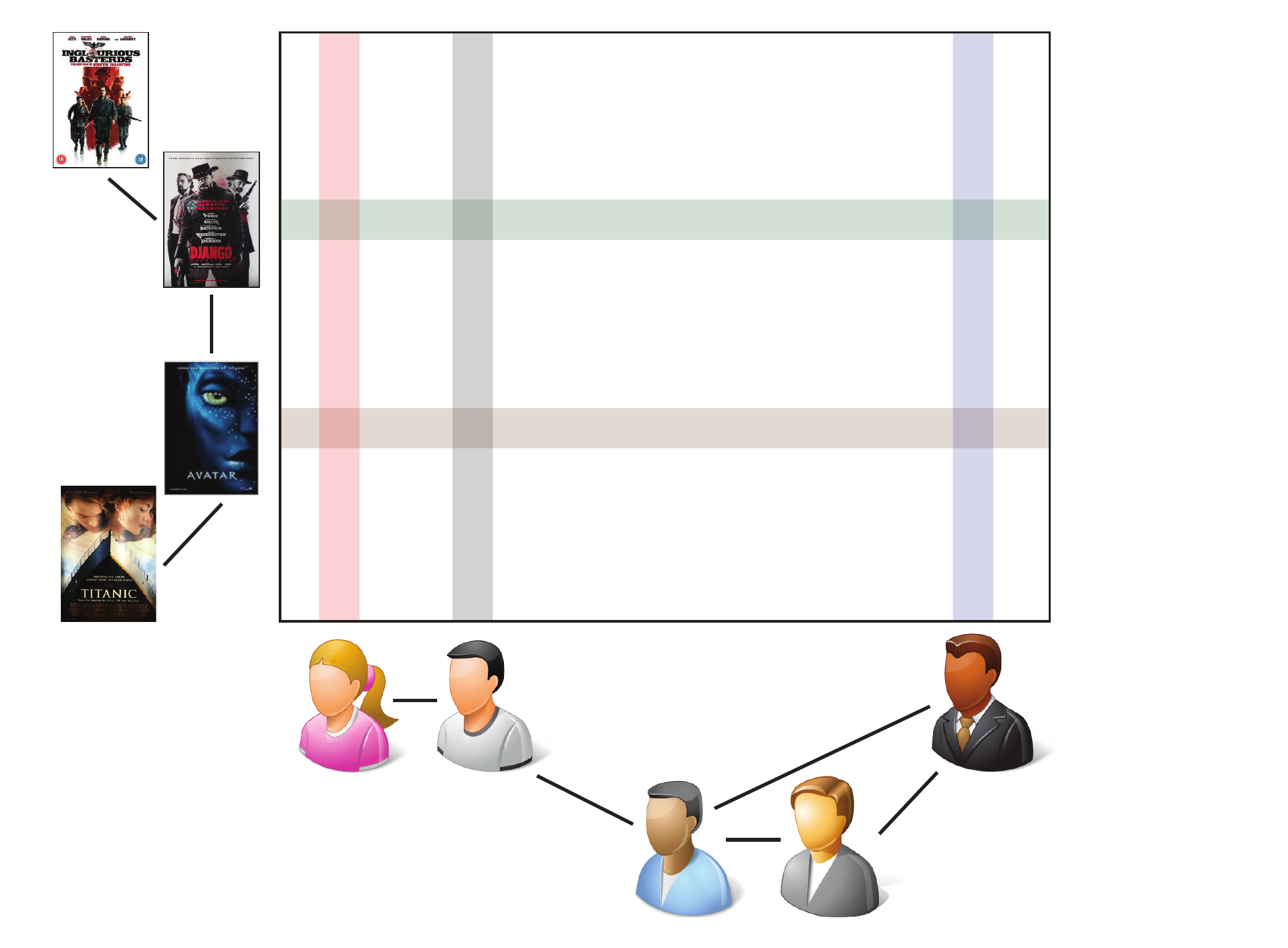}
			\put(46.5,22){\footnotesize $n$ users}
			\put(1,42){\rotatebox{90}{\footnotesize $m$ items}}
			%
%			\put(49,46.5){\footnotesize $\mathbf{X}$}			
			\put(25,73){\footnotesize $j_1$}
			\put(35,73){\footnotesize $j_2$}
			\put(50,73){\footnotesize $\hdots$}
			\put(73.75,73){\footnotesize $j_3$}
			\put(83,55){\footnotesize $i_2$}
			\put(83,46){\footnotesize $\vdots$}
			\put(83,39){\footnotesize $i_1$}
	\end{overpic}
	}\vspace{-3mm}
\caption{\editMB{Geometric matrix completion exemplified on the famous Netflix movie recommendation problem.  
The column and row graphs represent the relationships between users and items, respectively.  }
 }
\label{fig:matrixcomp}
\end{figure}
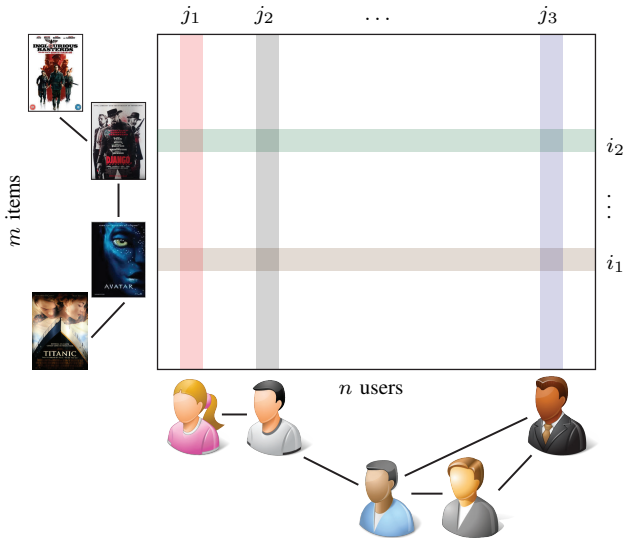

\editMB{

\paragraph*{\bf Recommender systems}

Recommending movies on Netflix, friends on Facebook, or products on Amazon are a few examples of {\em recommender systems} that have recently become ubiquitous in a broad range of applications. 
%%
%Two major approach to recommender systems are collaborative \cite{pro:BreeseHeckermanKadie98CollFilt} and content \cite{art:PazzaniBillsus07ContFilt} filtering techniques. Systems based on collaborative filtering use collected ratings of products by customers and offer new recommendations by finding similar rating patterns. Systems based on content filtering make use of similarities between products and customers to recommend new products. Hybrid systems combine collaborative and content techniques. 
%
Mathematically, a recommendation method can be posed as a {\em matrix completion} problem \cite{candes2012exact}, where columns and rows represent users and items, respectively, and matrix values represent a score determining whether a user would like an item or not. Given a small subset of known elements of the matrix, the goal is to fill in the rest. 
%
%This problem aims at recovering the missing values of a matrix given a (very) small set of its entries. 
A famous example is the Netflix challenge \cite{art:KorenBellVolinsky09MatFac} 
offered in 2009 and carrying a 1M\$ prize for the algorithm that can best predict user ratings for movies based on previous ratings. The size of the Netflix matrix is 480K movies $\times$ 18K users (8.5B elements), with only 0.011\% known entries.

Several recent works proposed to incorporate geometric structure into matrix completion problems \cite{art:MaZhouLiuLyuKing11RecomSys,kalofolias2014matrix,rao2015collaborative,kuang2016harmonic} in the form of column- and row graphs representing similarity of users and items, respectively (see Figure~\ref{fig:matrixcomp}). 
Such a {\em geometric matrix completion} setting makes meaningful e.g. the notion of  smoothness of the matrix values, and was shown beneficial for the performance of recommender systems.

%matrix completion \cite{candes2012exact} 
%geometric matrix completion on graphs \cite{kalofolias2014matrix}

%factorized \cite{srebro2004maximum}
%factorized geometric \cite{rao2015collaborative} 

In a recent work, Monti et al. \cite{monti2017mc} proposed addressing the geometric matrix completion problem by means of a learnable model combining a {\em Multi-Graph CNN} (MGCNN) and a recurrent neural network (RNN). 
Multi-graph convolution can be thought of a generalization of the standard bi-dimensional image convolution, where  the domains of the rows and the columns are now different (in our case, user- and item graphs). The features extracted from the score matrix by means of the MGCNN are then passed to an RNN, which produces a sequence of incremental updates of the score values.  
Overall, the model can be considered as a learnable diffusion of the scores, with the main advantage compared to traditional approach being a fixed number of variables independent of the matrix size.
MGCNN achieved state-of-the-art results on several classical matrix completion challenges and, on a more conceptual level, could be a very interesting practical application of geometric deep learning to a classical signal processing problem.
}

\paragraph*{\bf Computer vision and graphics}
The computer vision community has recently shown an increasing interest in working with 3D geometric data, mainly due to the emergence of affordable range sensing technology such as Microsoft Kinect or Intel RealSense. Many machine learning techniques successfully working on images were tried ``as is'' on 3D geometric data, represented for this purpose in some way ``digestible'' by standard frameworks, e.g. as range images \cite{su2015multi,wei2016dense} or rasterized volumes \cite{wu20153d,qi2016volumetric}. 
The main drawback of such approaches is their treatment of geometric data as Euclidean structures. 
% (see Figure~\ref{fig:extint}). 
First, for complex 3D objects, Euclidean representations such as depth images or voxels may lose significant parts of the object or its fine details, or even break its topological structure. Second, Euclidean representations are not intrinsic, and vary when changing pose or deforming the object. Achieving invariance to shape deformations, a common requirement in many vision applications, demands very complex models and huge training sets due to the large number of degrees of freedom involved in describing non-rigid deformations (Figure~5, left).

%\begin{figure}[h!]
%\centering
%%\includegraphics[width=0.42\linewidth]{features_corr.pdf} \hspace{20mm}
%%\includegraphics[width=0.42\linewidth]{features_sim.pdf}
%\begin{overpic}
%[width=1\linewidth]{extint.pdf}
%\put(17,-4){\small Extrinsic}
%\put(70,-4){\small Intrinsic}
%\end{overpic}
%\label{fig:extint}
%\caption{
%\small Left: extrinsic methods such as volumetric CNNs treat 3D
%geometric data in its Euclidean representation. Such a representation is
%not invariant to deformations (e.g., in the shown example, the filter that
%responds to features on a straight cylinder would not respond to a bent
%one). Right: in an intrinsic representation, the filter is applied on the
%surface itself, thus being invariant to deformations.
%}
%\end{figure}

In the domain of computer graphics, on the other hand, working intrinsically with geometric shapes is a standard practice. In this field, 3D shapes are typically modeled as Riemannian manifolds and are discretized as meshes. Numerous studies (see, e.g. \cite{bronstein2006generalized,bronstein2010scale,kim2011blended,bronstein2011shape,ovsjanikov2012functional}) have been devoted to designing local and global features e.g. for establishing similarity or correspondence between deformable shapes with guaranteed invariance to isometries.   %properties.  
%%
%Two well-studied classes of deformations are {\em isometries} (metric-preserving transformations, consequently also preserving local areas and angles) and {\em conformal} (angle-preserving) deformations. 
%%
%The former model suits well inelastic and articulated motions, such as different poses of the human body, but is unable to capture significant shape variability (e.g. matching people of different stature or complexion). 
%%
%The class of conformal maps, on the other hand, is way too large: a classical result in differential geometry known as the {\em Uniformization Theorem} states that any closed simply-connected surface can be conformally mapped to a sphere \cite{poincare1908uniformisation}. 
%%
%Apparently, there are no other deformation classes that are larger than isometries but smaller than conformal. 

\begin{figure}[t!]
\centering
\begin{overpic}
[width=1\linewidth]{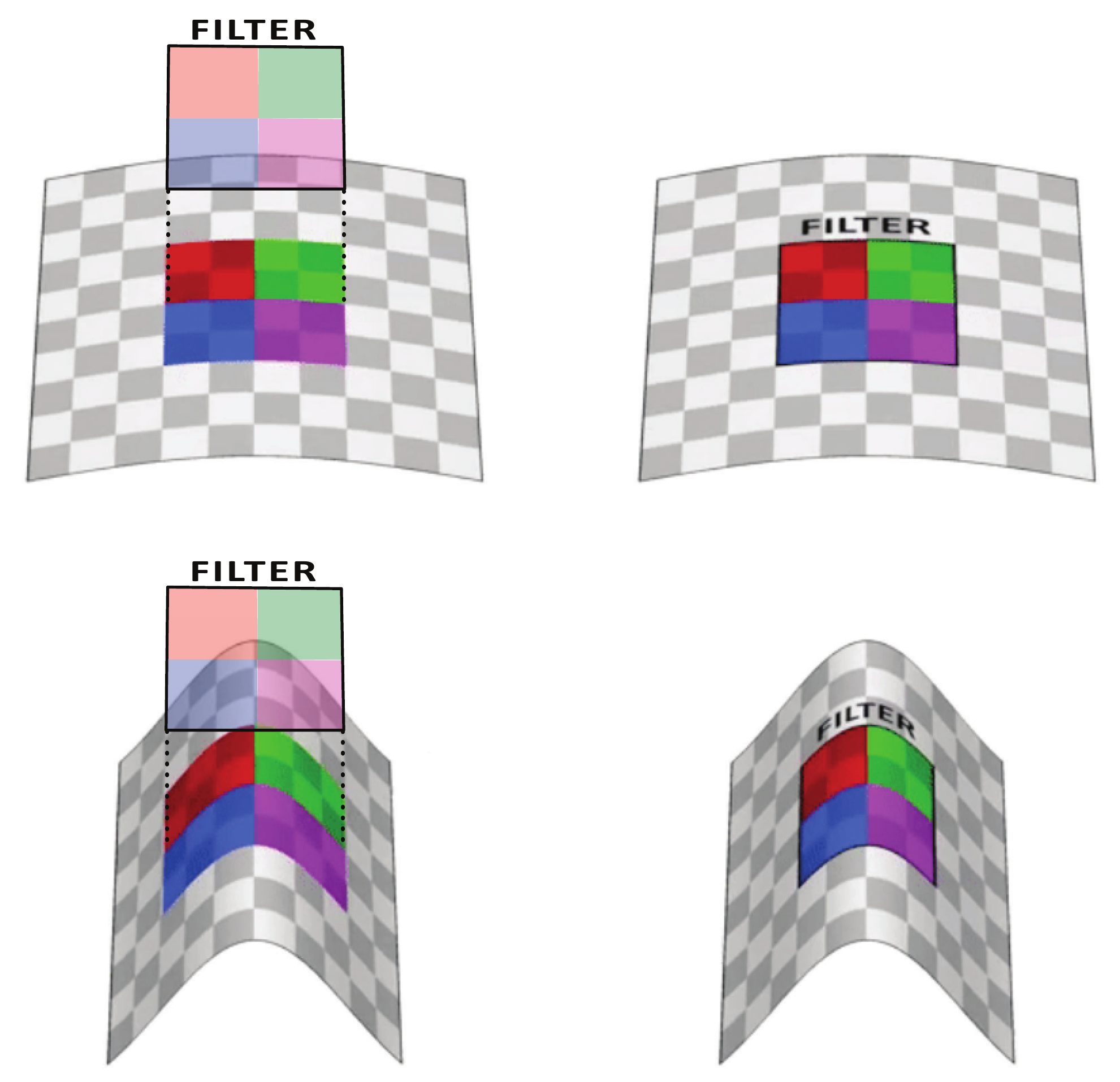}
\put(12,-4){\small Euclidean CNN}
\put(65.5,-4){\small Geometric CNN}
\end{overpic}
\label{fig:icnn1}
\caption{
\small \editMB{Illustration of the difference between classical CNN (left) applied to a 3D shape (checkered surface) considered as a Euclidean object, and a geometric CNN (right) applied intrinsically on the surface. In the latter case, the convolutional filters (visualized as a colored window) are deformation-invariant by construction.  }
}
\end{figure}

\begin{figure}[h!]
\centering
\begin{overpic}
[width=1\linewidth]{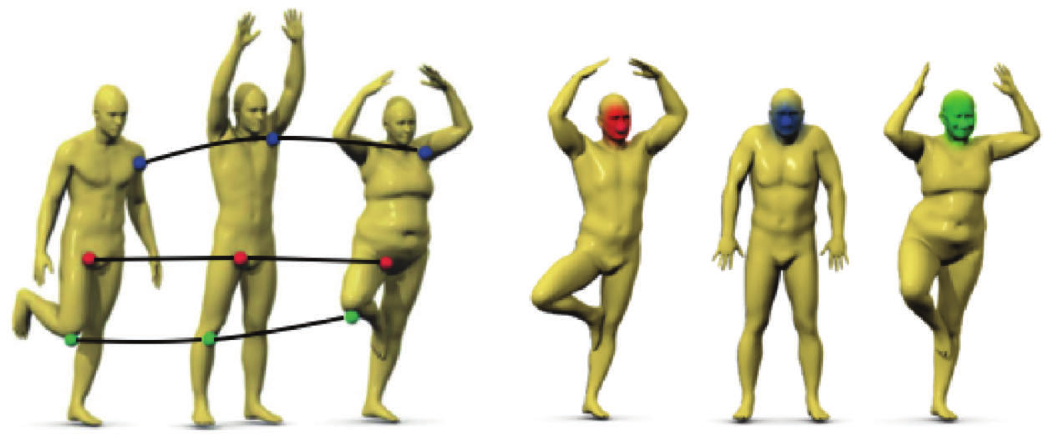}
\put(12,-4){\small Correspondence}
\put(68,-4){\small Similarity}
\end{overpic}
\label{fig:graphics1}
\caption{
\small Left: features used for shape correspondence should ideally manifest invariance across the shape class (e.g., the ``knee feature'' shown here should not depend on the specific person). Right: on the contrary, features used for shape retrieval should be specific to a shape within the class to allow distinguishing between different people. Similar features are marked with same color.  
Hand-crafting the right feature for each application is a very challenging task. 
}
\end{figure}

Furthermore, different applications in computer vision and graphics may require completely different features: for instance, in order to establish feature-based correspondence between a collection of human shapes, one would desire the descriptors of corresponding anatomical parts (noses, mouths, etc.) to be as similar as possible across the collection. In other words, such descriptors should be invariant to the collection variability. Conversely, for shape classification, one would like descriptors that emphasize the subject-specific characteristics, and for example, distinguish between two different nose shapes (see Figure~6). %\ref{fig:graphics1}). 
Deciding a priori which structures should be used and which should be ignored is often hard or sometimes even impossible. Moreover, axiomatic modeling of geometric noise such as 3D scanning artifacts turns out to be extremely hard. 

%The gap between the computer vision and graphics communities is owed to the fact that 

%On the one hand, t

%Put in a somewhat oversimplified manner, the computer vision community works with real-world 3D data, but uses Euclidean techniques originally developed for images that are not suitable for geometric data. 
%%
%At the same time, the mathematically rigorous models used in computer graphics to describe geometric objects can hardly deal with noisy data, leading to a tendency to work with idealized synthetic shapes.  
%%
%%Somewhat trivializing, we can say that in the computer vision community traditionally real-world 3D data using techniques suitable for images
%We believe that the gap between the two communities 
%%(computer vision, working with real-world 3D data using techniques suitable for images, and computer graphics, working with idealized clean data but with geometrically correct methods) 
%can be bridged with the development of geometric deep learning methods. 
%
By resorting to intrinsic deep neural networks, the invariance to isometric deformations is automatically built into the model, thus vastly reducing the number of degrees of freedom required to describe the invariance class. Roughly speaking, the intrinsic deep model will try to learn `residual' deformations that deviate from the isometric model. 
Geometric deep learning can be applied to several problems in 3D shape analysis, which can be divided in two classes. 
First, problems such as local descriptor learning \cite{masci2015geodesic,boscaini2016anisotropic} or correspondence learning \cite{boscaini2016learning} (see example in the insert IN7), in which the output of the network is {\em point-wise}. 
The inputs to the network are some point-wise features, for example, color texture or simple geometric features such as normals. 
% such as SHOT capturing the local normal vector orientations \cite{tombari2010unique}. 
Using a CNN architecture with  multiple intrinsic convolutional layers, it is possible to produce non-local features that capture the context around each point. % (which is important for finding correspondence between ambiguous regions). 
The second type of problems such as shape recognition require  the network to produce a  {\em global} shape descriptor, aggregating all the local information into a single vector using e.g. the covariance pooling \cite{masci2015geodesic}. %~(\ref{eq:covpooling}). 
% learning correspond

%\begin{figure*}[t!]
%\centering
%\includegraphics[width=0.42\linewidth]{features_corr.pdf} \hspace{20mm}
%\includegraphics[width=0.42\linewidth]{features_sim.pdf}
\myframedtext*[\linewidth]{
\vspace{-5mm}
\begin{multicols}{2}
\paragraph*{\bf [IN7] 3D shape correspondence application}
Finding intrinsic correspondence between deformable shapes is a classical tough problem that underlies a broad range of vision and graphics applications, including texture mapping, animation, editing, and scene understanding \cite{biasotti2015recent}. 
From the machine learning standpoint, correspondence can be thought of as a classification problem, where each point on the query shape is assigned to one of the points on a reference shape (serving as a ``label space'') \cite{rodola2014dense}. 
It is possible to learn the correspondence with a deep intrinsic network applied to some input feature vector $\mathbf{f}(x)$ at each point $x$ of the query shape $\mathcal{X}$, producing an output $U_{\boldsymbol{\Theta}}(\mathbf{f}(x))(y)$, which is interpreted as the conditional probability $p(y|x)$ of $x$ being mapped to $y$. %The hypervector $\boldsymbol{\Theta}$ denotes  of the network parameters). 
Using a training set of points with their ground-truth correspondence $\{ x_i, y_i \}_{i \in \mathcal{I}}$, supervised learning is performed minimizing the {\em multinomial regression loss} 
\begin{eqnarray}
    \label{eq:corresp}
   % \ell_{\text{reg}}(\boldsymbol{\Theta}) 
    %&=& - \sum_{(x, y^*(x))\in \mathcal{T}} \langle \delta_{y^*(x)} , \log \boldsymbol{f}_{\Theta}(x) \rangle_Y \nonumber \\
    \min_{\boldsymbol{\Theta}} \,\,\,\,\, -  \sum_{i \in \mathcal{I}} \log U_{\boldsymbol{\Theta}}(\mathbf{f}(x_i))(y_i)
    %f_{\boldsymbol{\Theta}}(x,y^*(x)),
\end{eqnarray}
w.r.t. the network parameters $\boldsymbol{\Theta}$. The loss penalizes for the deviation of the predicted correspondence from the groundtruth. 
We note that, while producing impressive result, such an approach essentially learns point-wise correspondence, which then has to be post-processed in order to satisfy certain properties such as smoothness or bijectivity. 
Correspondence is an example of structured output, where the output of the network at one point depends on the output in other points (in the simplest setting, correspondence should be smooth, i.e., the output at nearby points should be similar). 
Litany et al. \cite{litany2017} proposed {\em intrinsic structured prediction} of shape correspondence by integrating a layer computing functional correspondence \cite{ovsjanikov2012functional} into the deep neural network. 
\begin{center}
\begin{overpic}
[width=0.9\linewidth]{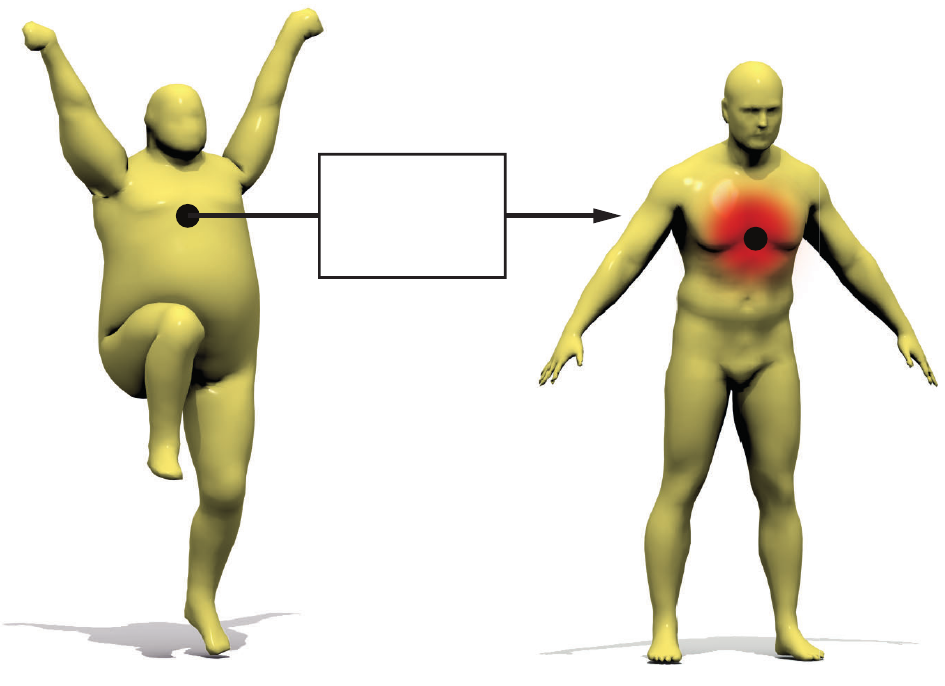}
\put(41.5,49){\small $U_{\Theta}$}
\put(18.5,46){\small $x_i$}
\put(78,43){\small $y_i$}
\put(27,3){\small $\mathcal{X}$}
\put(75,3){\small $\mathcal{Y}$}
\end{overpic}
\end{center}%\vspace{10mm}
{\small {\bf [FIGS7a]} Learning shape correspondence: an intrinsic deep network $U_{\boldsymbol{\Theta}}$ is applied point-wise to some input features defined at each point. The output of the network at each point $x$ of the query shape $\mathcal{X}$ is a probability distribution of the reference shape $\mathcal{Y}$ that can be thought of as a soft correspondence. }
\end{multicols}
\vspace{-2mm}
\begin{center}
\begin{overpic}
[width=1\linewidth]{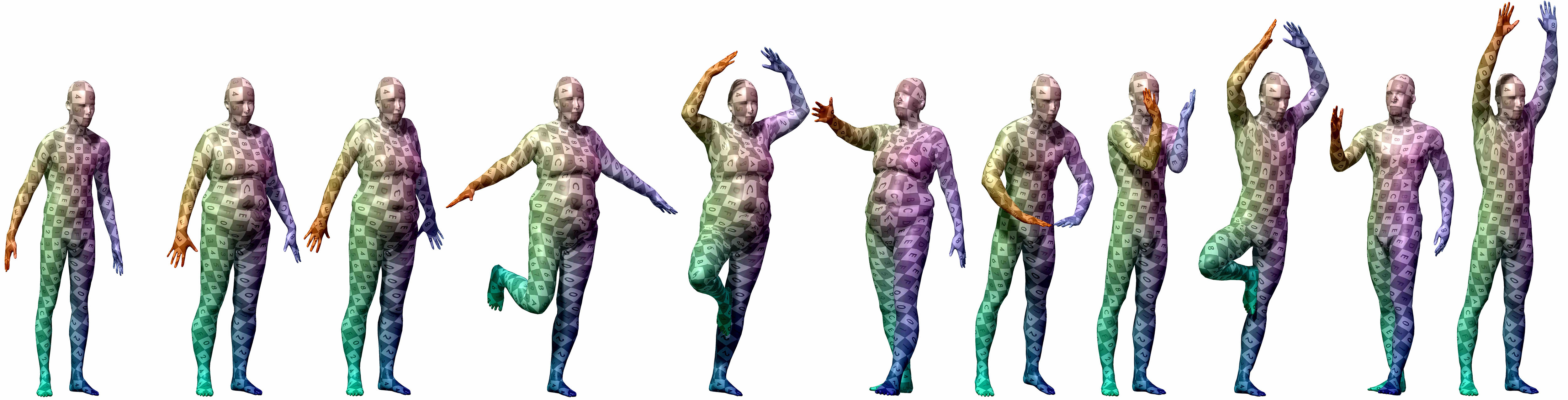} %acnn.pdf}
%\put(99.5,34.5){\footnotesize 0}
%\put(99.5,49.25){\footnotesize 10\%}
%\put(39,32){\small Blended intrinsic maps}
%\put(41.5,-0.5){\small Intrinsic CNN}
\end{overpic}%\vspace{1mm}
%
%\begin{overpic}
%[width=0.8\linewidth]{correspondence.pdf}
%%\put(10,-0.5){\small Non-Euclidean}
%\end{overpic}\\
%\label{fig:graphics1}
\end{center}
\editMB{\small {\bf [FIGS7b]} Intrinsic correspondence established between human shapes using intrinsic deep architecture (MoNet \cite{monti2016geometric} with three convolutional layers). SHOT descriptors capturing the local normal vector orientations \cite{tombari2010unique} were used in this example as input features. The correspondence is visualized by transferring texture from the leftmost reference shape. For additional examples, see \cite{monti2016geometric}. }
%{\small {\bf [FIGS2b]} Quality of intrinsic correspondence established between human shapes using a state-of-the-art non-learning approach (blended intrinsic maps \cite{kim2011blended}, first row) and learned using intrinsic deep architecture (Anisotropic CNN \cite{boscaini2016learning} with three convolutional layers, second row). SHOT descriptors capturing the local normal vector orientations \cite{tombari2010unique} were used in this example as input features. Shown is the correspondence deviation from the ground truth at each point, measured in $\%$ of geodesic diameter. Hotter colors represent larger errors.}
%
%Third and fourth rows: examples of correspondence between different shapes affected by extreme transformations (missing parts) learned by intrinsic CNN . Corresponding points are encoded in similar color. Reference shape is shown on the left. 
%\caption{
%}
%\end{figure*}
}

\editJB{
\paragraph*{\bf Particle physics and Chemistry}

Many areas of experimental science are interested in studying systems of discrete particles defined over a low-dimensional phase space. 
For instance, the chemical properties of a molecule are determined by the relative positions of its atoms, and the classification of events 
in particle accelerators depends upon position, momentum, and spin of all the particles involved in the collision. 

The behavior of an $N$-particle system is ultimately derived from solutions of the Schr{\"o}dinger equation, but its exact solution 
involves diagonalizing a linear system of exponential size. In this context, an important question is whether one can approximate the dynamics with a tractable model that incorporates by construction the geometric stability postulated by the Schr{\"o}dinger equation, and at the same time has enough flexibility to adapt to data-driven scenarios and capture complex interactions. 

An instance $l$ of an $N_l$-particle system can be expressed as
$$f_l(t) = \sum_{j=1}^{N_l} \alpha_{j,l} \delta(t - x_{j,l})~,$$
where $(\alpha_{j,l})$ model particle-specific information such as the spin, and $(x_{j,l})$ are the locations
of the particles in a given phase-space. Such system can be recast as a signal defined over a graph with $|\mathcal{V}_l| = N_l$  vertices and edge weights $\mathbf{W}_l = (\phi( \alpha_{i,l}, \alpha_{j,l}, x_{i,l}, x_{j,l}))$
%$( \mathcal{V}_l, \mathbf{W}_l )$, where $|\mathcal{V}_l| = N_l$ and $w_{ij,l} = \phi( \alpha_i, \alpha_j, x_i, x_j)$ 
expressed through a similarity kernel capturing the appropriate priors. 
Graph neural networks %(see Section \ref{sec:spat}) 
are currently being applied to perform event classification, energy regression, and anomaly detection in high-energy physics experiments such as the Large Hadron Collider (LHC) and neutrino detection in the IceCube Observatory. %\cite{hep_gnn}. 
Recently, models based on graph neural networks have been applied to predict the dynamics of $N$-body systems \cite{battaglia2016interaction, chang2016compositional} showing excellent prediction performance.

}

\editMB{

\paragraph*{\bf Molecule design}
A key problem in material- and drug design is predicting the physical, chemical, or biological properties of a novel molecule (such as solubility of toxicity) from its structure. 
State-of-the-art methods rely on hand-crafted molecule descriptors such as circular fingerprints \cite{morgan65generation,glem06circular,rogers10extended}. 
%
%Recently, Duvenaud et al. 
A recent work from Harvard university \cite{duvenaud2015convolutional} proposed modeling molecules as graphs (where vertices represents atoms and edges represent chemical bonds) and employing graph convolutional neural networks to learn the desired molecule properties. Their approach has significantly outperformed hand-crafted features. 
This work opens a new avenue in molecule design that might revolutionize the field.

\paragraph*{\bf Medical imaging}
An application area where signals are naturally collected on non-Euclidean domains and where the methodologies we reviewed could be very useful is brain imaging. A recent trend in neuroscience is to associate functional MRI traces with a pre-computed connectivity rather than inferring it from the traces themselves \cite{preti2016dynamic}. In this case, the challenge consists in processing and analyzing an array of signals collected over a complex topology, which results in subtle dependencies. 
In a recent work from Imperial College \cite{ktena2017distance}, graph CNNs were used to detect  disruptions of the brain functional networks associated with autism. 
%Ktena et al. \cite{rueckert}
%One should expect geometry-aware machine learning approaches to play an ever greater role in this field to help classify activity patterns and relate them to neurological cues. 
}

\section{Open problems and future directions}
\label{sec:concl}

%First, axiomatic approaches are limited in their ability to capture the complex structure of geometric data. 

The recent emergence of geometric deep learning methods in various communities and application domains, which we tried to overview in this paper, allow us to proclaim, perhaps with some caution, that we might be witnessing a new field being born. We expect the following  years to bring exciting new approaches and results, and conclude our review with a few observations of current key difficulties and potential directions of future research.

Many disciplines dealing with geometric data employ some empirical models or ``handcrafted'' features. This is a typical situation in  geometry processing and computer graphics, where axiomatically-constructed features are used to analyze 3D shapes, or computational sociology, where it is common to first come up with a hypothesis and then test it on the data \cite{lazer2009life}. Yet, such models assume some prior knowledge (e.g. isometric shape deformation model), and often fail to correctly capture the full complexity and richness of the data. 
In computer vision, departing from ``handcrafted'' features towards generic models learnable from the data in a task-specific manner has brought a breakthrough in performance and led to an overwhelming trend in the community to favor deep learning methods. 
Such a shift has not occurred yet in the fields dealing with geometric data due to the lack of adequate methods, but there are first indications of a coming paradigm shift.

%, which has occurred in computer vision in the past few years, is very slow or even not happening at all in other fields due to the lack of adequate learning techniques suitable for geometric data.  

%In some applications, geometric data can also be handled as a Euclidean structure, allowing to resort to classical deep learning techniques. 
%%However, Euclidean representation often fails to correctly capture the geometric structure of the data. 
%In deformation-invariant 3D shape correspondence application we mentioned in the context of computer graphics, 3D shapes can be considered both as 2D manifolds and as subsets of the 3D Euclidean space. The latter representation fails to correctly capture the geometric structure of the data, as it is extrinsic and not invariant under non-rigid deformations. While in principle it is possible to apply classical deep learning to Euclidean representations of non-rigid shapes, such models tend to be very complex and require large amounts of training data \cite{wei2016dense}. 
%%
%The main contribution of geometric deep learning in these settings is using a more suitable model with guaranteed invariance properties that appear to be much simpler than the Euclidean ones.

\paragraph*{\bf Generalization}
%Another important aspect is generalization capabilities and transfer learning. 
Generalizing deep learning models to geometric data requires not only finding non-Euclidean counterparts of basic building blocks (such as convolutional and pooling layers), but also generalization {\em across} different domains. 
Generalization capability is a key requirement in many applications, including computer graphics, where a model is learned on a training set of non-Euclidean domains (3D shapes) and then applied to previously unseen ones. 
%
%Recalling the approaches we mentioned in this review, 
Spectral formulation of convolution allows designing CNNs on a graph, but the model learned this way on one graph cannot be straightforwardly applied to another one, since the spectral representation of convolution is domain-dependent.  
\editMB{
A possible remedy to the generalization problem of spectral methods is the recent architecture proposed in \cite{spectralxformers}, applying the idea of spatial transformer networks \cite{xformers} in the spectral domain. This approach is reminiscent of the construction of compatible orthogonal bases by means of joint Laplacian diagonalization \cite{eynard2015multimodal}, which can be interpreted as an alignment of two Laplacian eigenbases in a $k$-dimensional space.
}

The spatial methods, on the other hand, allow generalization across different domains, but the construction of low-dimensional local spatial coordinates on graphs turns to be rather challenging. In particular, the construction of anisotropic diffusion on general graphs is an interesting research direction. 

\editADS{
The spectrum-free approaches also allow generalization across graphs, at least in terms of their functional form.   However, if multiple 
layers of equation \eqref{eq:gnn} used with no non-linearity or learned parameters $\boldsymbol{\theta}$, simulating a high power of the diffusion, the model may behave differently on different kinds of graphs.  Understanding under what circumstances and to what extent these methods generalize across graphs is currently being studied.}

\editMB{
\paragraph*{\bf Time-varying domains}
An interesting extension of geometric deep learning problems discussed in this review is coping with signals defined over a dynamically changing structure. In this case, we cannot assume a fixed domain and must track how these changes affect signals. This could prove useful to tackle applications such as abnormal activity detection in social or financial networks. In the domain of computer graphics and vision, potential applications deal with dynamic shapes (e.g. 3D video captured by a range sensor). 
}

\editMB{
\paragraph*{\bf Directed graphs}
Dealing with directed graphs is also a challenging topic, as such graphs typically have non-symmetric Laplacian matrices that do not have orthogonal eigendecomposition allowing easily interpretable spectral-domain constructions. 
Citation networks, which are directed graphs, are often treated as undirected graphs (including in our example in IN7) considering citations between two papers without distinguishing which paper cites which. This obviously may loose important information. 
}

\paragraph*{\bf Synthesis problems}
Our main focus in this review was primarily on {\em analysis} problems on non-Euclidean domains. 
Not less important is the question of data {\em synthesis}. 
%
%In the computer vision community, 
There have been several recent attempts to try to learn a {\em generative model} allowing to synthesize new images \cite{dosovitskiy2016learning} and speech waveforms  \cite{dieleman2016wavenet}. Extending such methods to the geometric setting seems a promising direction, though the key difficulty is the need to reconstruct the geometric structure (e.g., an embedding of a 2D manifold in the 3D Euclidean space modeling a deformable shape) from some intrinsic representation \cite{boscaini2015shape}.

%[GENERATIVE MODELS, SHAPE SYNTHESIS]

\paragraph*{\bf Computation}
The final consideration is a computational one. All existing deep learning software frameworks are primarily optimized for Euclidean data. One of the main reasons for the computational efficiency of deep learning architectures (and one of the factors that contributed to their renaissance) is the assumption of regularly structured data on 1D or 2D grid, allowing to take advantage of modern GPU hardware. Geometric data, on the other hand, in most cases do not have a grid structure, requiring different ways to achieve efficient computations. It seems that computational paradigms developed for large-scale graph processing are more adequate frameworks for such applications.

\section*{Acknowledgement}

The authors are grateful to Federico Monti, Davide Boscaini, Jonathan Masci, Emanuele Rodol{\`a}, Xavier Bresson, Thomas Kipf, and Micha{\"e}l Defferard for comments on the manuscript and for providing some of the figures used in this paper. %figures FIGS4a, FIGS5b, and Fig 5. 
This work was supported in part by the ERC Grants Nos. 307047 (COMET) and 724228 (LEMAN), Google Faculty Research Award, Radcliffe fellowship, Rudolf Diesel fellowship, and Nvidia equipment grants. 

%\ifCLASSOPTIONcaptionsoff
%  \newpage
%\fi

\bibliographystyle{IEEEtran}
\bibliography{refs}

%\input{bios.tex}

\end{document}